\documentclass[10pt, a4paper, conference]{IEEEtran}

\IEEEoverridecommandlockouts
\usepackage{cite}

\usepackage[utf8]{inputenc}
\usepackage{amsmath}
\usepackage{amssymb}

\usepackage{booktabs}

\usepackage[table]{xcolor}
\usepackage{graphicx}
\usepackage{tikz}
\usetikzlibrary{arrows}
\usepackage{subfig}
\DeclareCaptionFont{tinier}{\fontsize{6}{6}\selectfont}

\usepackage{xspace}
\usepackage{balance}
\usepackage[colorlinks=true, allcolors=black]{hyperref}

\newcommand{\slm}{$\Sigma\Lambda$\xspace}
\newcommand{\thead}[1]{{\small\textbf{#1}}}

\newcommand{\fliph}[1]{\scalebox{1}[-1]{#1}}
\newcommand{\rotae}[1]{\reflectbox{\rotatebox[origin=c]{-90}{#1}}}

\begin{document}

\title{Human or Machine? \\ It Is Not What You Write, But How You Write It}

\author{
    \IEEEauthorblockN{Luis A. Leiva}
    \IEEEauthorblockA{Aalto University, Finland\\
        name.surname@aalto.fi}
\and
    \IEEEauthorblockN{Moises Diaz}
    \IEEEauthorblockA{Universidad del\\ Atlantico Medio, Spain\\
        moises.diaz@atlanticomedio.es}
\and
    \IEEEauthorblockN{Miguel A. Ferrer}
    \IEEEauthorblockA{Universidad de Las Palmas\\ de Gran Canaria, Spain\\
        miguelangel.ferrer@ulpgc.es}
\and
    \IEEEauthorblockN{Réjean Plamondon}
    \IEEEauthorblockA{Polytechnique Montréal, Canada\\
        rejean.plamondon@polymtl.ca}
}

\maketitle

\begin{abstract}
Online fraud often involves identity theft.
Since most security measures are weak or can be spoofed,
we investigate a more nuanced and less explored avenue:
behavioral biometrics via handwriting movements.
This kind of data can be used to verify whether a user is operating a device or a computer application,
so it is important to distinguish between human and machine-generated movements reliably.
For this purpose, we study handwritten symbols (isolated characters, digits, gestures, and signatures)
produced by humans and machines, and compare and contrast several deep learning models.
We find that if symbols are presented as static images,
they can fool state-of-the-art classifiers (near 75\%\,accuracy in the best case)
but can be distinguished with remarkable accuracy
if they are presented as temporal sequences (95\%\,accuracy in the average case).
We conclude that an accurate detection of fake movements has more to do with \emph{how} users write,
rather than \emph{what} they write.
Our work has implications for computerized systems that need to authenticate or verify legitimate human users,
and provides an additional layer of security to keep attackers at bay.
\end{abstract}


\IEEEpeerreviewmaketitle

\section{Introduction}

Online fraud often involves identity theft,
and most of today's security measures are weak or can be spoofed.
Passwords can be guessed or are jotted down imprudently.
Two-factor authentication can be broken with SIM swap attacks~\cite{Andrews18}.
Newer phones, tablets, and laptops often include fingerprint and facial recognition,
but these can be spoofed as well~\cite{Menotti15}.
A plausible next level of security is to identify people using behavioral information,
since it is much harder to copy or imitate.
On the web, for example, it is possible to analyze mouse movements at no cost and at large scale~\cite{Leiva13_smt}.
Furthermore, some websites are starting to ask their users to solve
some form of handwritten captcha~\cite{Shirali-Shahreza11, Leiva15_mucaptcha, Ramahia14_capctha},
based on the assumption that handwriting input is very natural for humans.
Indeed, today's mobile devices have a touchscreen,
so users can simply hand-write with their fingers or a stylus both quickly and effortlessly.

In this context, we can think of a new form of biometric verification for online services based on handwritten symbols
such as gestures (geometric shapes or marks), characters, digits, and signatures
that users would have to enter on some touch-sensitive surface such as a mobile phone or tablet.
The main advantage of entering handwriting symbols, as opposed to regular handwriting,
is that they are short and easy to articulate.
In addition, gestures and digits are language-independent, so they are equally easy to learn for everyone.
Finally,
signatures comprise ballistic movements that people articulate almost without thinking~\cite{diaz2015modeling}.
Therefore, we can expect an increasing adoption of some form of handwriting-based verification in the future~\cite{Plamondon18_pdb}.
But then a practical question remains: is it possible to tell human and machine-generated handwriting movements apart?
Previous work~\cite{Galbally12b, 8052226, Leiva17_slm, Leiva17_gestures, Bhattacharya17, Reznakova17, plamondon2014recent, Almaksour11}
showed that users are unable to distinguish between them,
however the data were presented as static images, i.e. only spatial information was available to the users for assessment.
Therefore, it is unclear if a computer can do better in this classification task.
Furthermore, research on signature forgery detection and handwriting recognition~\cite{diaz2019perspective, Plamondon00}
has suggested that incorporating temporal information often improves classification performance.
If online services are to rely on behavioral data to prevent online fraud,
then we have to ensure that it is possible to distinguish between human and synthetic data reliably.
This idea is similar to the concept of ``liveness detection'' in the biometrics community~\cite{Ghiani15_fingerprint, Xin17_face, Chen18_iris}.
To the best of our knowledge, we are the first to tackle this problem in \emph{generic} handwriting movements using deep learning models.

In this paper, we contribute computational models that can tell human and machine-generated handwriting apart.
We build and contrast convolutional and recurrent neural networks to handle both off-line and on-line handwritten symbols
articulated on different devices (smartphone and tablet) using different input methods (stylus and finger).
We find that legitimate handwriting is hard to distinguish in off-line form even for a computer
(near 75\%\,accuracy in the best case),
however it is possible to achieve remarkable classification performance in on-line form
(95\%\,accuracy in the average case).
In other words, what really matters it is not \emph{what} you write but \emph{how} you write it.
Our work has implications for computerized systems that need to authenticate or verify legitimate human users,
and provides an additional layer of security that contributes to keeping attackers at bay.
Our code, models, and datasets are publicly available at \url{https://github.com/luileito/handwriting-biometrics}.

\section{Related Work}

Previous work that has investigated handwriting input as online verification
has been based on some form of CAPTCHAs (Completely Automated Public Turing Tests to Tell Computers and Humans Apart).
In Highlighting CAPTCHA~\cite{Shirali-Shahreza11} the user must trace an obfuscated word with a stylus,
however relying on precise user handwriting is cumbersome to perform on a mobile device
due to the inaccuracy of user input~\cite{Hourcade08, Plamondon00}.
Other approaches require the user to trace a symbol with their mouse or finger,
for example a gesture\footnote{\url{http://josscrowcroft.com/demos/motioncaptcha/}}
or a math equation~\cite{Leiva15_mucaptcha},
arguing that these symbols are language-independent.
Finally, BeCAPTCHA-Mouse~\cite{Acien20_becaptcha} analyzes mouse pointing behavior when the user clicks
on different CAPTCHA images.

To verify legitimate human presence on websites and online services,
a more fundamental and straightforward approach is the one we envision in this paper:
Just ask the user to handwrite any symbol
and verify (computationally) if a human actually produced it.
A key aspect of this kind of handwritten data is that it is of sequential nature,
therefore we can assume spatiotemporal sequences of $(x,y,t)$ tuples.
In this context, recent advancements on generative models have shown impressive results
on handwriting~\cite{graves2013generating}, sketching~\cite{Ha18},
and gesturing~\cite{Taranta16_gpsr}, but so far they have ignored the temporal information.
This relevant aspect was highlighted by Elarian et al.~\cite{Elarian14},
who concluded that on-line handwriting velocity is particularly difficult to simulate reliably,
specially for trajectories comprising several strokes~\cite{Acien20_becaptcha}.

One of the most successful techniques to generate handwriting data are movement simulation approaches,
which are mainly based on the human neuromotor control system and feed-forward models of locomotion.
To this aim, the oscillatory theory~\cite{hollerbach1981oscillation}
has been employed to generate handwriting as a result of horizontal and vertical oscillations (i.e., constrained modulation);
see e.g.~\cite{gangadhar2007oscillatory, choudhury2019synthesis}.
Another notable approach is the Kinematic Theory of rapid human movements~\cite{Plamondon95a, Plamondon95b}
and its associated Sigma-Lognormal (\slm) model~\cite{Plamondon06}.
According to this theory, \emph{aimed} human movements (i.e. ``movements with a purpose'')
are defined by elementary movement units
that are superimposed to produce the resulting trajectory~\cite{plamondon2014recent}.
The main idea is that the neuromuscular system involved in the production of an aimed movement
can be considered as a linear system made up of coupled subsystems,
and the impulse response of such a system converges toward a lognormal function.
The \slm model has been successfully used for synthesizing stroke gestures~\cite{Almaksour11, Leiva15_g3, Reznakova17}
and signatures~\cite{7775072, 8052226, 8585714},
among other types of human movements~\cite{MartinAlbo16_ipm, plamondon2014recent}
with high levels or realism. In addition, the \slm model is the only approach we are aware of
that can generate human-like \emph{generic} spatio-temporal handwriting sequences.
Therefore, we adopted this model to create synthetic data in this paper.

Previous work has analyzed the quality of synthetic data from several angles,
including e.g. input device and size of symbol vocabulary~\cite{Leiva15_g3},
and reproducing wrist movement and eye saccades~\cite{Plamondon95b}.
However, no previous work has attempted to differentiate human and synthetic on-line handwriting data from a computational modeling perspective.
This is an essential issue to prevent spoofing attacks, i.e.,
when a malicious attacker tries to defeat a biometric system
through the introduction of fake biometric samples.
In fact, it has been shown that it is feasible to use the \slm model
to attack biometric systems through the generation of synthetic signatures~\cite{8585714}.

\section{Method and Experiments}

Previous work analyzed the ``look and feel'' of \emph{off-line} machine-generated
handwritten symbols~\cite{Galbally12b, 8052226, Leiva17_slm, Leiva17_gestures} with the \slm model,
by conducting online surveys where users were shown one symbol \emph{image} at a time
and had to tell if it was human or machine-generated.
The results revealed that users were unable to tell them apart most of the time
(around 50\% of classification accuracy, which is essentially random performance).
In this paper, we implement several state-of-the-art convolutional neural nets
to see if a trained computer can do better than regular users.
We also implement several recurrent neural nets
that are able to handle both spatial and temporal information,
to see if incorporating movement dynamics contributes to improving classification performance.

\subsection{Handwriting Datasets}

\begin{figure*}[!ht]
  \centering
  \setlength\tabcolsep{2pt}

  \subfloat[\$1-GDS dataset examples\label{fig:data-1dollar}]{
    \def\imh{0.65cm}
    \begin{tabular}[b]{*6c}
      \multicolumn{4}{c}{\scriptsize Human} \\
      \midrule
      \includegraphics[height=\imh]{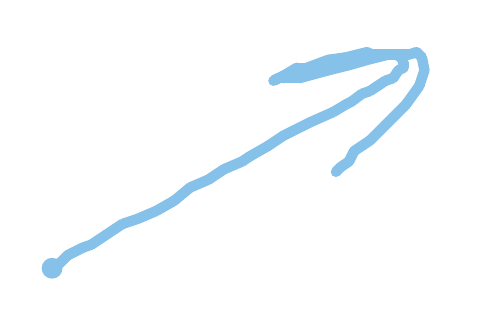}             &
      \includegraphics[height=\imh]{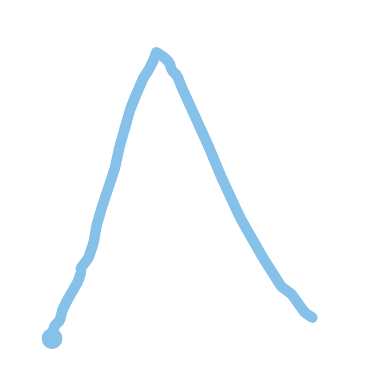}             &
      \includegraphics[height=\imh]{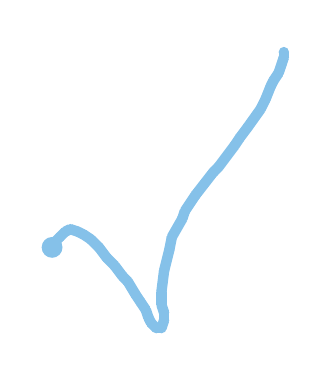}             &
      \includegraphics[height=\imh]{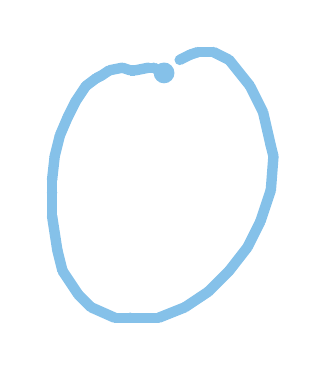}            \\
      \includegraphics[height=\imh]{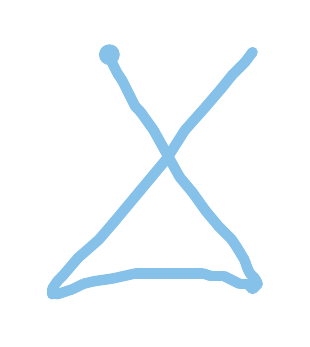}       &
      \includegraphics[height=\imh]{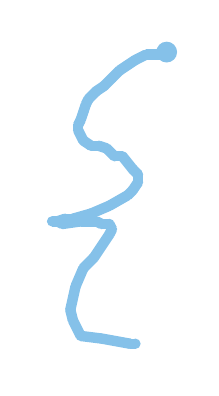}  &
      \includegraphics[height=\imh]{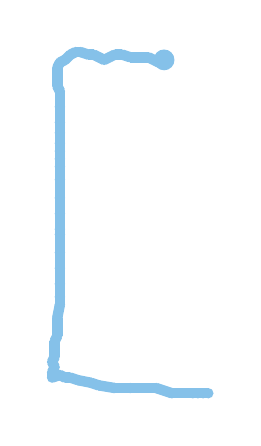}   &
      \includegraphics[height=\imh]{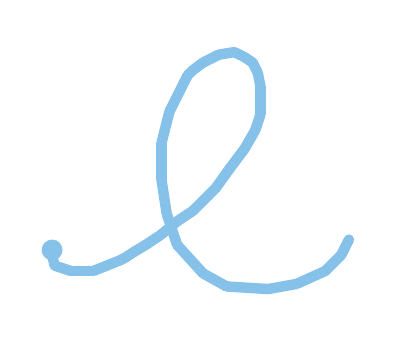}           \\
      \includegraphics[height=\imh]{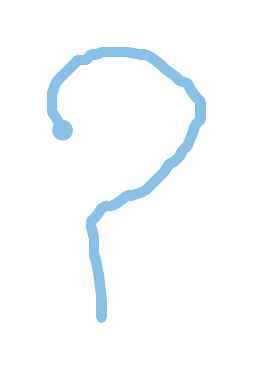}     &
      \includegraphics[height=\imh]{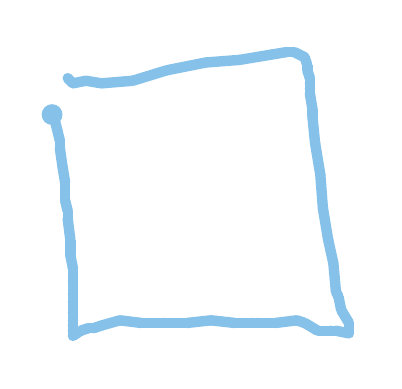}         &
      \includegraphics[height=\imh]{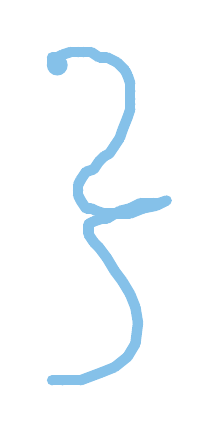} &
      \includegraphics[height=\imh]{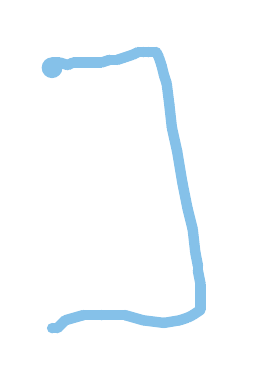}  \\
      \includegraphics[height=\imh]{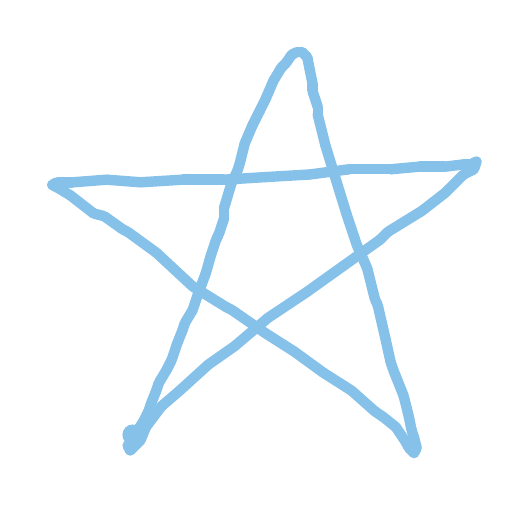}              &
      \includegraphics[height=\imh]{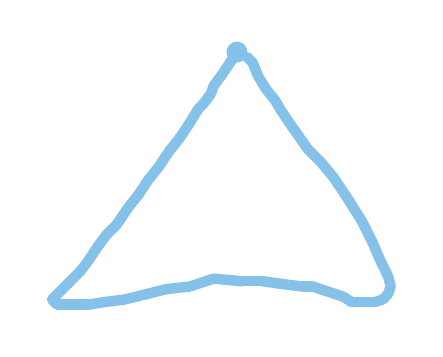}          &
      \includegraphics[height=\imh]{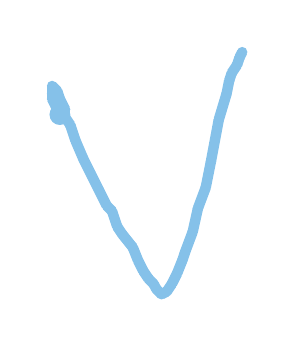}                 &
      \includegraphics[height=\imh]{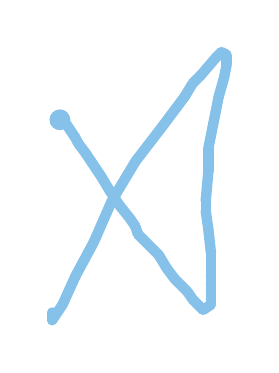}                 \\
      \midrule
    \end{tabular}
    \hspace{0.25em}
    \begin{tabular}[b]{*6c}
      \multicolumn{4}{c}{\scriptsize Machine} \\
      \midrule
      \includegraphics[height=\imh]{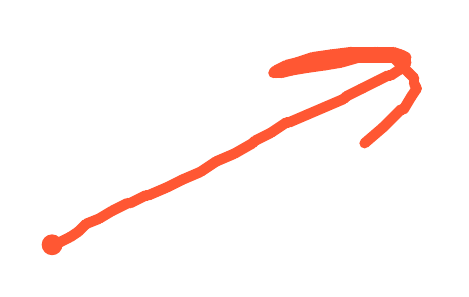}             &
      \includegraphics[height=\imh]{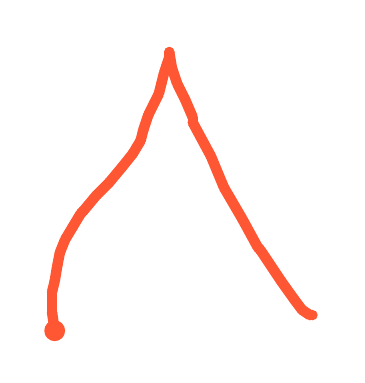}             &
      \includegraphics[height=\imh]{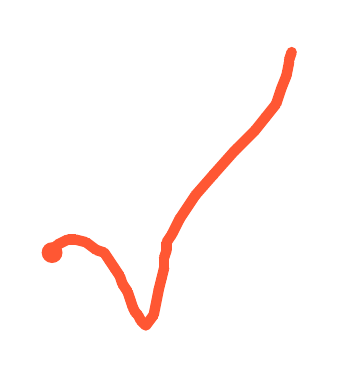}             &
      \includegraphics[height=\imh]{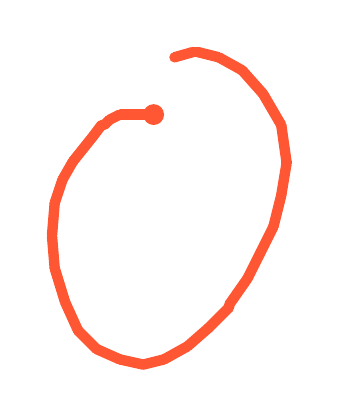}            \\
      \includegraphics[height=\imh]{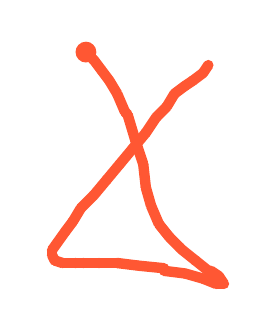}       &
      \includegraphics[height=\imh]{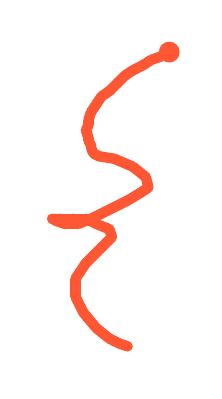}  &
      \includegraphics[height=\imh]{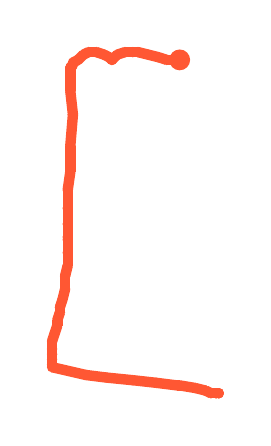}   &
      \includegraphics[height=\imh]{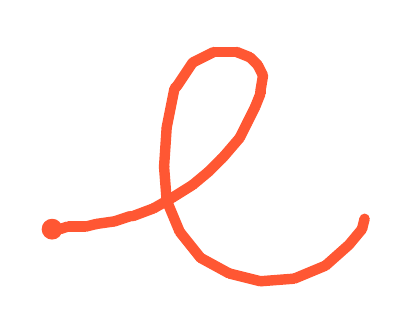}           \\
      \includegraphics[height=\imh]{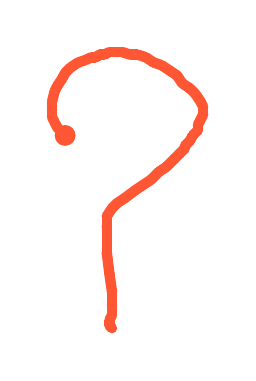}     &
      \includegraphics[height=\imh]{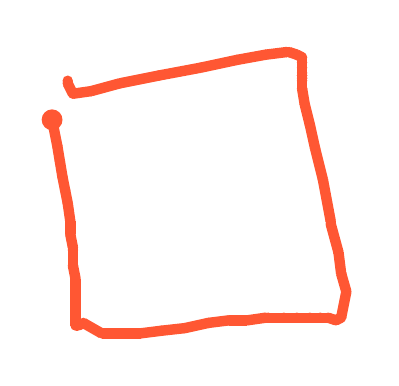}         &
      \includegraphics[height=\imh]{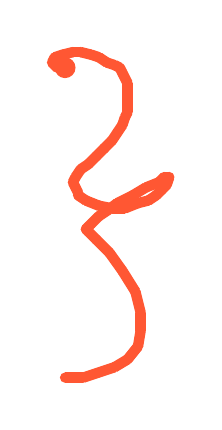} &
      \includegraphics[height=\imh]{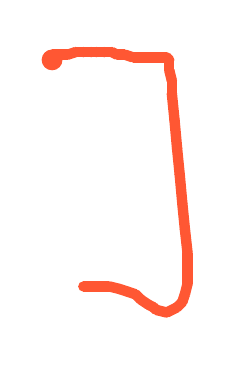}  \\
      \includegraphics[height=\imh]{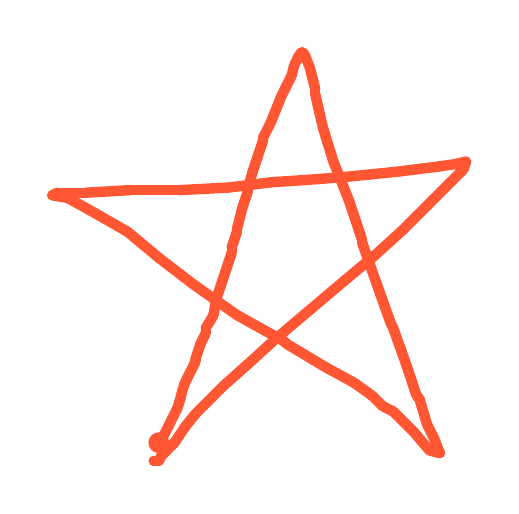}              &
      \includegraphics[height=\imh]{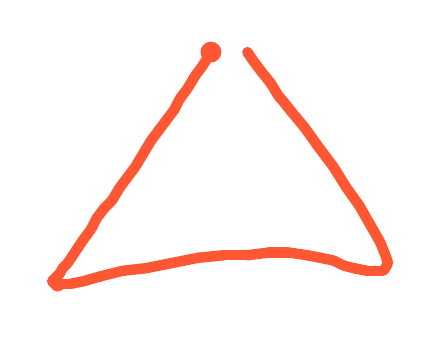}          &
      \includegraphics[height=\imh]{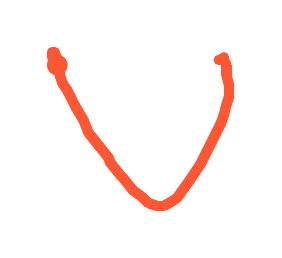}                 &
      \includegraphics[height=\imh]{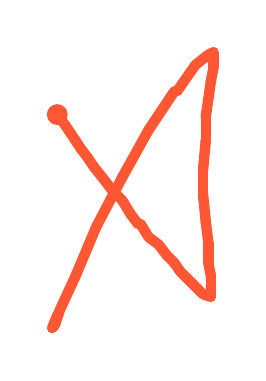}                 \\
      \midrule
    \end{tabular}
  }
  \hfill
  \subfloat[\$N-MMG dataset examples\label{fig:data-ndollar}]{
    \def\imh{0.65cm}
    \begin{tabular}[b]{*6c}
      \multicolumn{4}{c}{\scriptsize Human} \\
      \midrule
      \includegraphics[height=\imh]{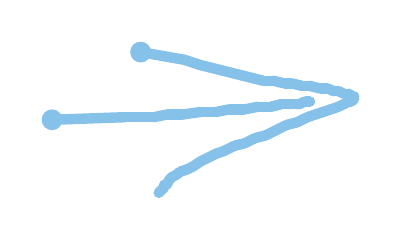}         &
      \includegraphics[height=\imh]{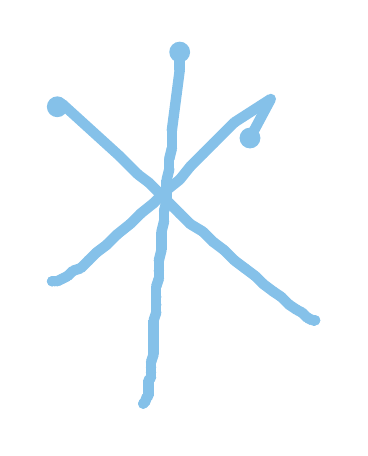}          &
      \includegraphics[height=\imh]{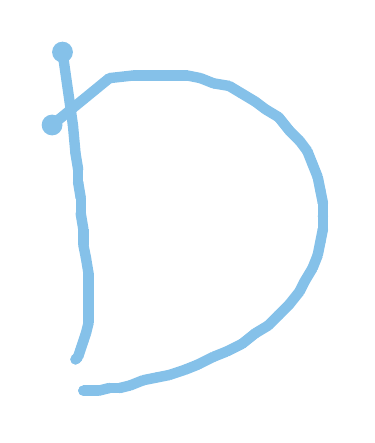}                 &
      \includegraphics[height=\imh]{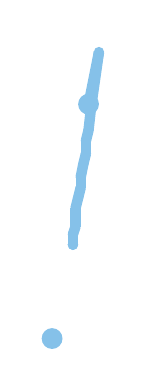} \\
      \includegraphics[height=\imh]{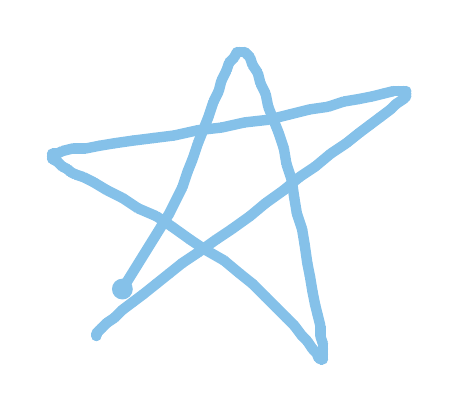}   &
      \includegraphics[height=\imh]{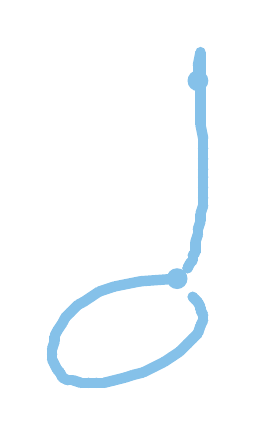}         &
      \includegraphics[height=\imh]{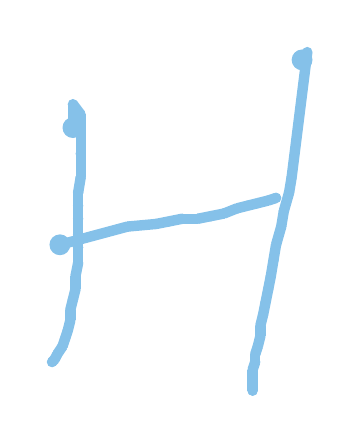}                 &
      \includegraphics[height=\imh]{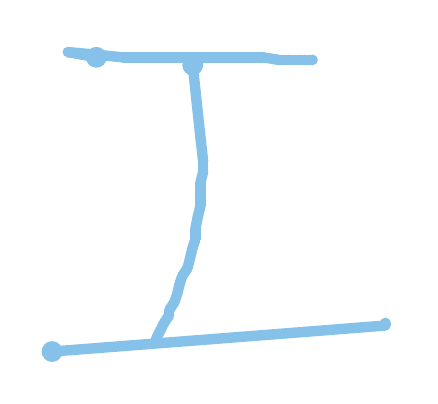}                 \\
      \includegraphics[width=\imh]{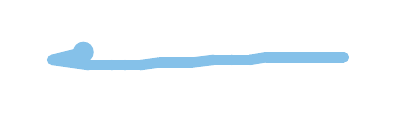}               &
      \includegraphics[height=\imh]{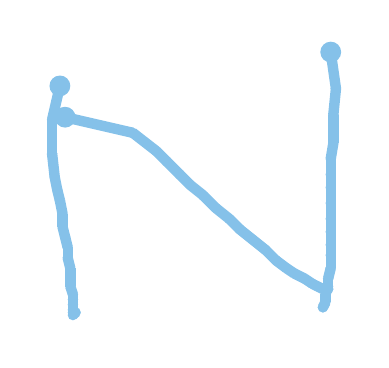}                 &
      \includegraphics[height=\imh]{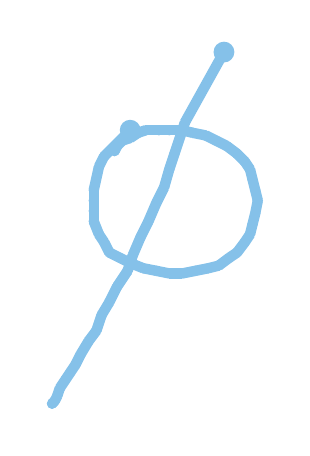}              &
      \includegraphics[height=\imh]{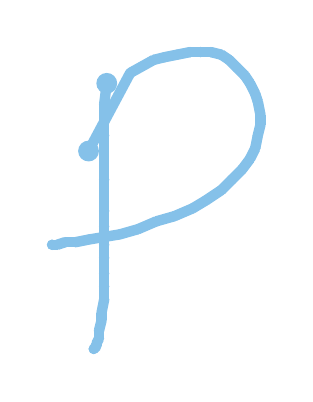}                 \\
      \includegraphics[height=\imh]{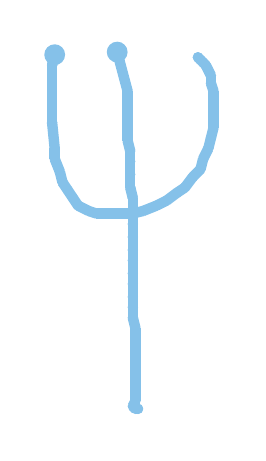}         &
      \includegraphics[height=\imh]{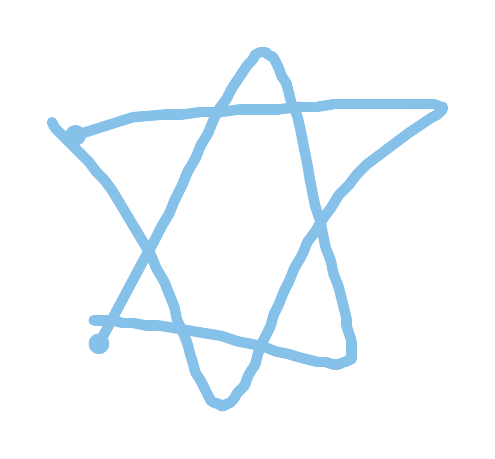}    &
      \includegraphics[height=\imh]{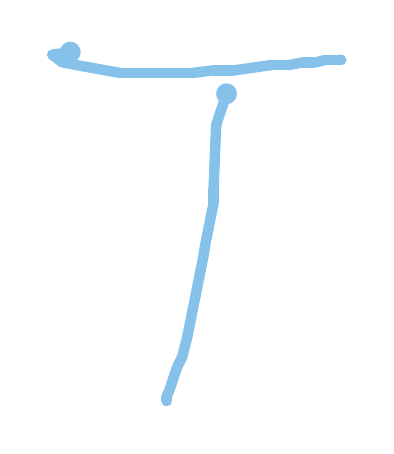}                 &
      \includegraphics[height=\imh]{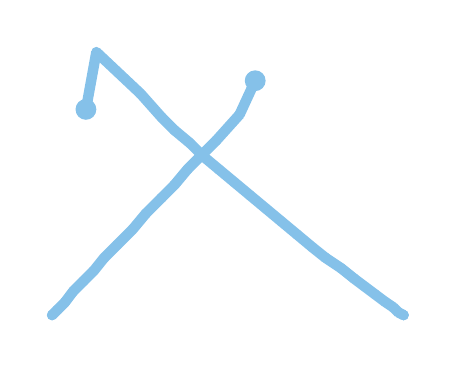}                 \\
      \midrule
    \end{tabular}
    \hspace{0.25em}
    \begin{tabular}[b]{*6c}
      \multicolumn{4}{c}{\scriptsize Machine} \\
      \midrule
      \includegraphics[height=\imh]{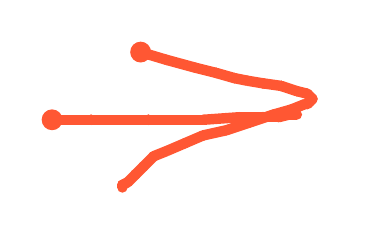}         &
      \includegraphics[height=\imh]{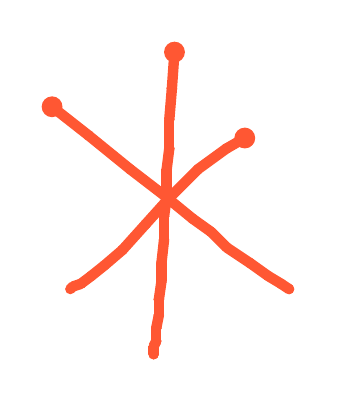}          &
      \includegraphics[height=\imh]{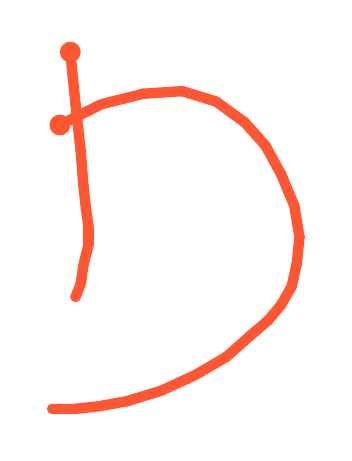}                 &
      \includegraphics[height=\imh]{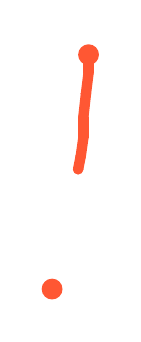} \\
      \includegraphics[height=\imh]{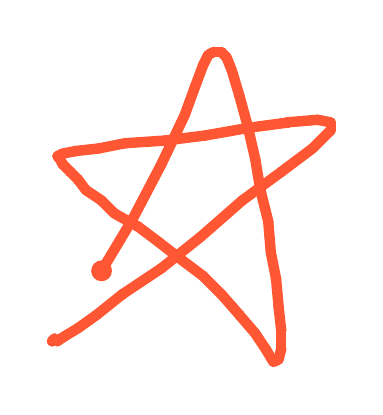}   &
      \includegraphics[height=\imh]{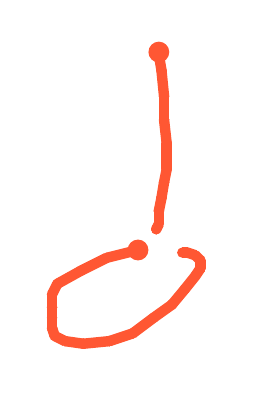}         &
      \includegraphics[height=\imh]{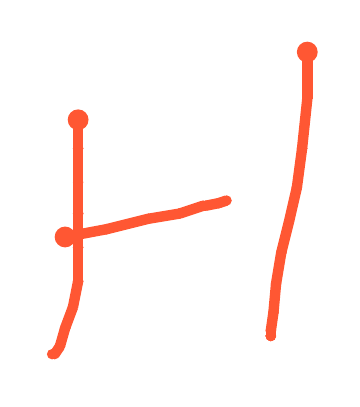}                 &
      \includegraphics[height=\imh]{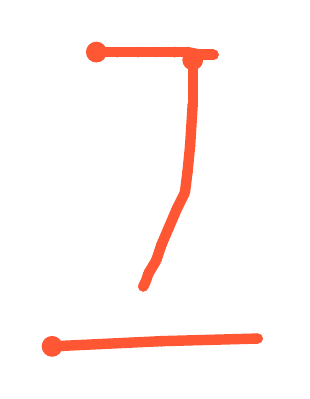}                 \\
      \includegraphics[width=\imh]{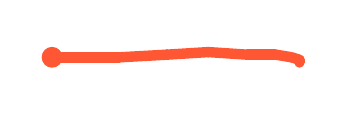}               &
      \includegraphics[height=\imh]{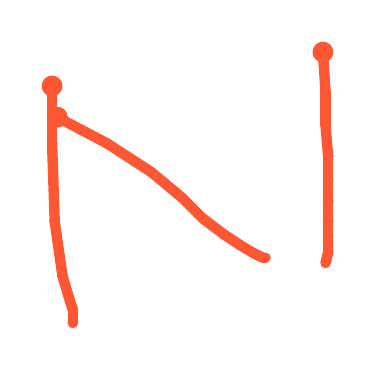}                 &
      \includegraphics[height=\imh]{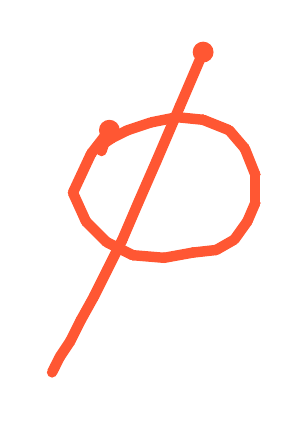}              &
      \includegraphics[height=\imh]{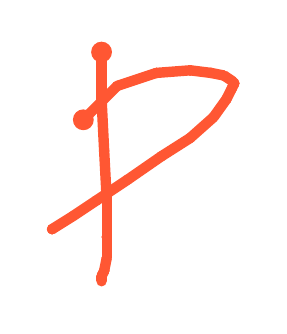}                 \\
      \includegraphics[height=\imh]{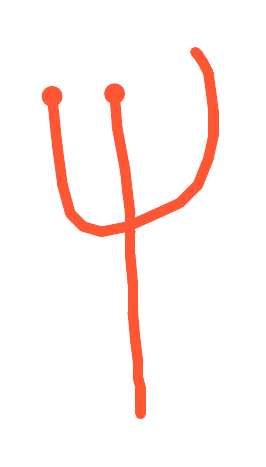}         &
      \includegraphics[height=\imh]{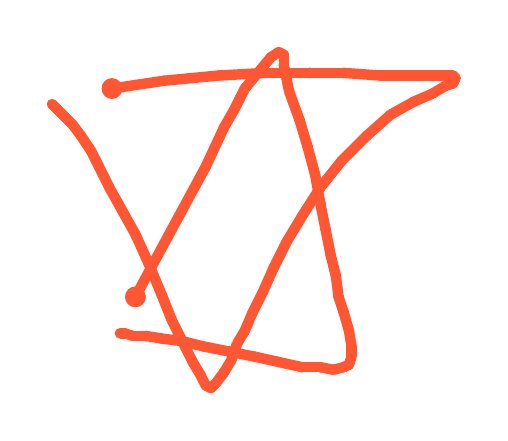}    &
      \includegraphics[height=\imh]{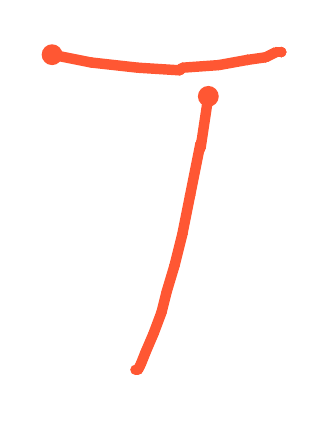}                 &
      \includegraphics[height=\imh]{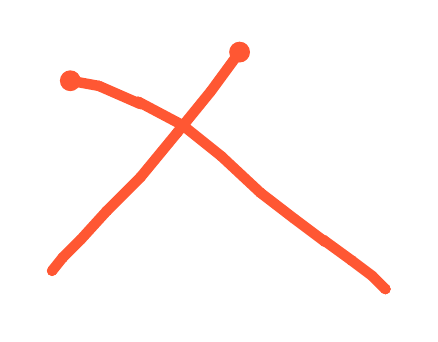}                 \\
      \midrule
    \end{tabular}
  }
  \hfill
  \subfloat[Chars74k dataset examples\label{fig:data-chars}]{
    \def\imh{0.6cm}
    \begin{tabular}[b]{*6c}
      \multicolumn{4}{c}{\scriptsize Human} \\
      \midrule
      \rotae{\includegraphics[width=\imh]{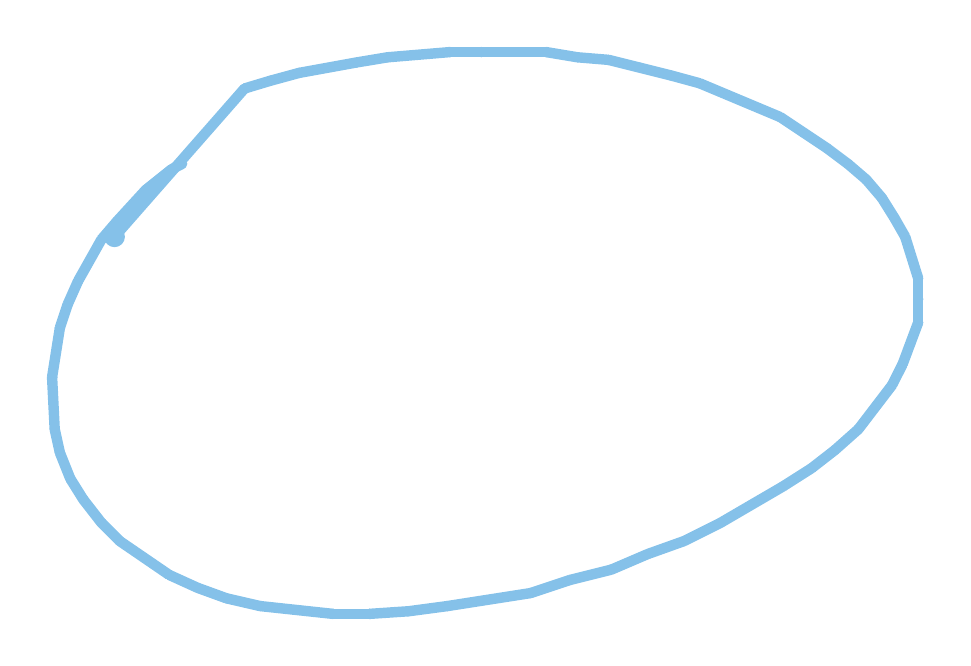}}  &
      \rotae{\includegraphics[height=\imh]{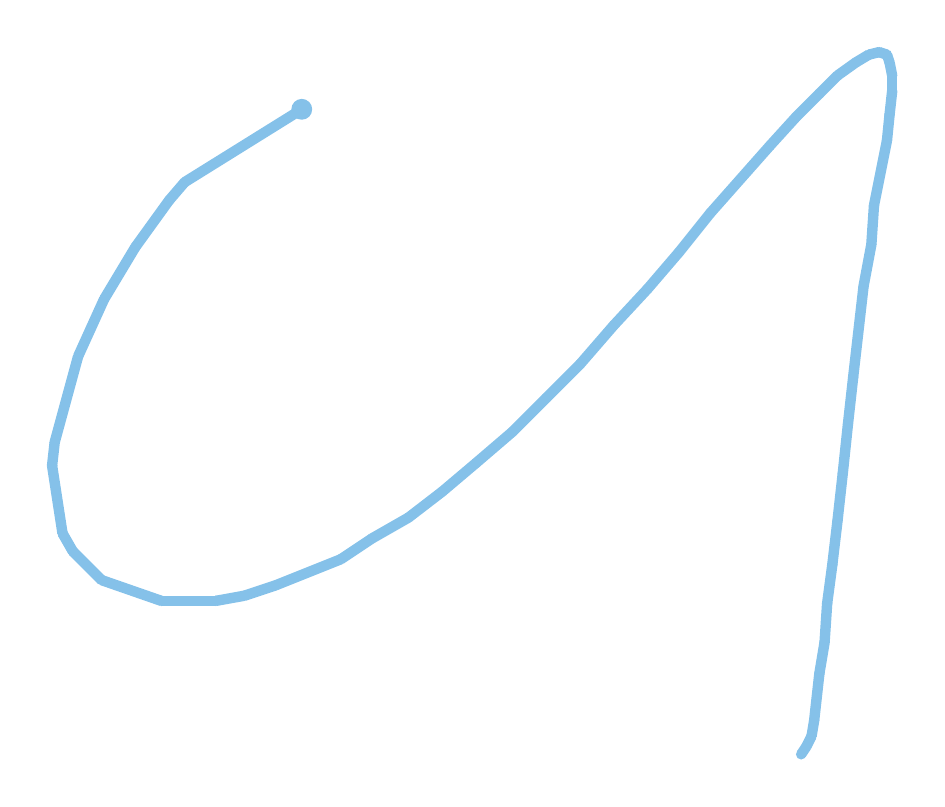}}  &
      \rotae{\includegraphics[width=\imh]{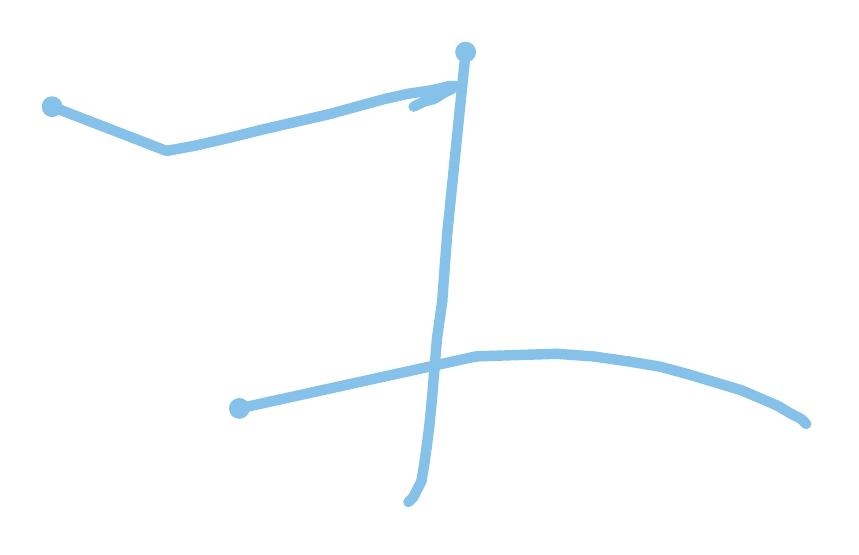}}  &
      \rotae{\includegraphics[width=\imh]{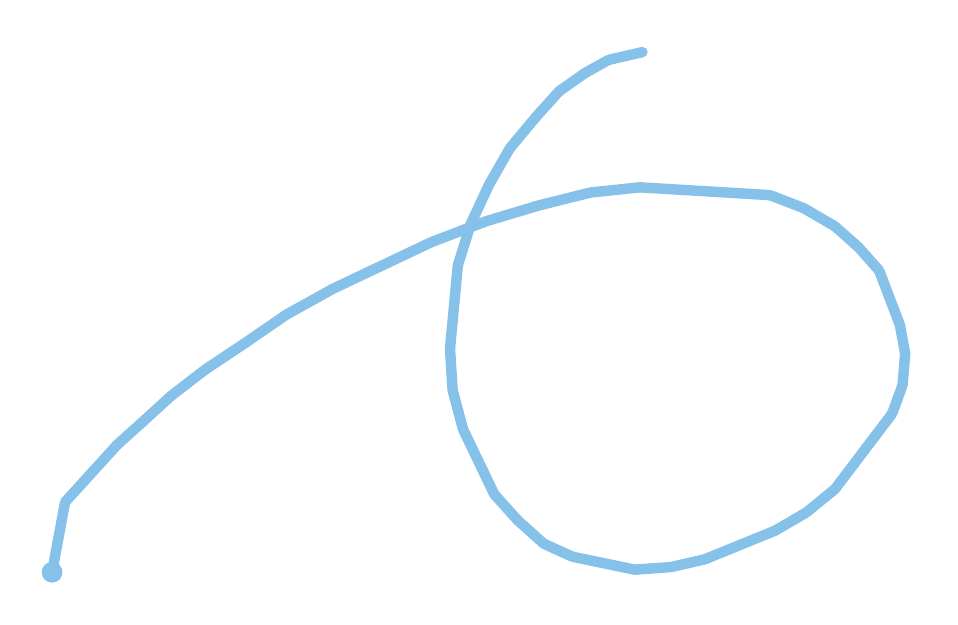}}  \\
      \rotae{\includegraphics[width=\imh]{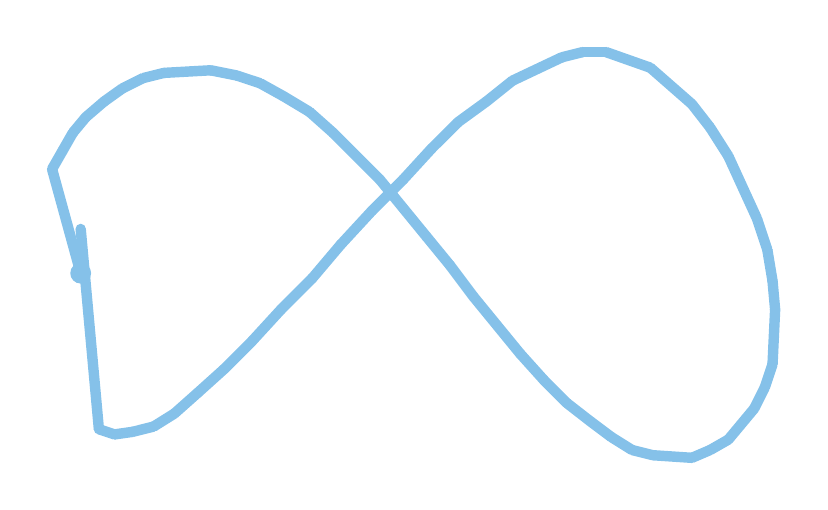}}  &
      \rotae{\includegraphics[height=\imh]{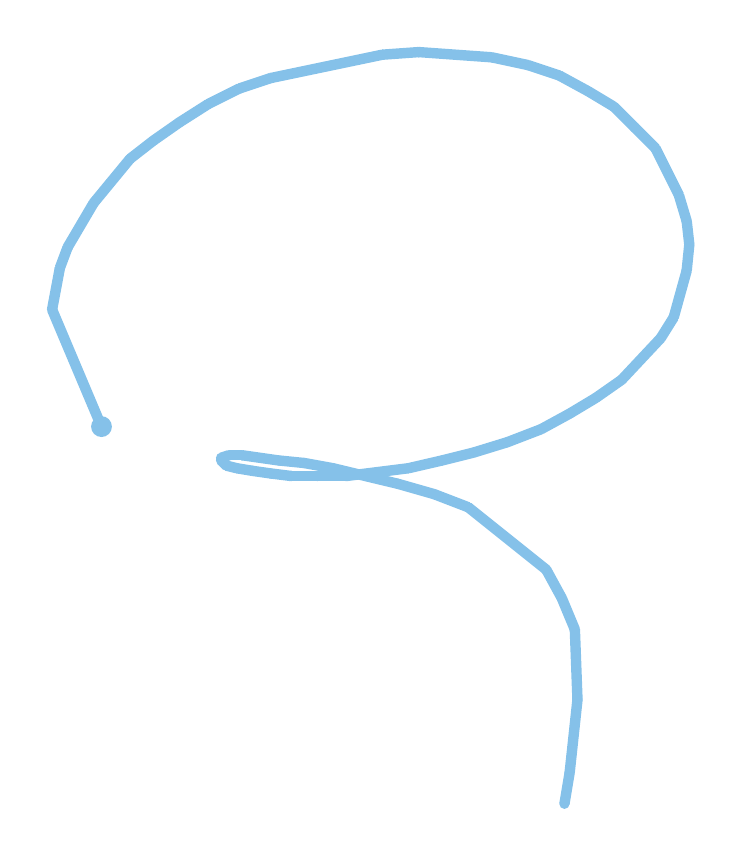}}  &
      \rotae{\includegraphics[height=\imh]{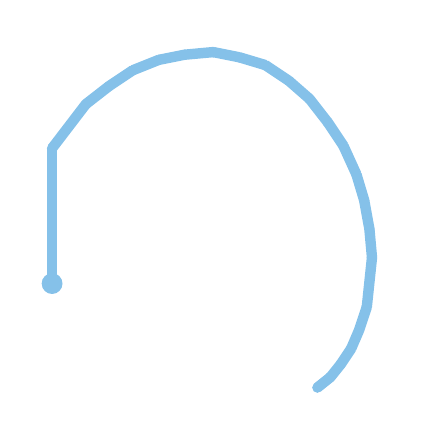}}  &
      \rotae{\includegraphics[height=\imh]{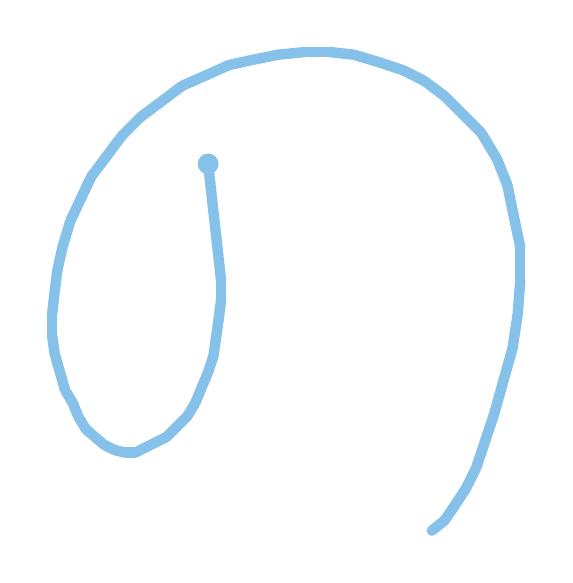}}  \\
      \rotae{\includegraphics[width=\imh]{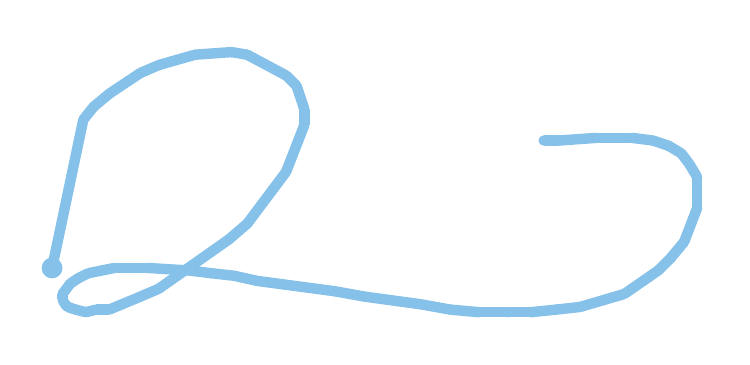}}  &
      \rotae{\includegraphics[height=\imh]{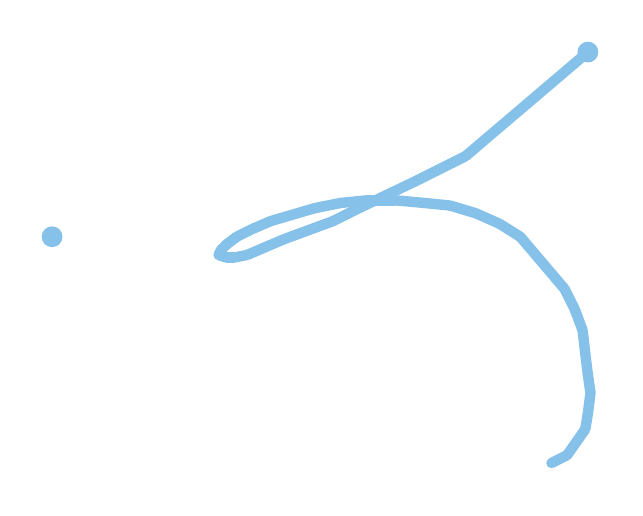}}  &
      \rotae{\includegraphics[width=\imh]{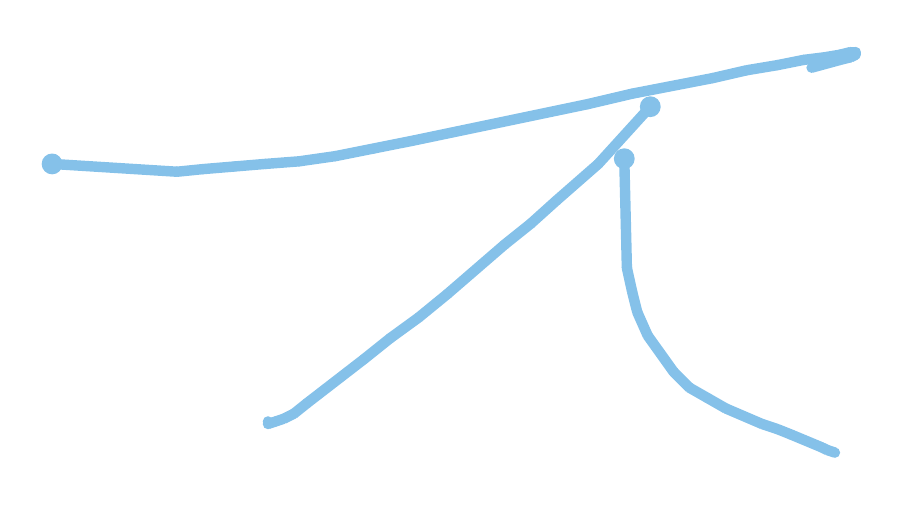}}  &
      \rotae{\includegraphics[height=\imh]{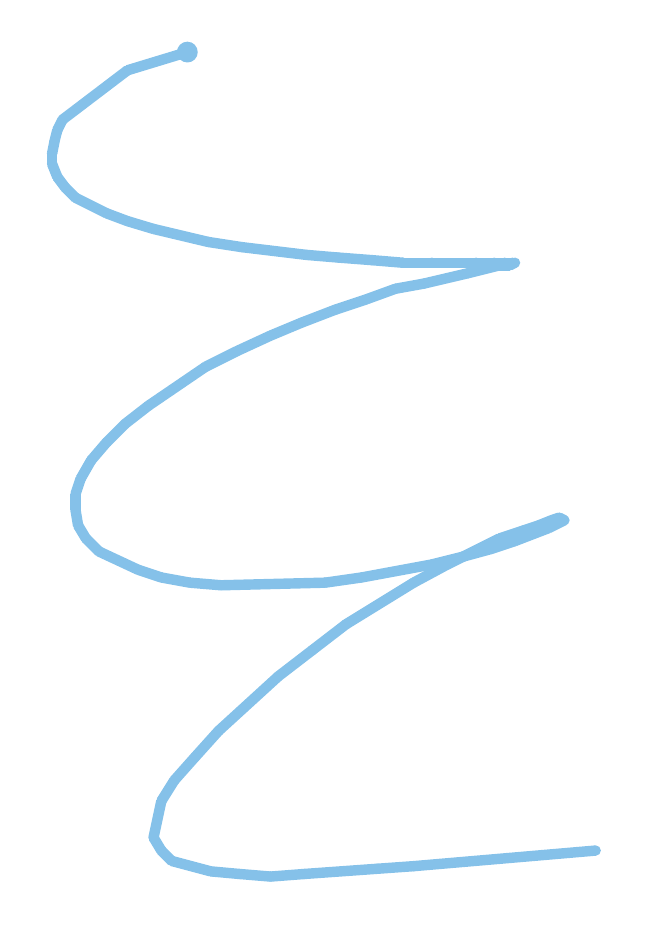}}  \\
      \rotae{\includegraphics[height=\imh]{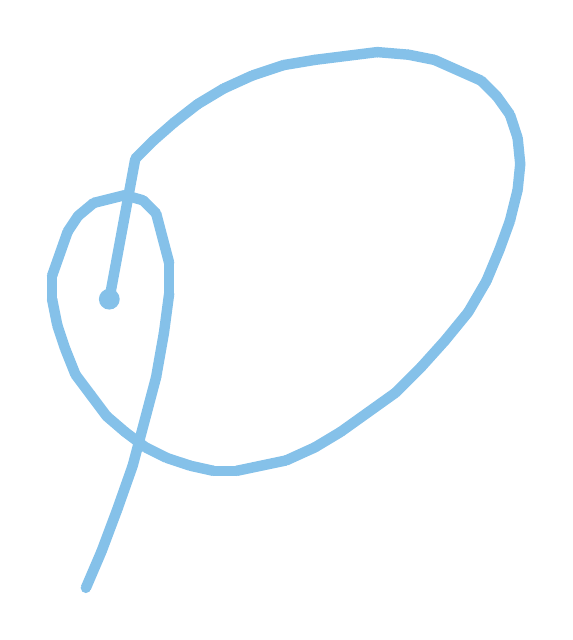}}  &
      \rotae{\includegraphics[width=\imh]{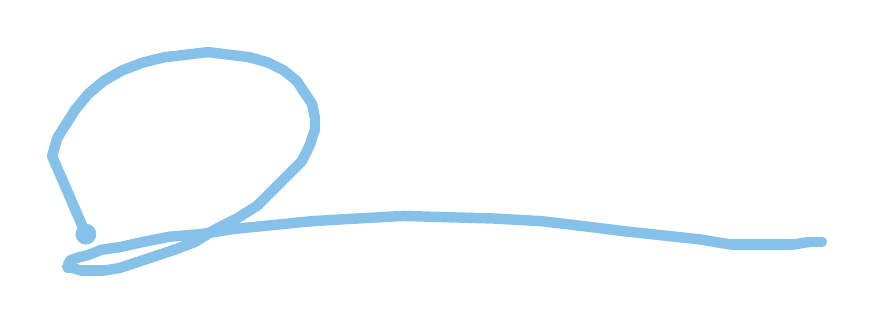}}   &
      \rotae{\includegraphics[height=\imh]{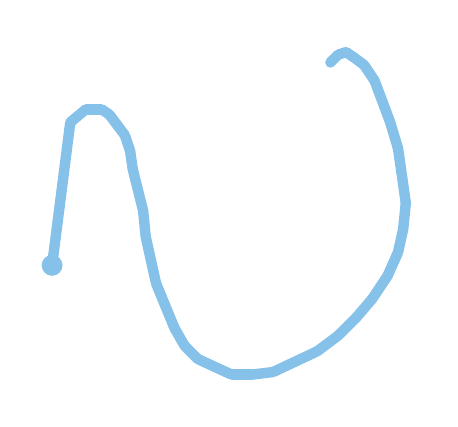}}  &
      \rotae{\includegraphics[height=\imh]{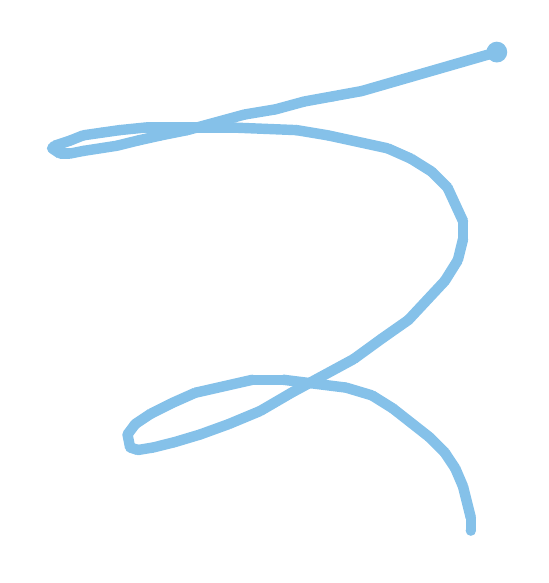}}  \\
      \midrule
    \end{tabular}
    \hspace{0.25em}
    \begin{tabular}[b]{*6c}
      \multicolumn{4}{c}{\scriptsize Machine} \\
      \midrule
      \rotae{\includegraphics[height=\imh]{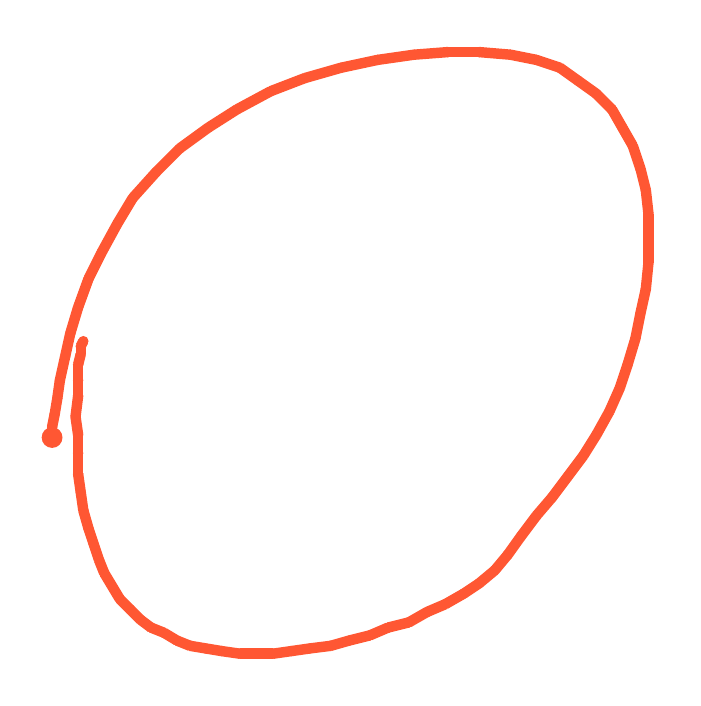}}  &
      \rotae{\includegraphics[height=\imh]{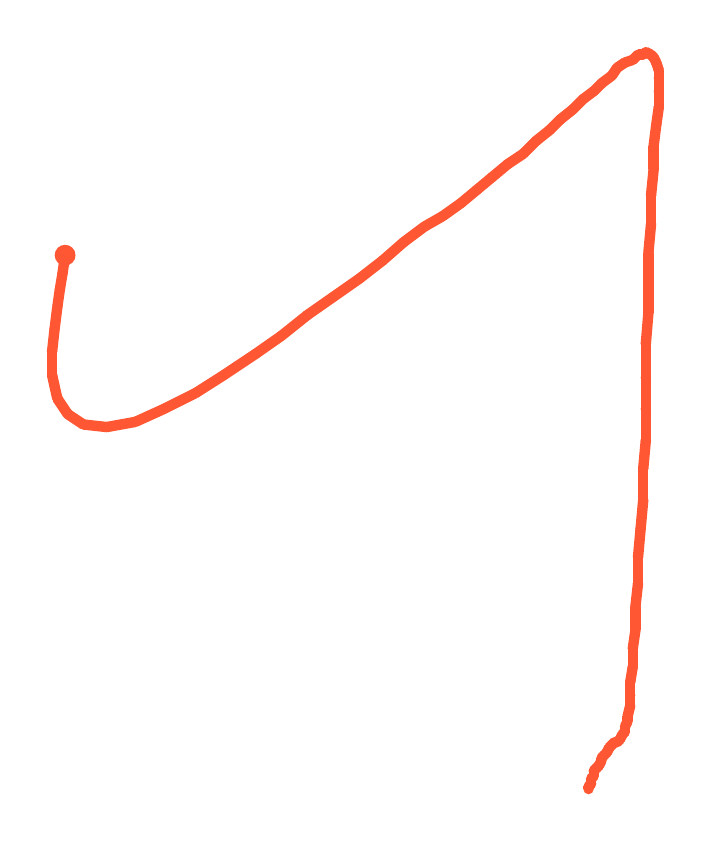}}  &
      \rotae{\includegraphics[height=\imh]{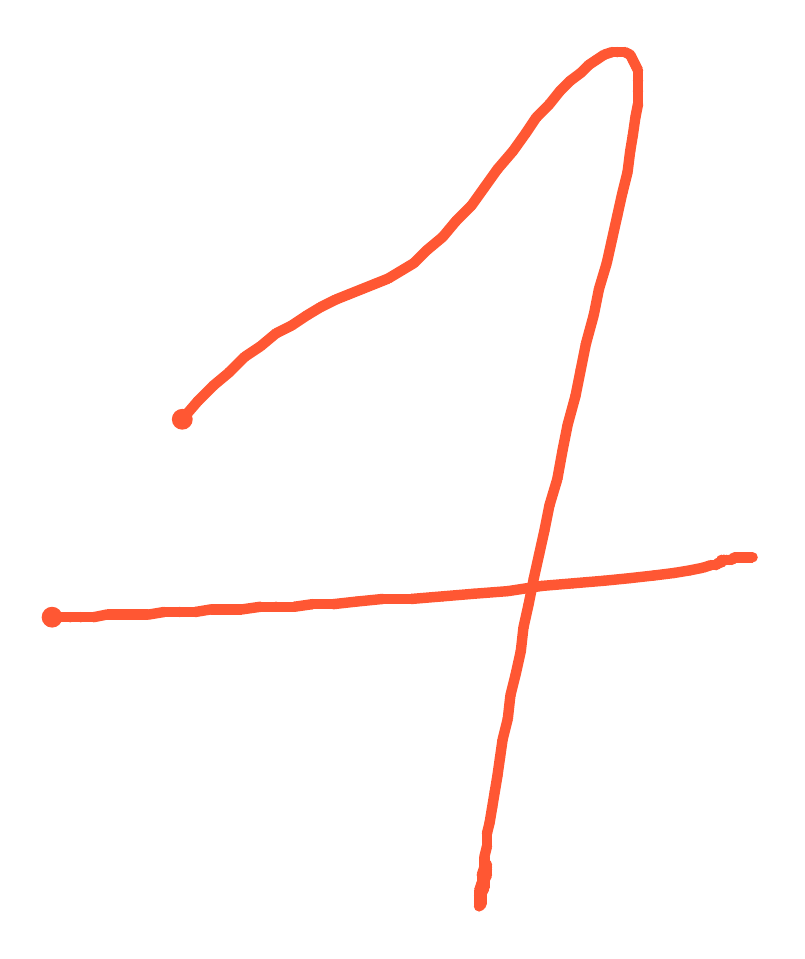}}  &
      \rotae{\includegraphics[width=\imh]{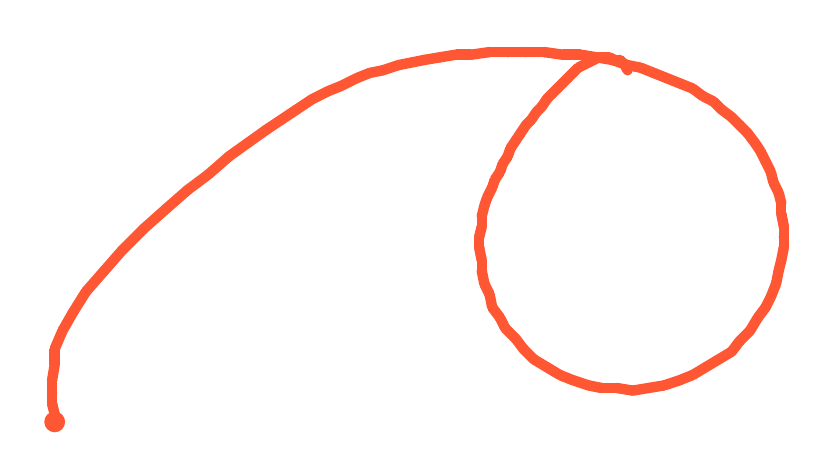}}  \\
      \rotae{\includegraphics[height=\imh]{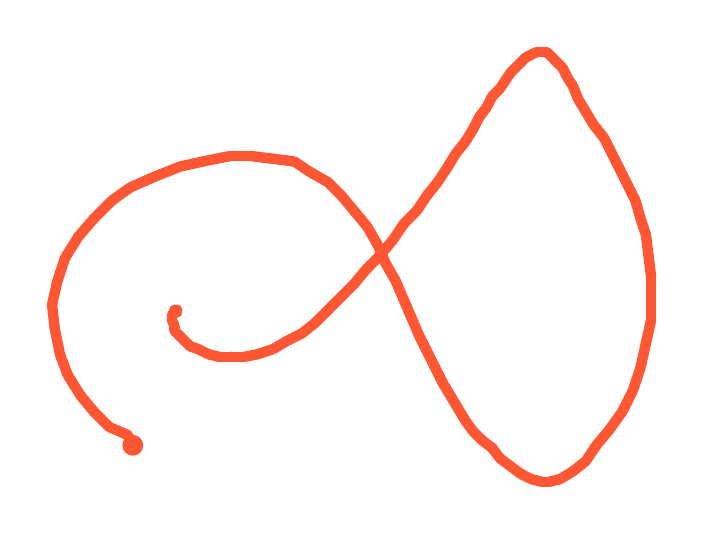}}  &
      \rotae{\includegraphics[height=\imh]{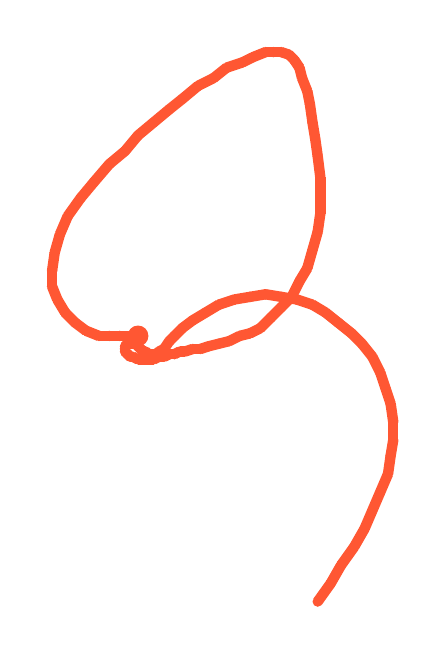}}  &
      \rotae{\includegraphics[height=\imh]{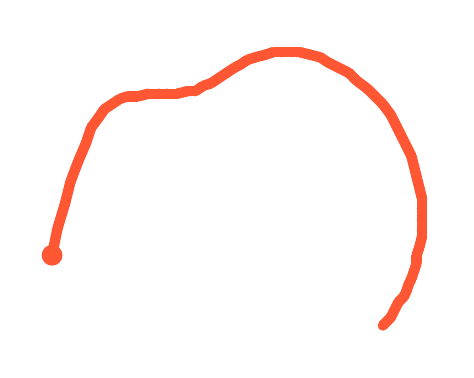}}  &
      \rotae{\includegraphics[height=\imh]{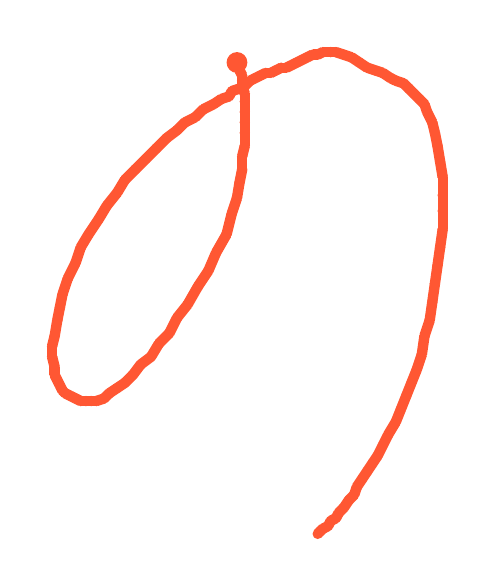}}  \\
      \rotae{\includegraphics[width=\imh]{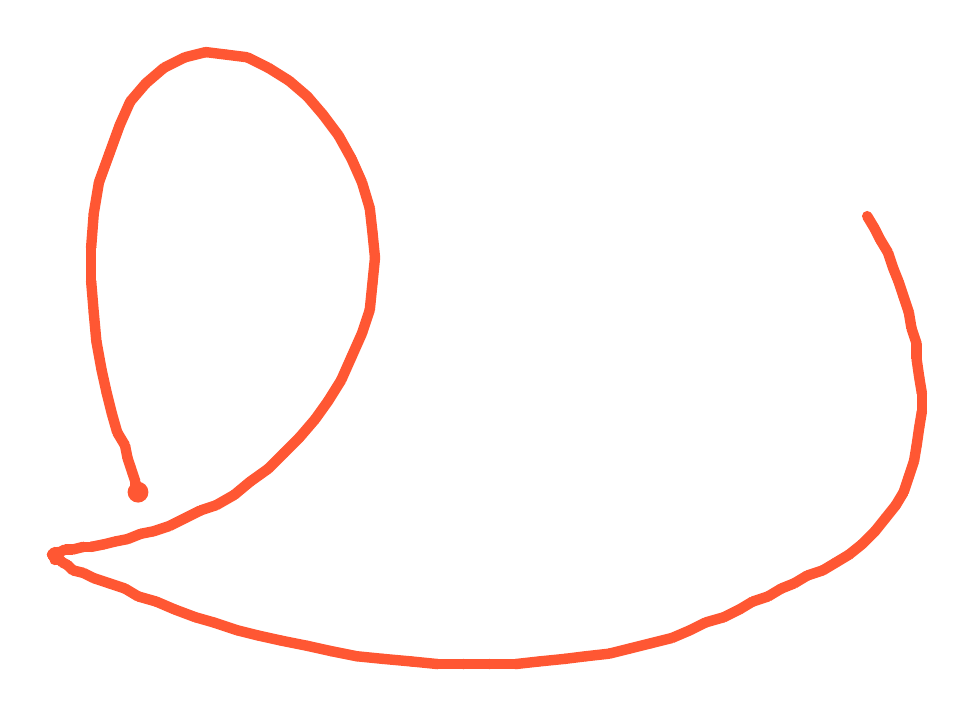}}  &
      \rotae{\includegraphics[width=\imh]{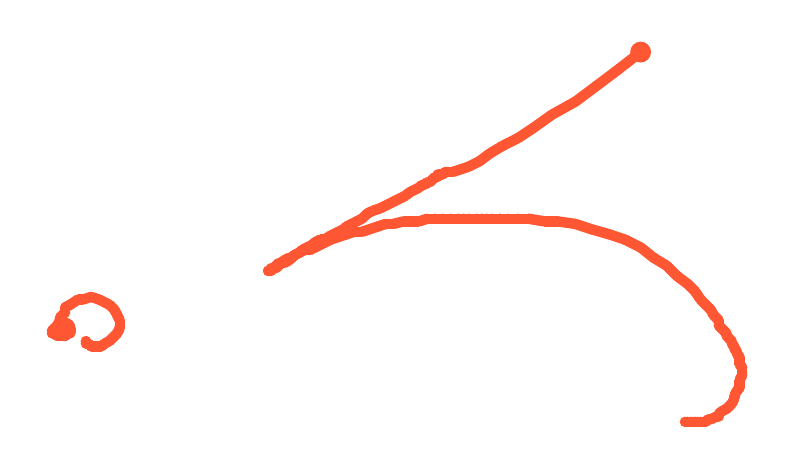}}  &
      \rotae{\includegraphics[height=\imh]{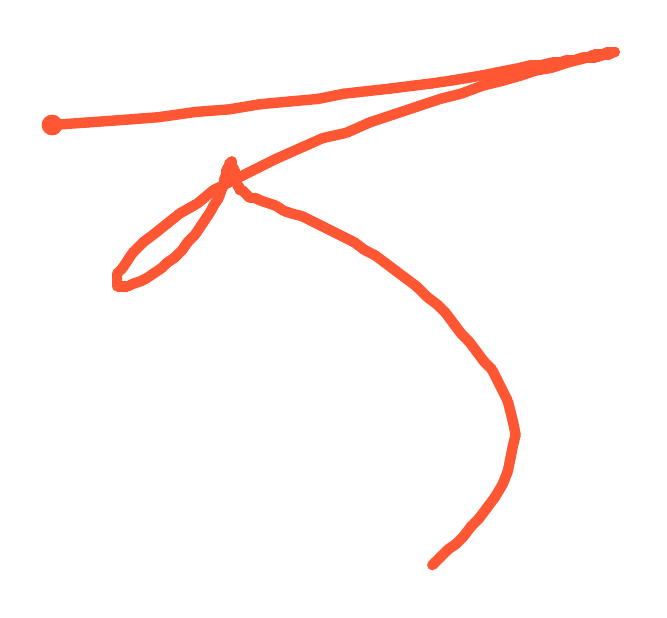}}  &
      \rotae{\includegraphics[height=\imh]{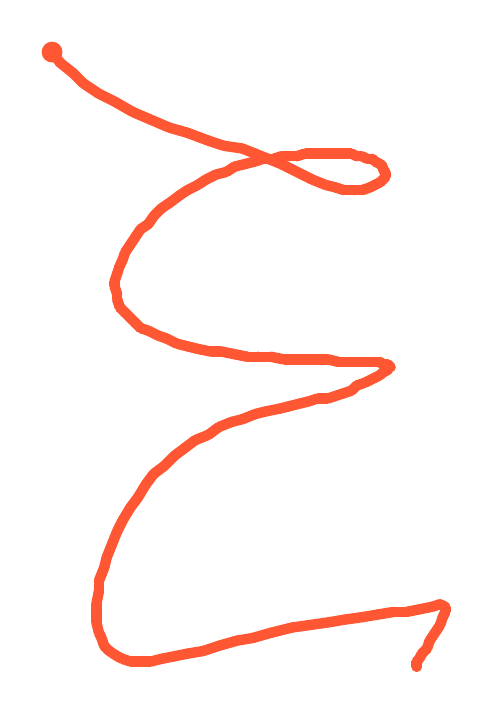}}  \\
      \rotae{\includegraphics[height=\imh]{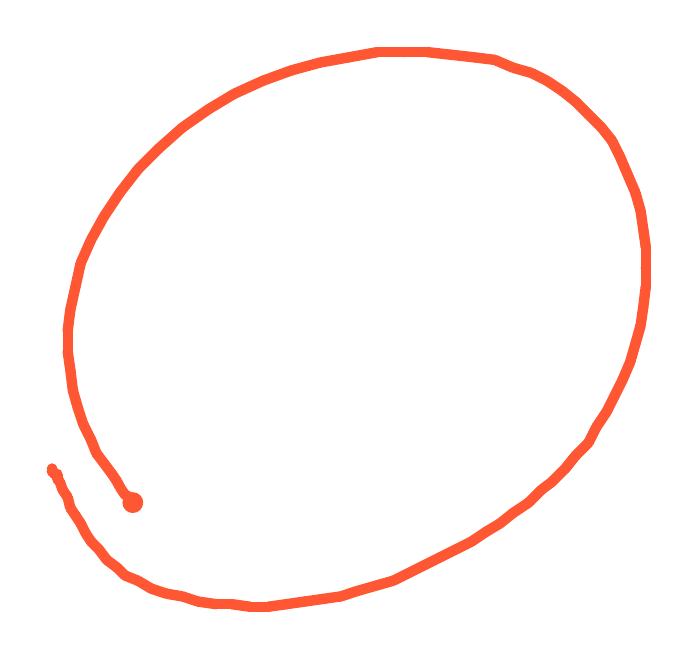}}  &
      \rotae{\includegraphics[width=\imh]{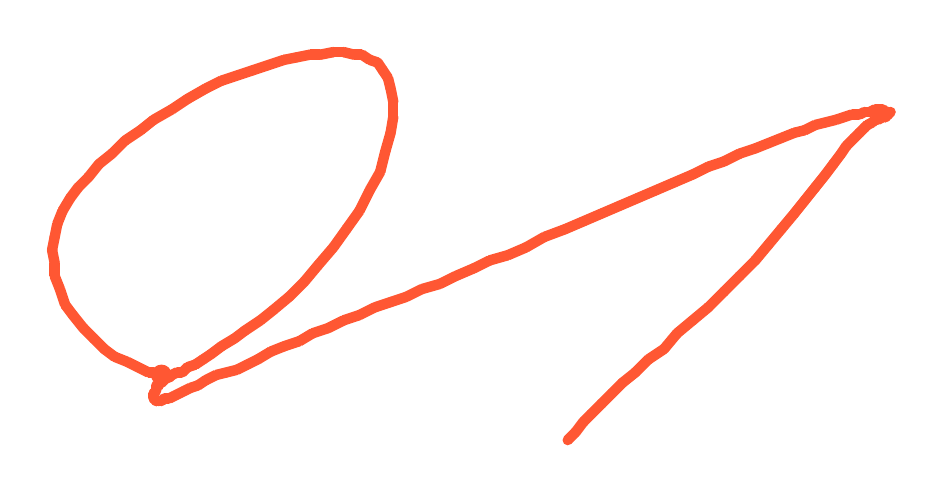}}  &
      \rotae{\includegraphics[height=\imh]{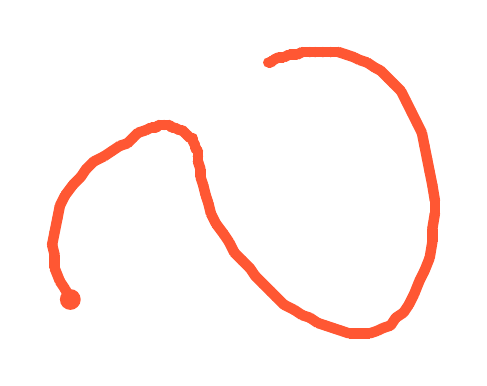}}  &
      \rotae{\includegraphics[height=\imh]{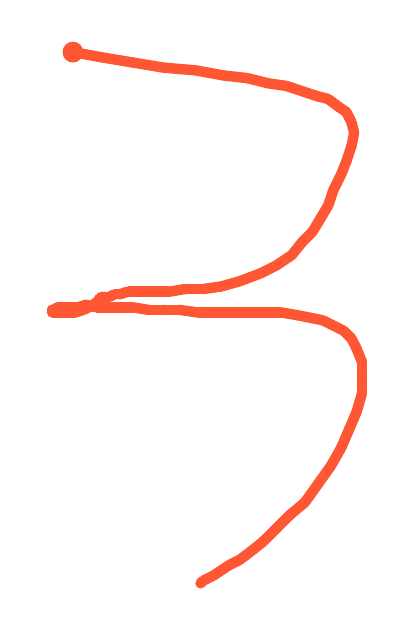}}  \\
      \midrule
    \end{tabular}
  }
  \hfill
  \subfloat[SUSIGv dataset examples\label{fig:data-susig}]{
    \def\imh{0.5cm}
    \begin{tabular}[b]{*6c}
      \multicolumn{4}{c}{\scriptsize Human} \\
      \midrule
      \fliph{\includegraphics[height=\imh]{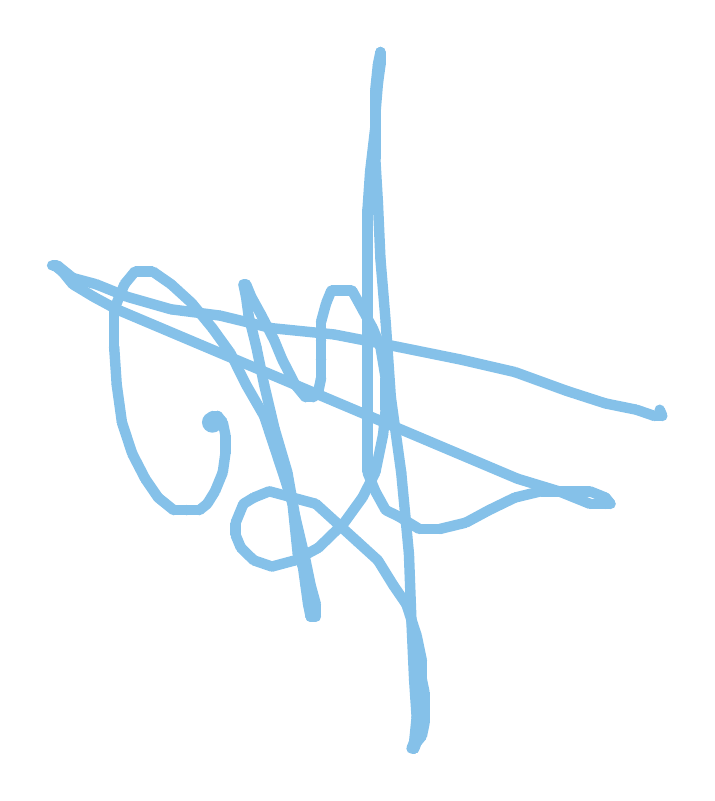}} &
      \fliph{\includegraphics[height=\imh]{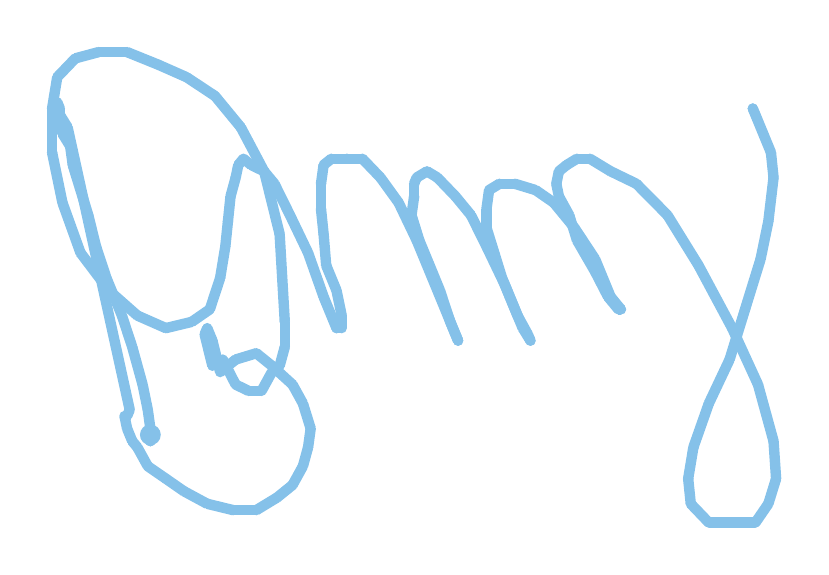}} &
      \fliph{\includegraphics[height=\imh]{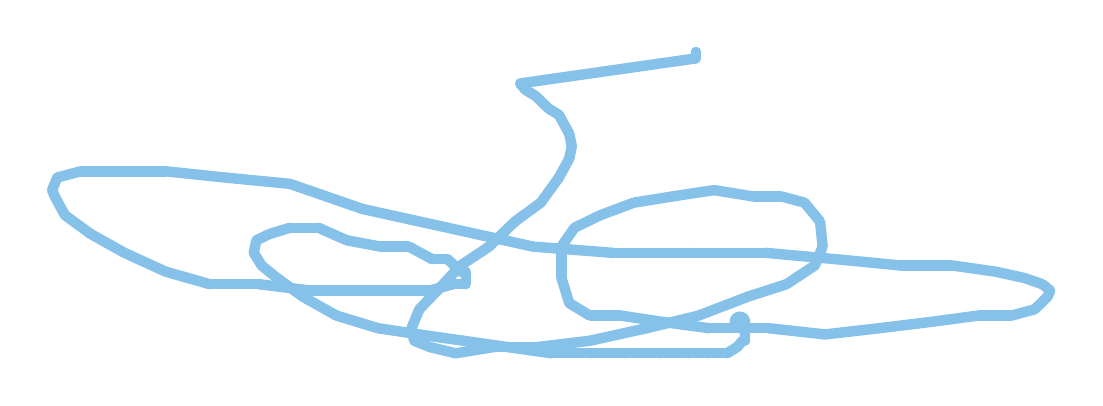}} &
      \fliph{\includegraphics[height=\imh]{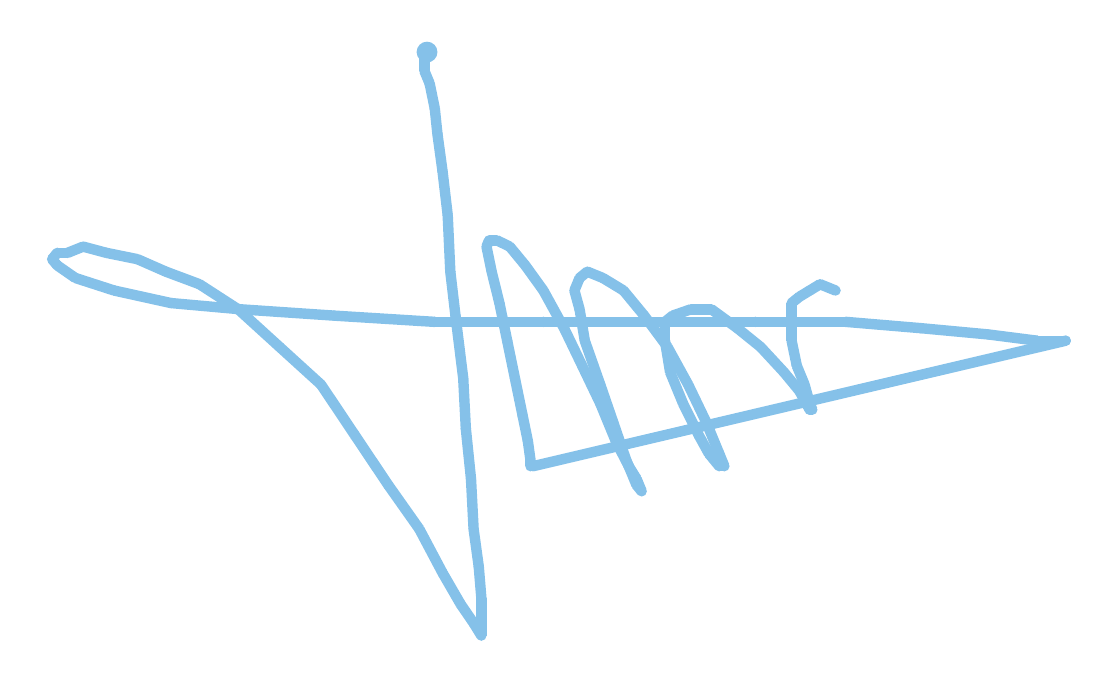}} \\
      \fliph{\includegraphics[height=\imh]{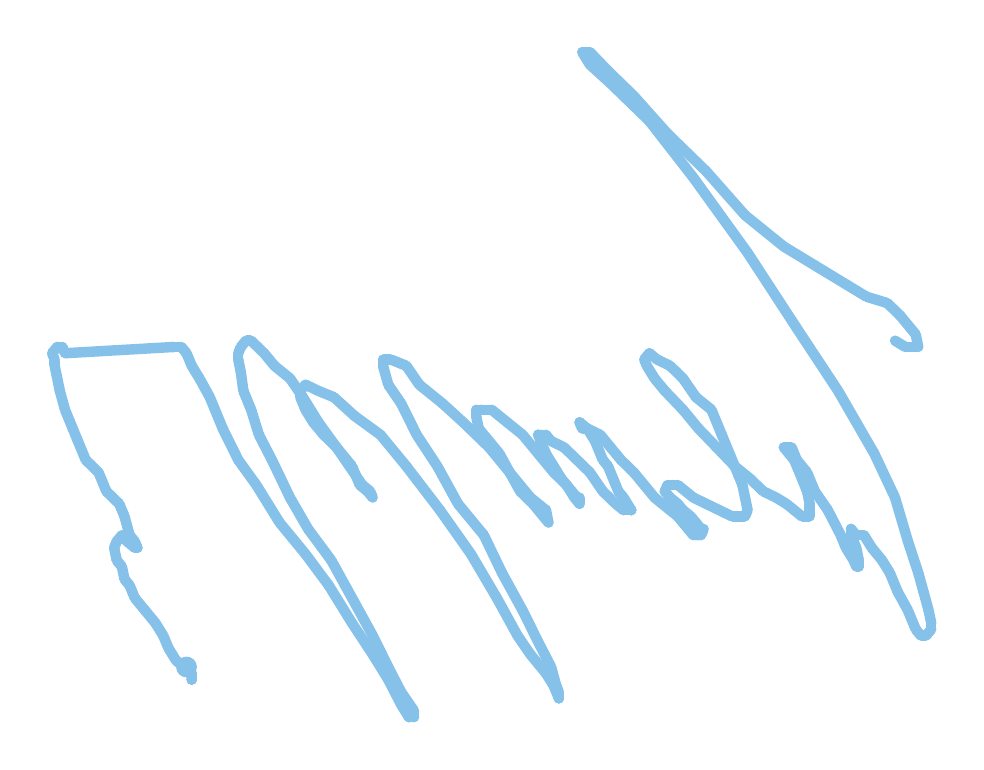}} &
      \fliph{\includegraphics[height=\imh]{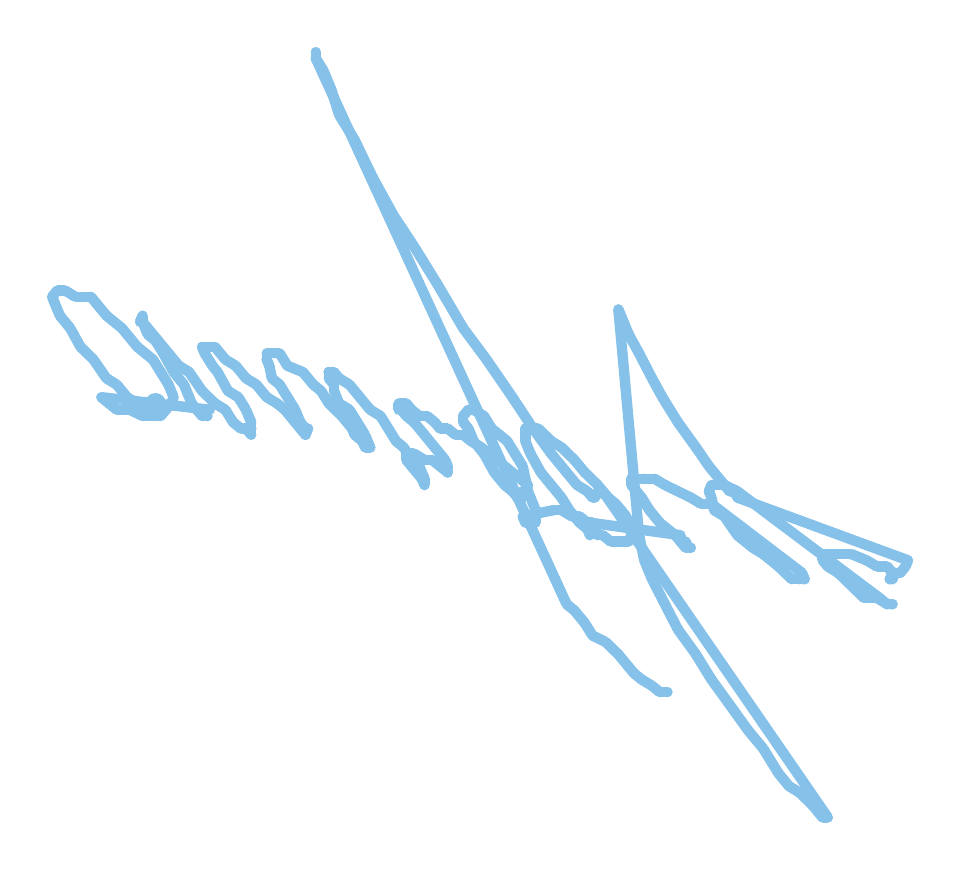}} &
      \fliph{\includegraphics[height=\imh]{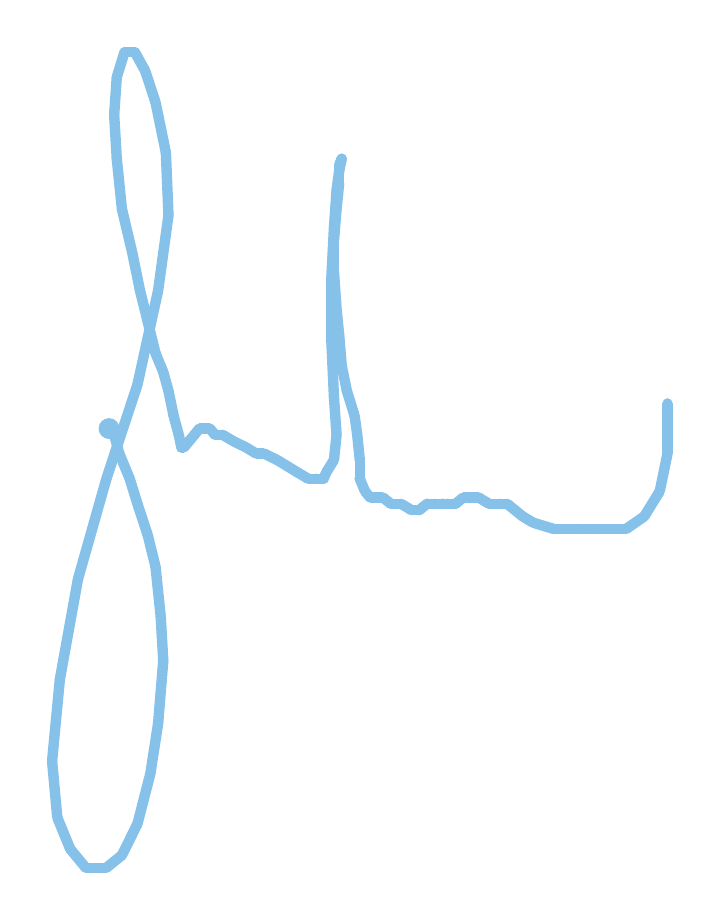}} &
      \fliph{\includegraphics[height=\imh]{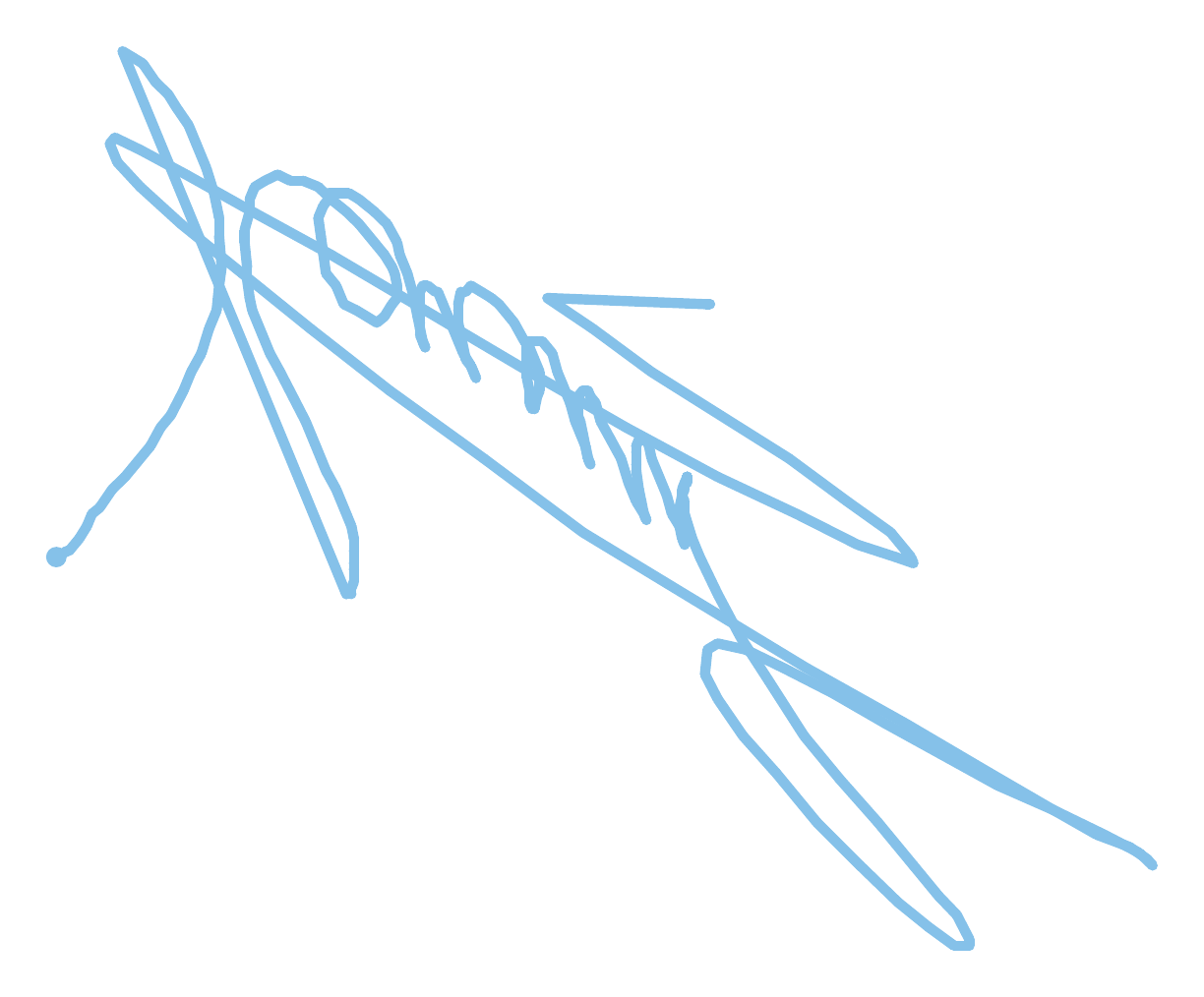}} \\
      \fliph{\includegraphics[height=\imh]{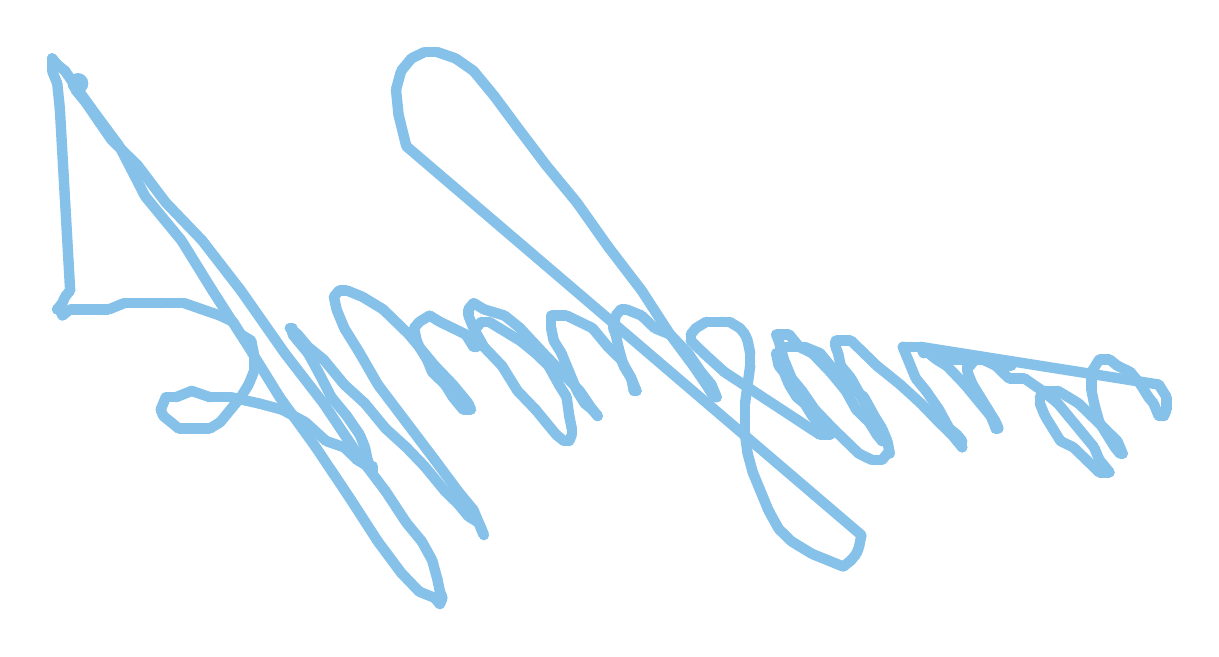}} &
      \fliph{\includegraphics[height=\imh]{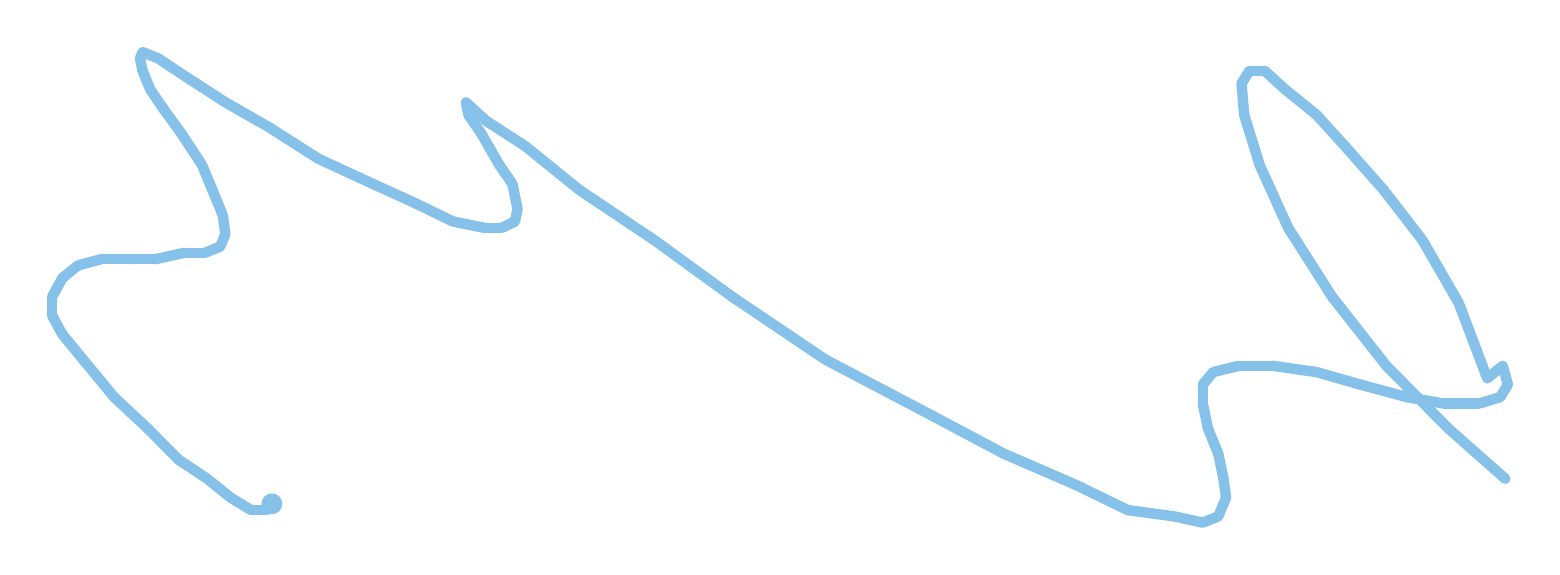}} &
      \fliph{\includegraphics[height=\imh]{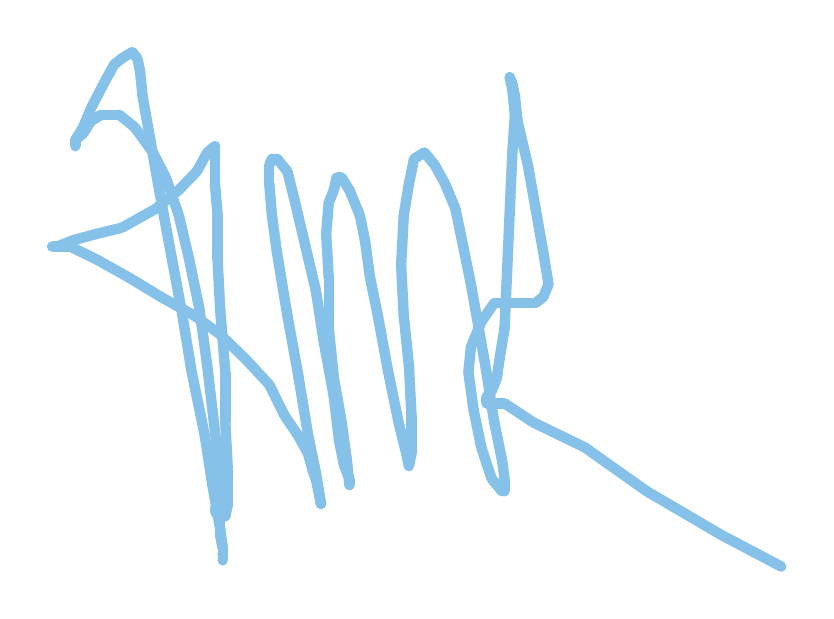}} &
      \fliph{\includegraphics[height=\imh]{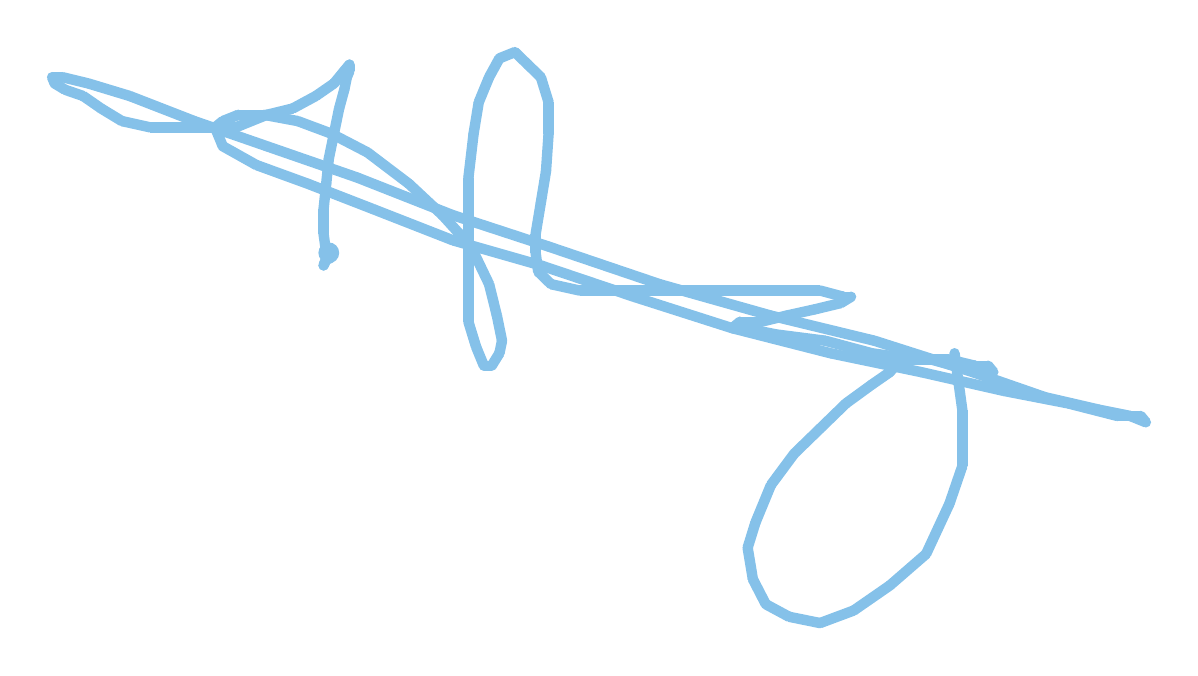}} \\
      \fliph{\includegraphics[height=\imh]{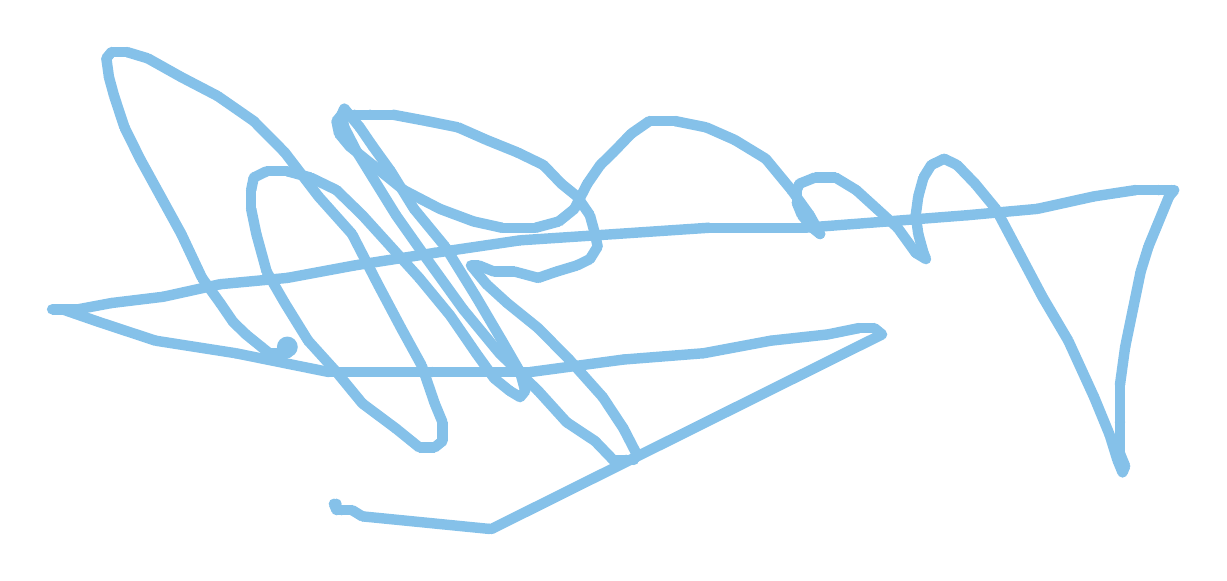}} &
      \fliph{\includegraphics[height=\imh]{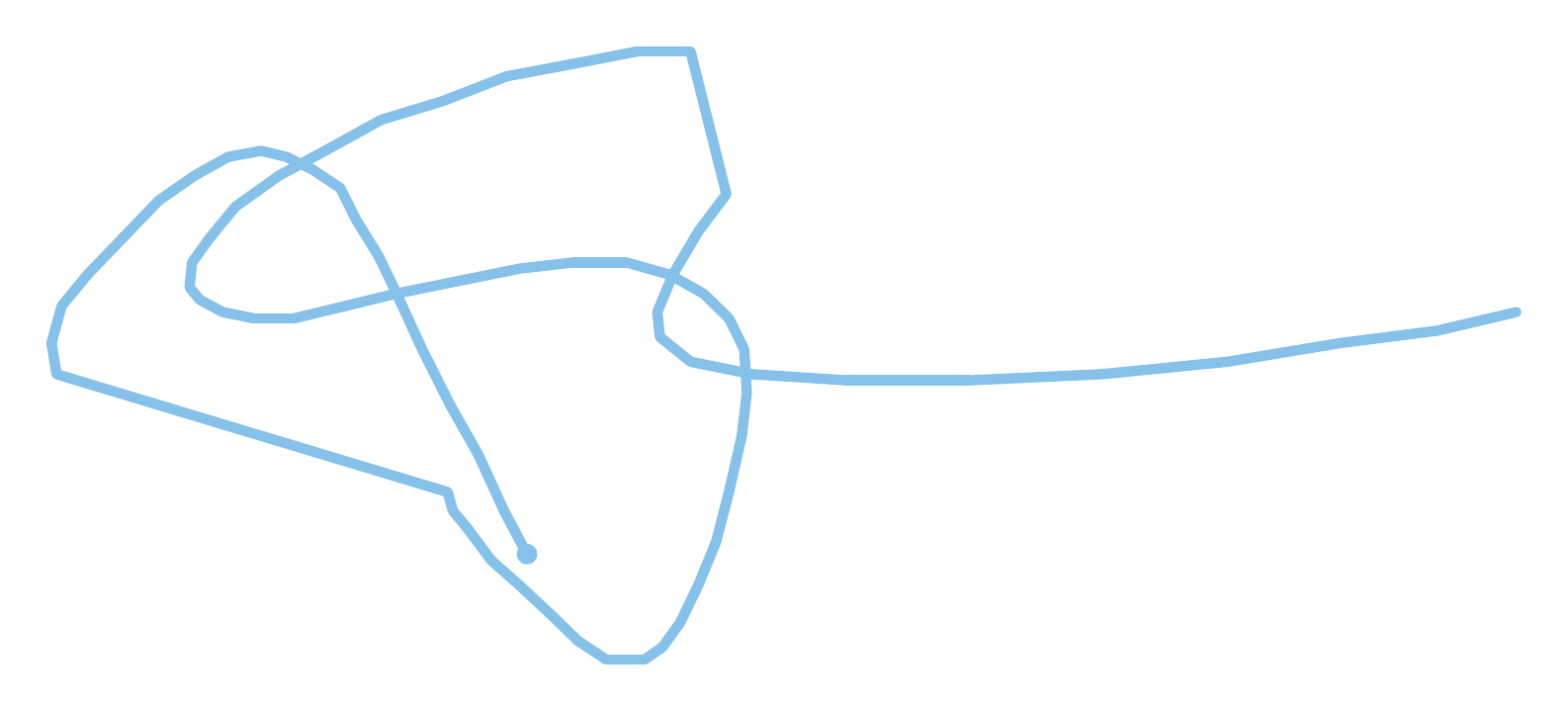}} &
      \fliph{\includegraphics[height=\imh]{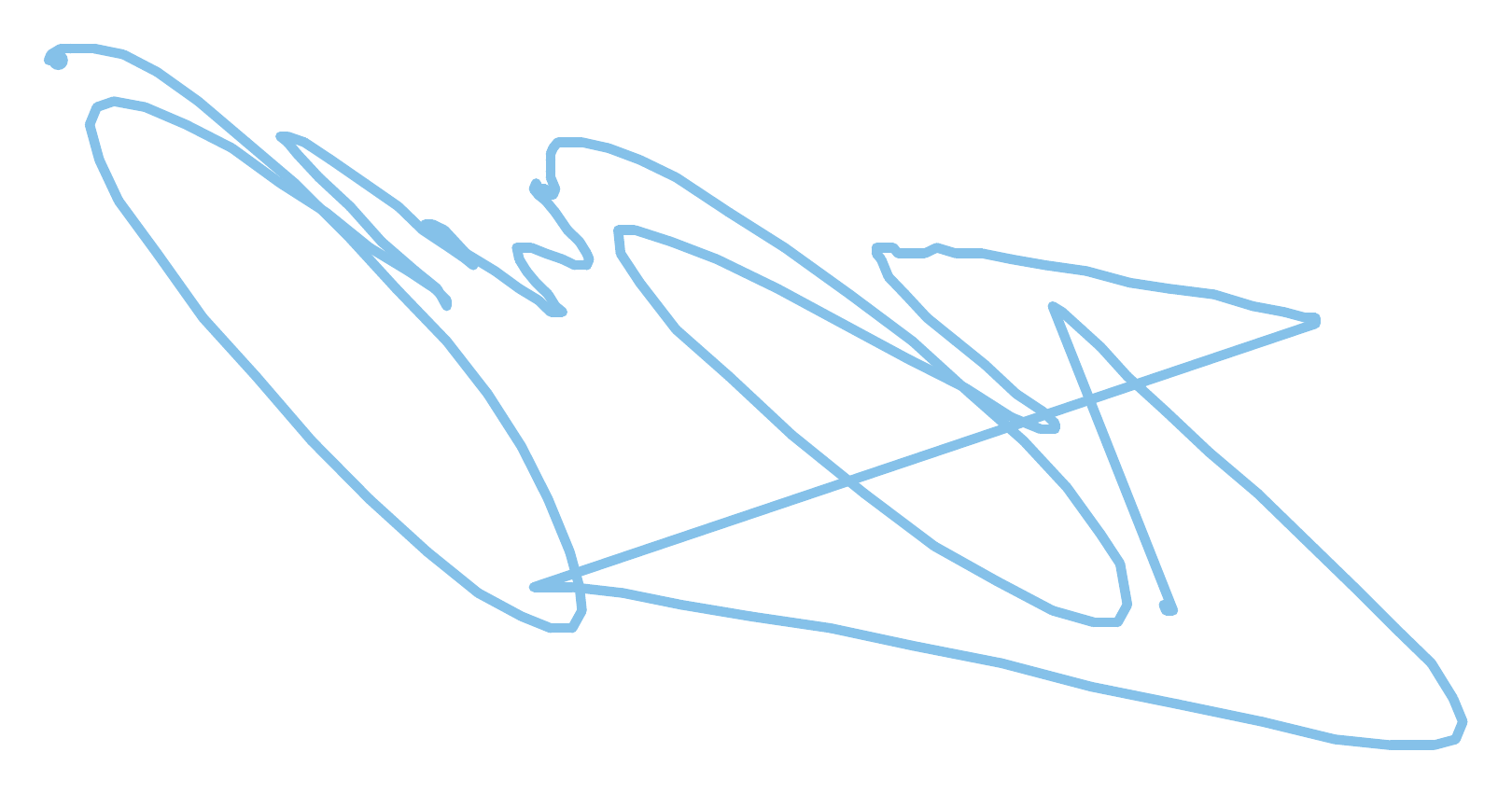}} &
      \fliph{\includegraphics[height=\imh]{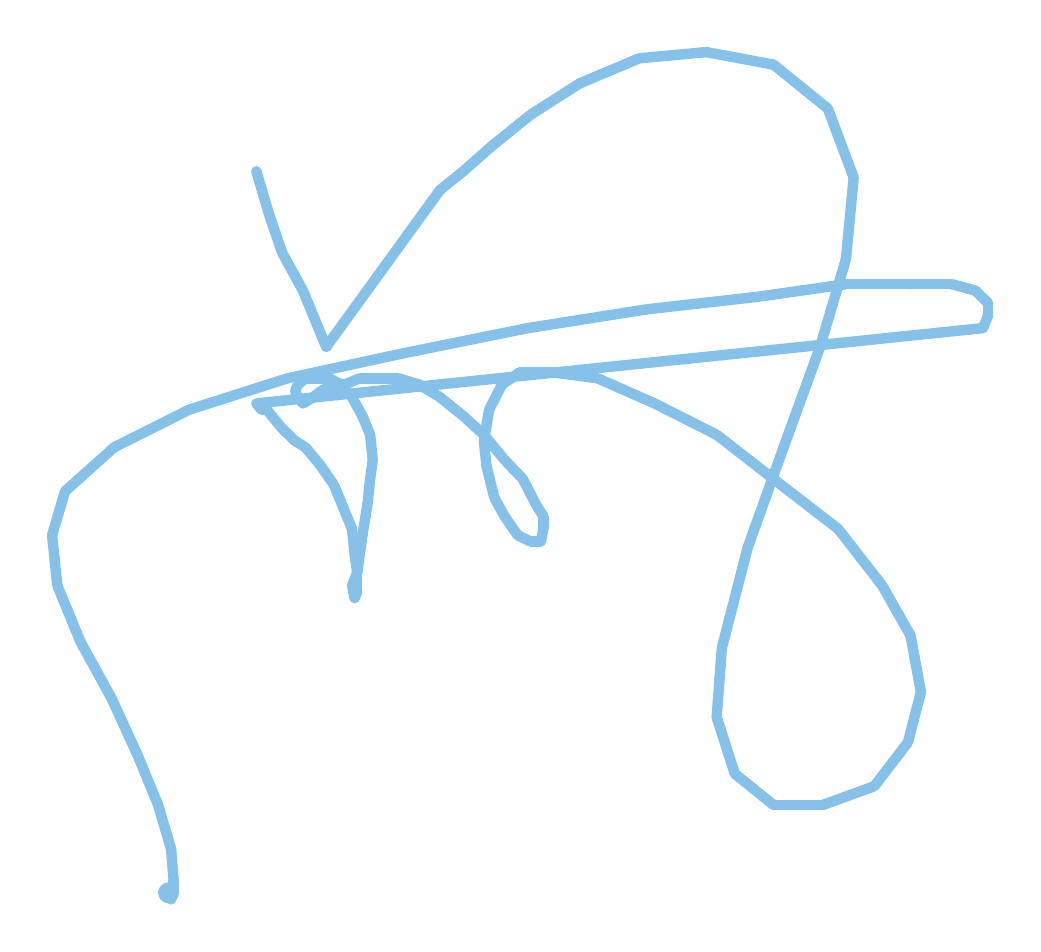}} \\
      \midrule
    \end{tabular}
    \hspace{0.25em}
    \begin{tabular}[b]{*6c}
      \multicolumn{4}{c}{\scriptsize Machine} \\
      \midrule
      \fliph{\includegraphics[height=\imh]{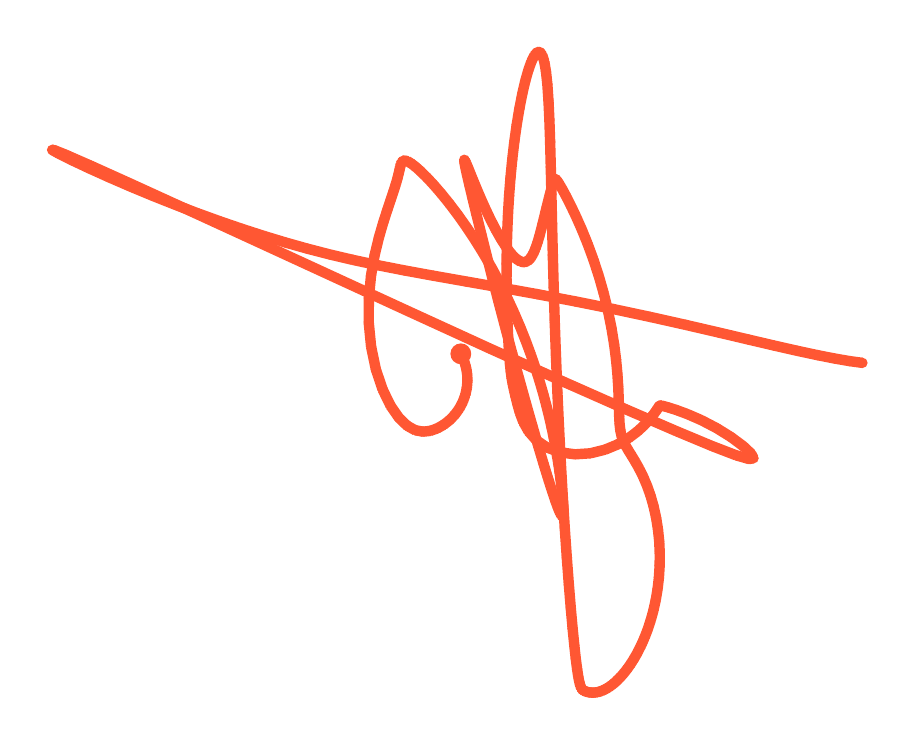}} &
      \fliph{\includegraphics[height=\imh]{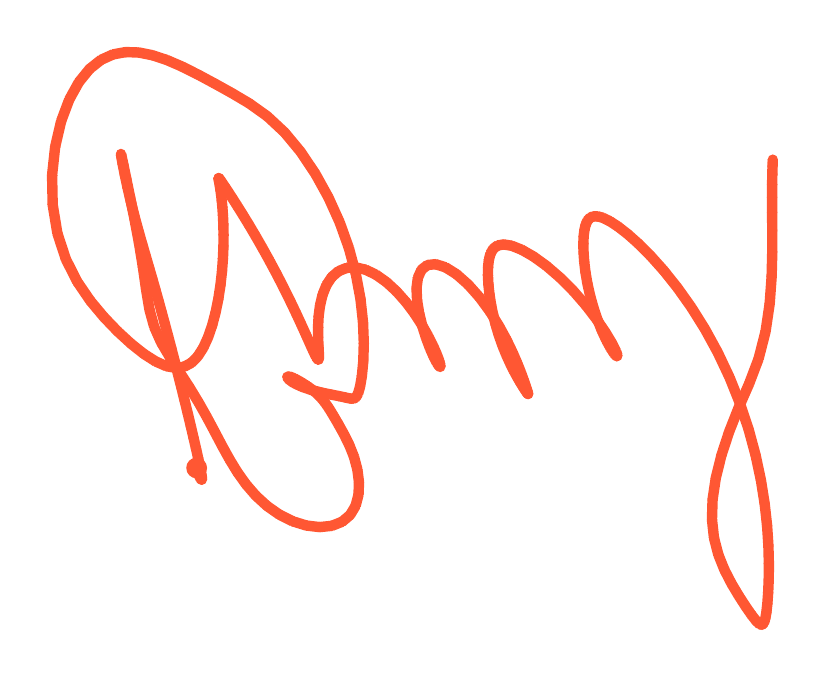}} &
      \fliph{\includegraphics[height=\imh]{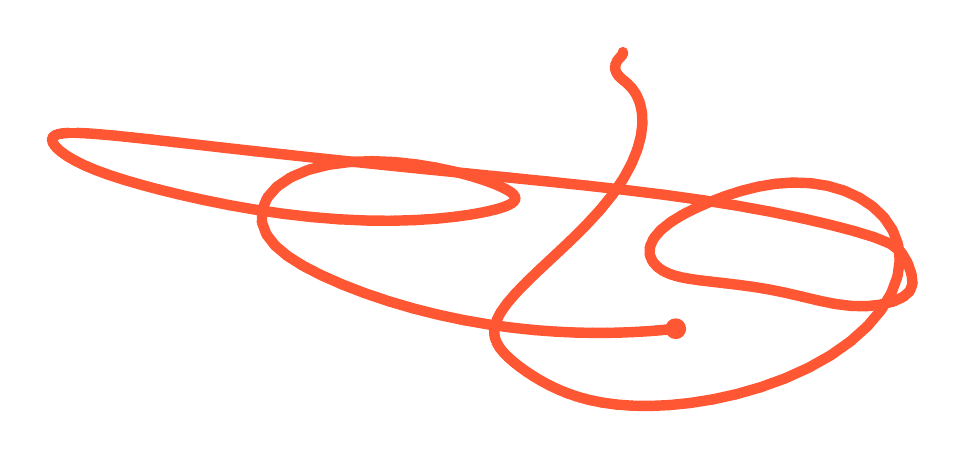}} &
      \fliph{\includegraphics[height=\imh]{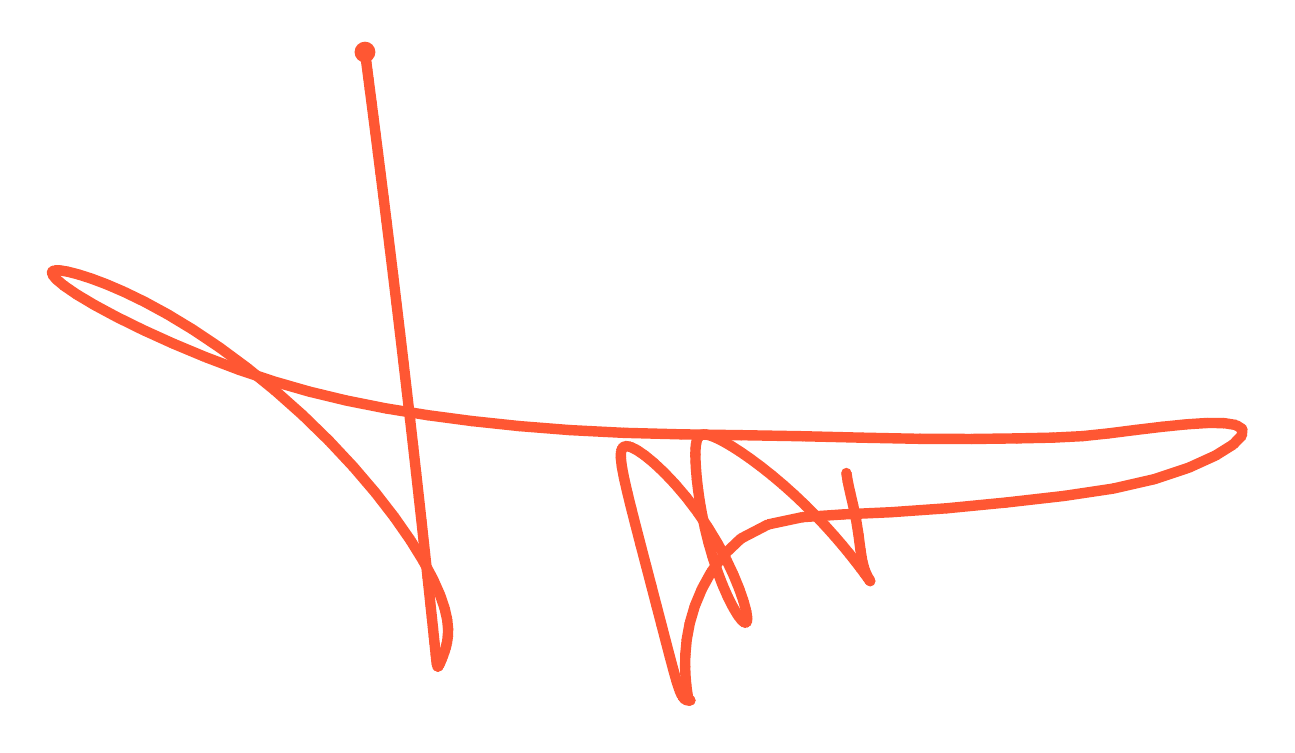}} \\
      \fliph{\includegraphics[height=\imh]{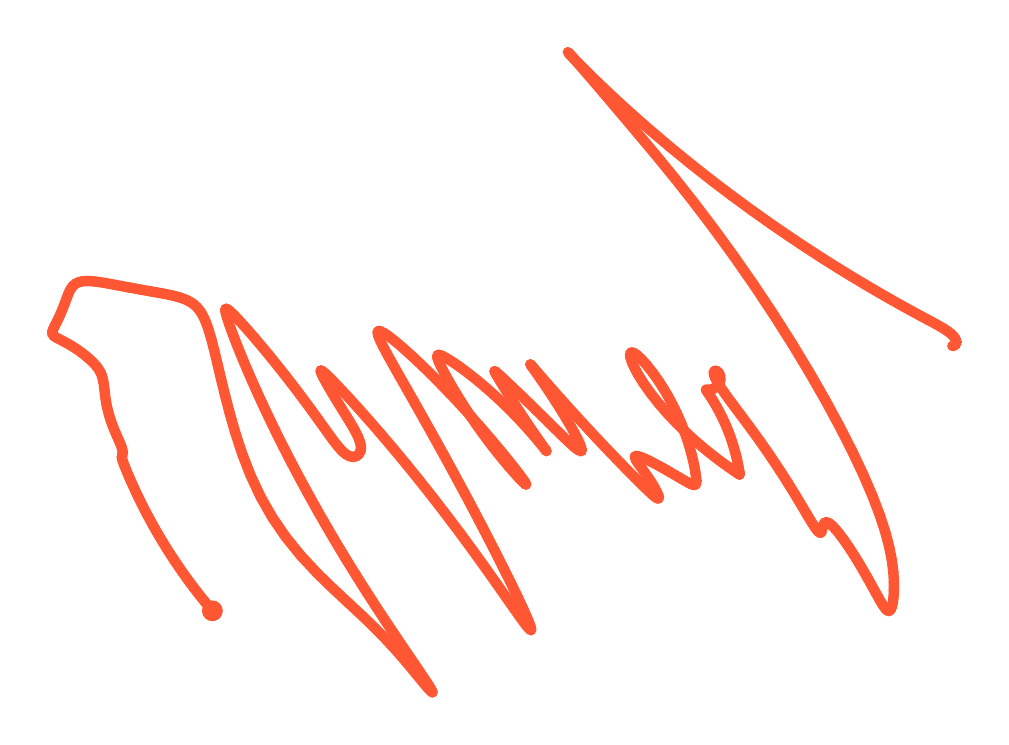}} &
      \fliph{\includegraphics[height=\imh]{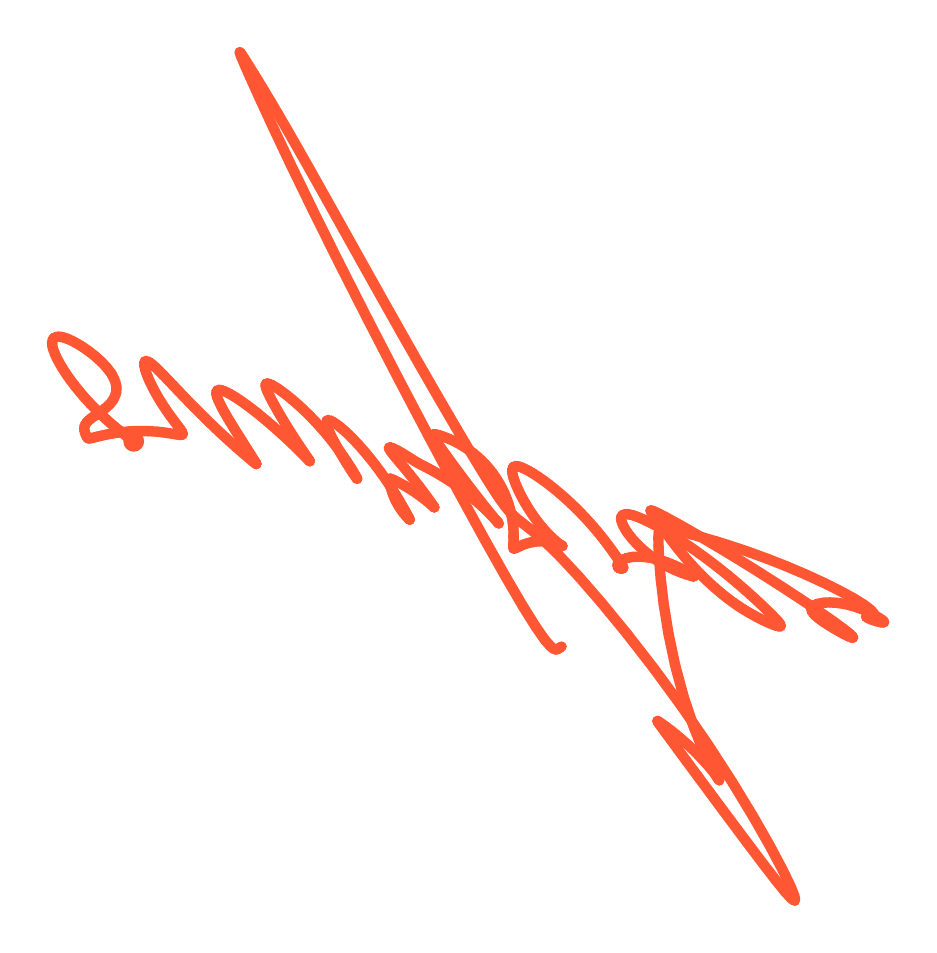}} &
      \fliph{\includegraphics[height=\imh]{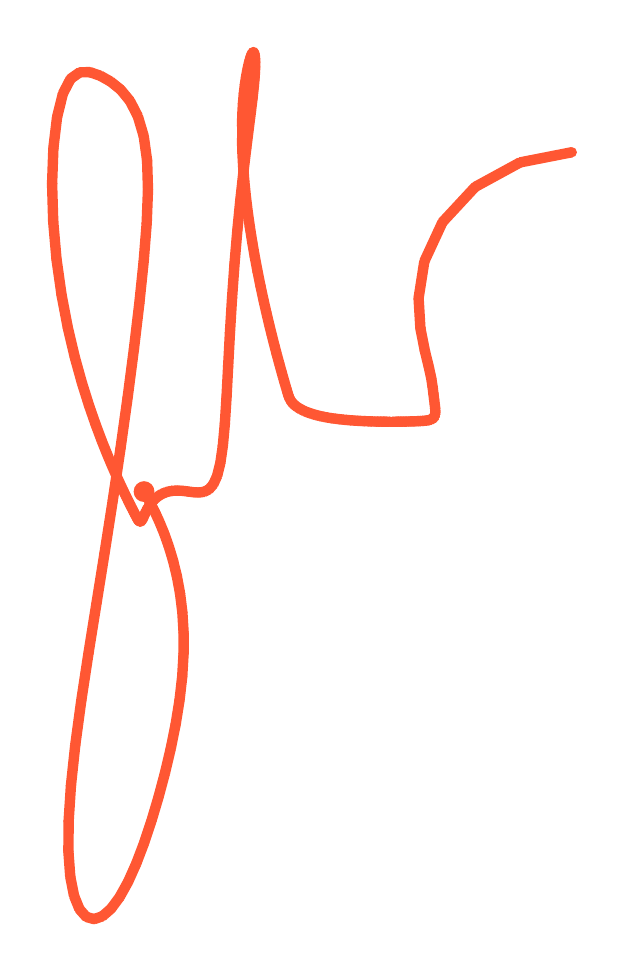}} &
      \fliph{\includegraphics[height=\imh]{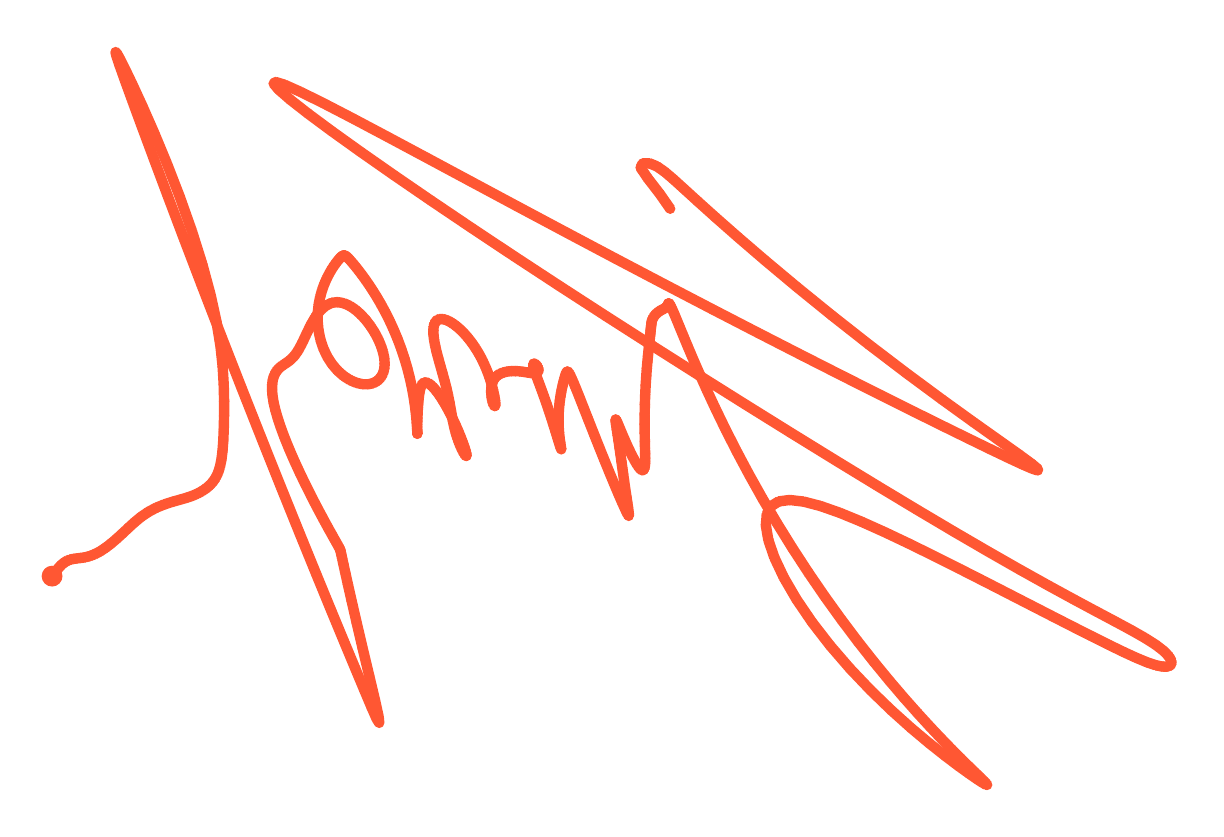}} \\
      \fliph{\includegraphics[height=\imh]{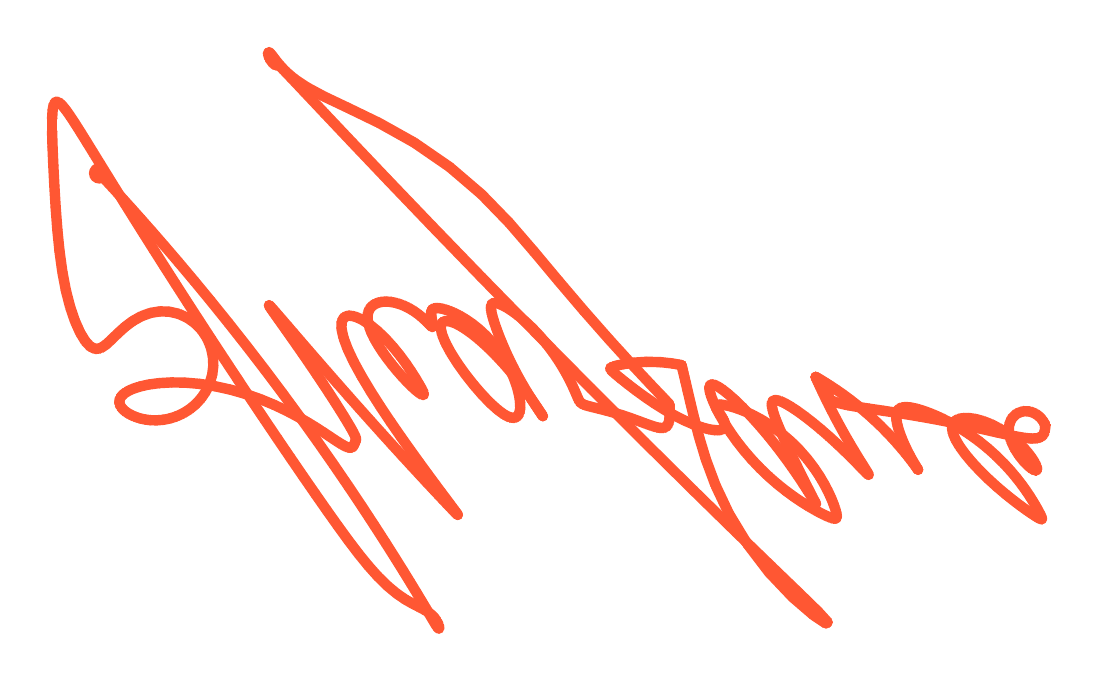}} &
      \fliph{\includegraphics[height=\imh]{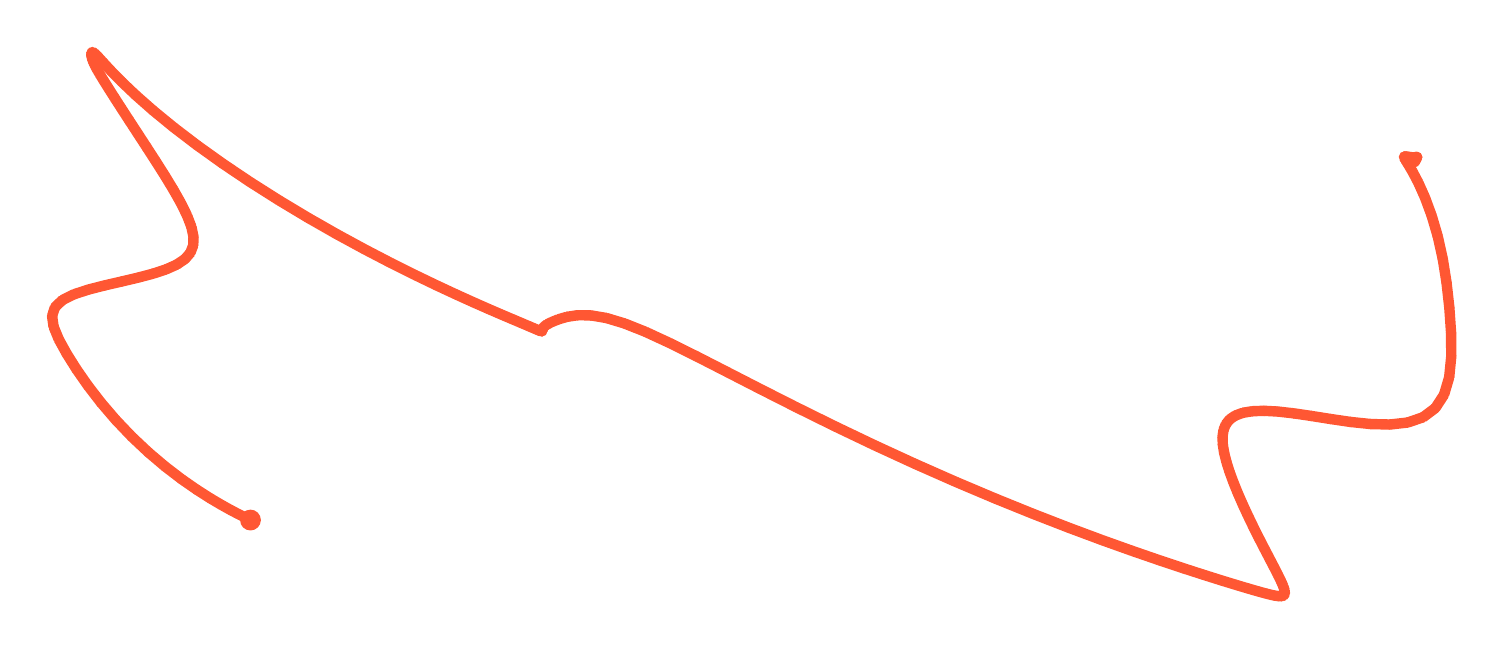}} &
      \fliph{\includegraphics[height=\imh]{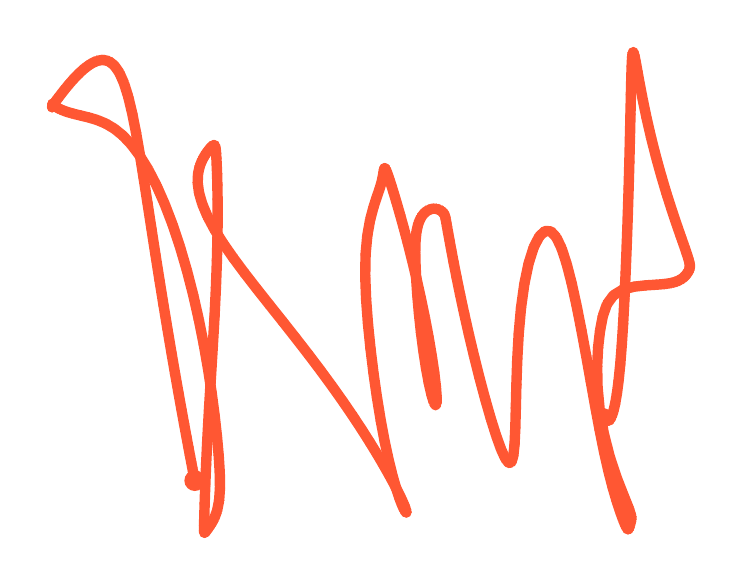}} &
      \fliph{\includegraphics[height=\imh]{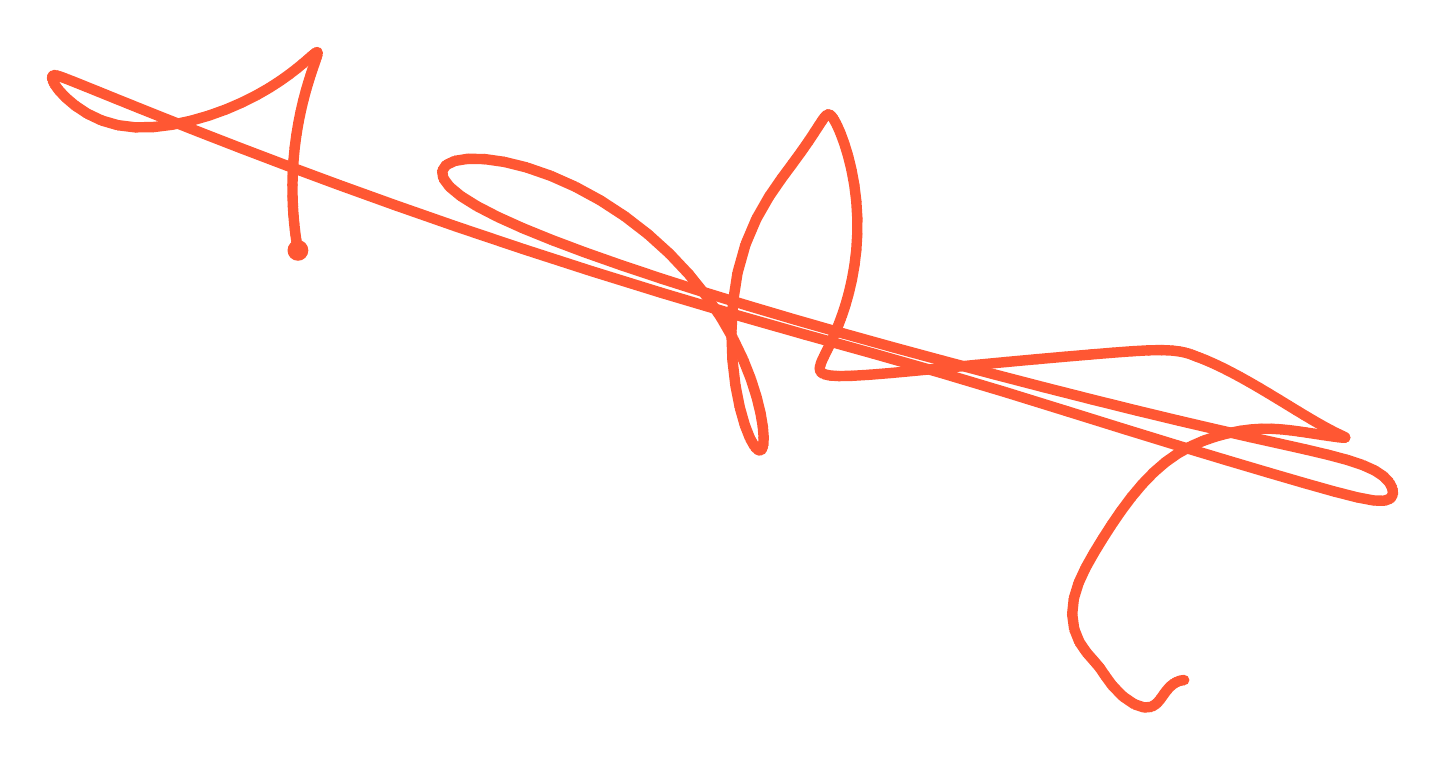}} \\
      \fliph{\includegraphics[height=\imh]{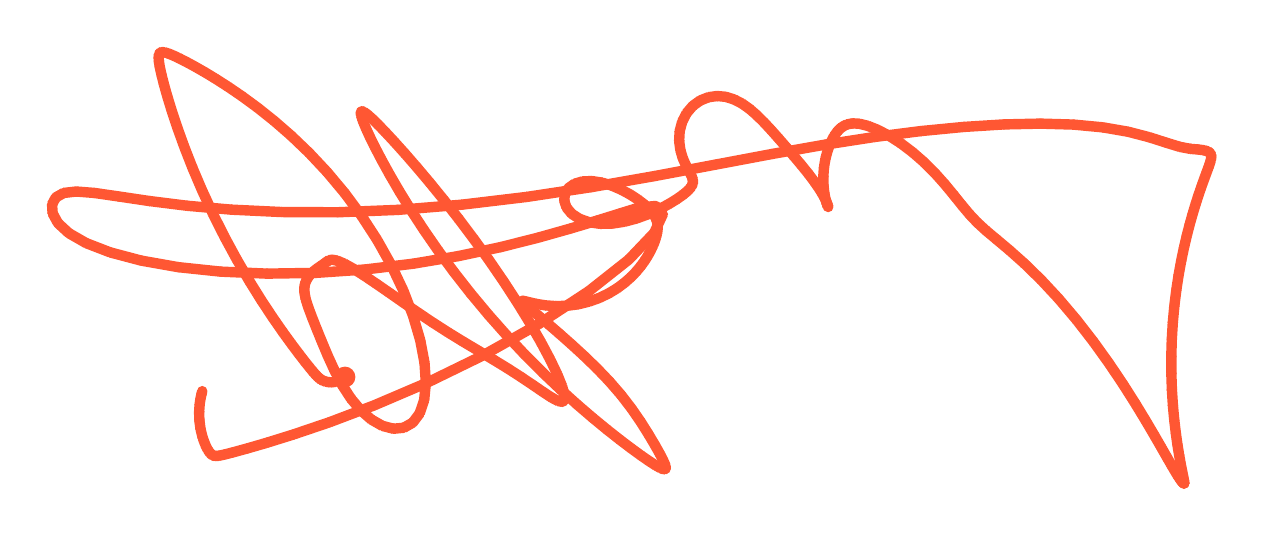}} &
      \fliph{\includegraphics[height=\imh]{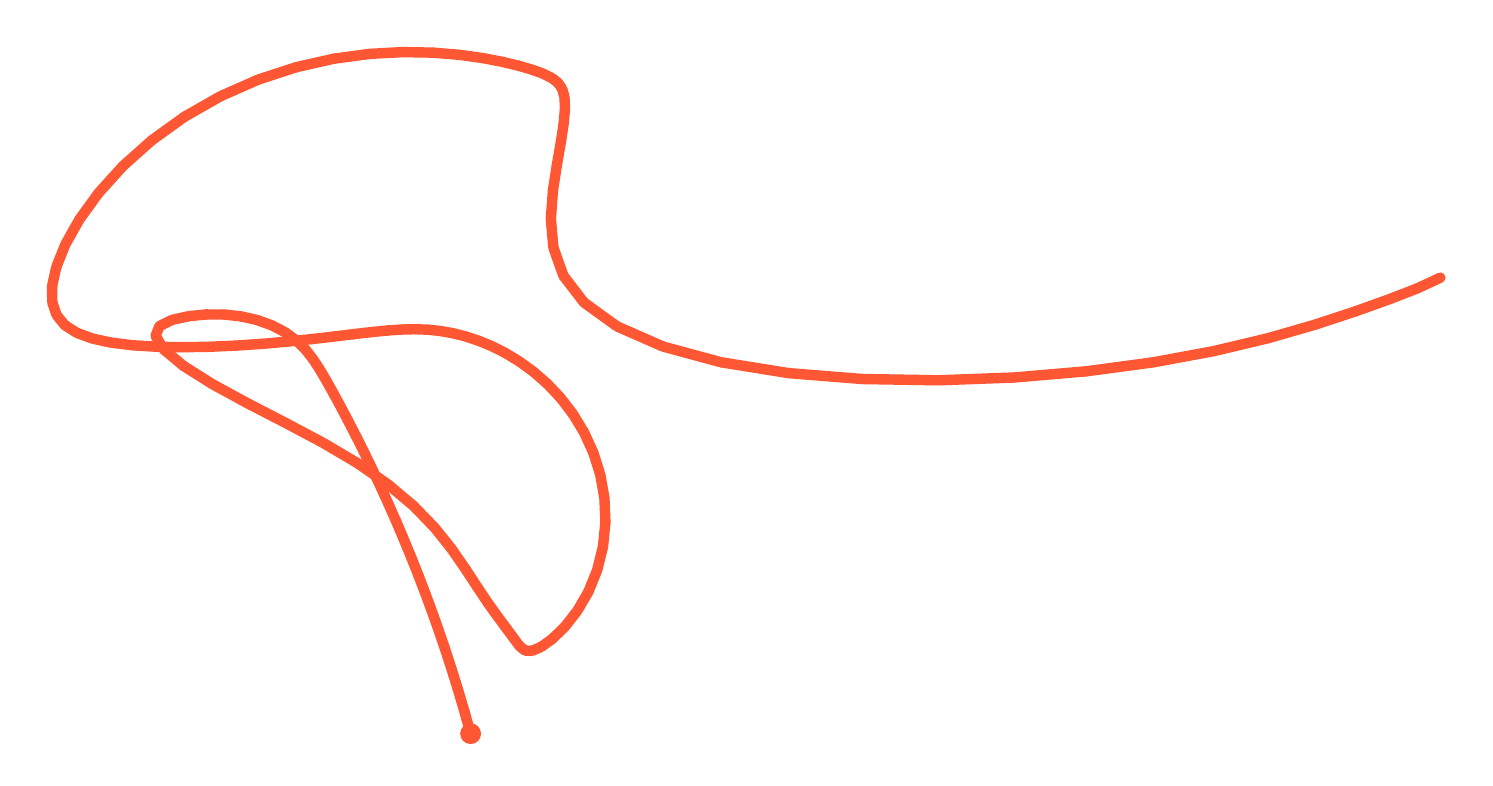}} &
      \fliph{\includegraphics[height=\imh]{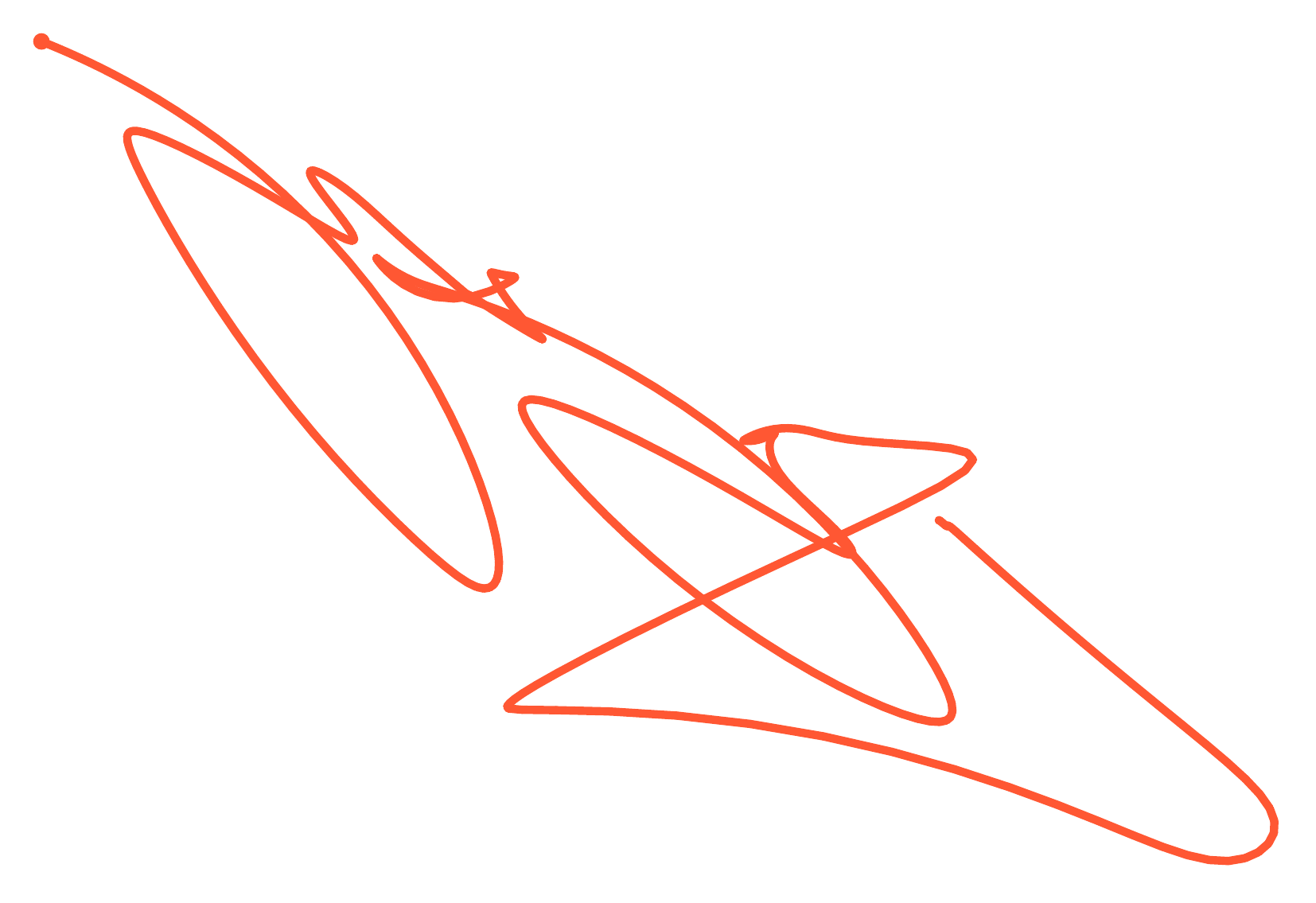}} &
      \fliph{\includegraphics[height=\imh]{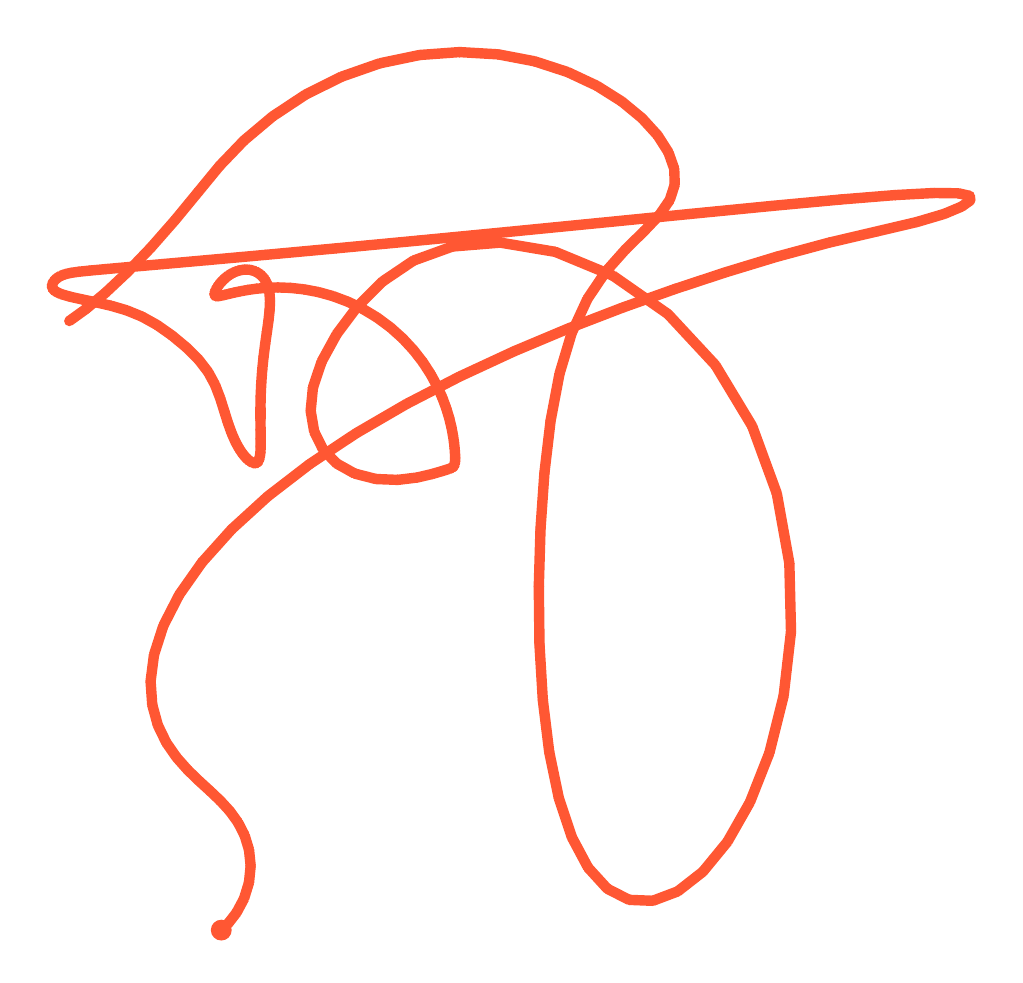}}\\
      \midrule
    \end{tabular}
  }

  \caption{
    Randomly selected examples of human and machine-generated specimens from each of the analyzed datasets.
  }
  \label{fig:datasets}
\end{figure*}

We analyze several handwriting datasets in order to cover a range of different application domains.
Specifically, we analyze two datasets of handwritten gestures,
a dataset of isolated characters and digits,
and a dataset of signatures.
In all datasets, users could see their own handwriting while articulating the symbols.
All datasets are publicly available in synthetic form~\cite{Leiva15_g3, 7775072},
an example of each is shown in \autoref{fig:datasets}.

\textbf{\$1-GDS}:
The dataset comprises 16 unistroke gesture classes, 5,280 samples in total~\cite{Wobbrock07}.
Ten users
provided 10 samples per symbol class
using an iPAQ Pocket PC (stylus as input device).

\textbf{\$N-MMG}:
The dataset comprises 16 multistroke gesture classes, 9,600 samples in total~\cite{Anthony10}.
Twenty users provided 10 samples per symbol class
using either finger (half of the users) or stylus as input devices on a Tablet PC.

\textbf{Chars74k}:
The dataset comprises 62 handwritten classes (0-9, A-Z, a-z), unistrokes and multistrokes, 3,410 samples in total.
Fifty-five users provided 1 sample per class using a Tablet PC at a constant sampling rate of 100\,Hz.

\textbf{SUSIGv}:
This is the visual subcorpus of the SUSIG database~\cite{kholmatov2009susig}, 1880 samples in total.
Ninety-four users provided 20 executions of their own signature using a stylus
on an Interlink Electronics' ePad-ink tablet at a constant sampling rate of 100\,Hz.
As we focused on genuine handwriting, the skilled forgeries were not analyzed.

\begin{figure*}[!ht]
  \centering
  \captionsetup[subfigure]{labelfont=tinier, position=top}

  \begin{minipage}[t]{0.49\textwidth}
      \hfill\subfloat[]{\includegraphics[height=0.6cm]{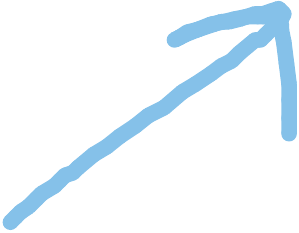}}\hspace{0.2em}
            \subfloat[]{\includegraphics[height=0.6cm]{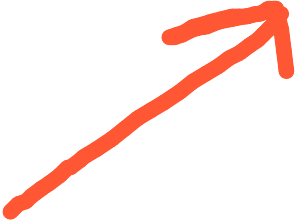}} \\
      \includegraphics[width=0.95\textwidth]{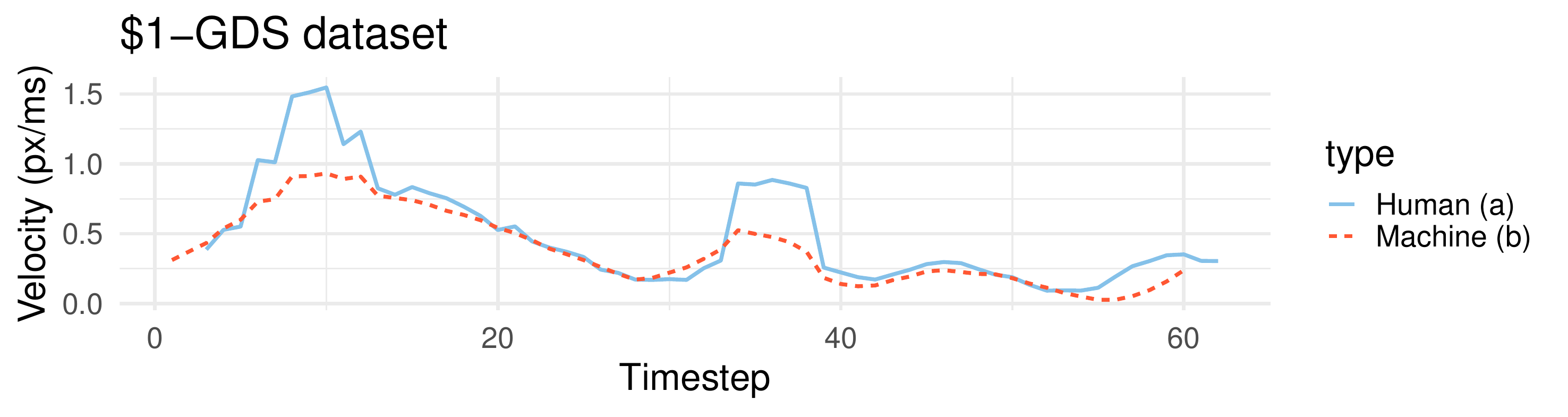}
  \end{minipage}
  \setcounter{subfigure}{0}
  \begin{minipage}[t]{0.49\linewidth}
      \hfill\subfloat[]{\includegraphics[height=0.7cm]{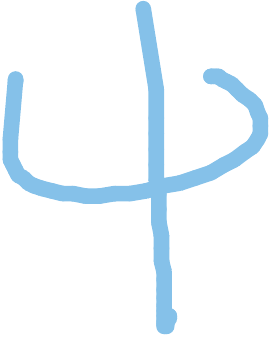}}\hspace{0.2em}
            \subfloat[]{\includegraphics[height=0.7cm]{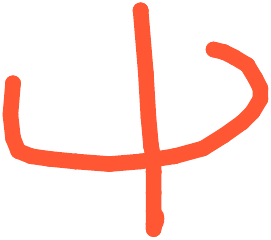}} \\
      \includegraphics[width=0.95\textwidth]{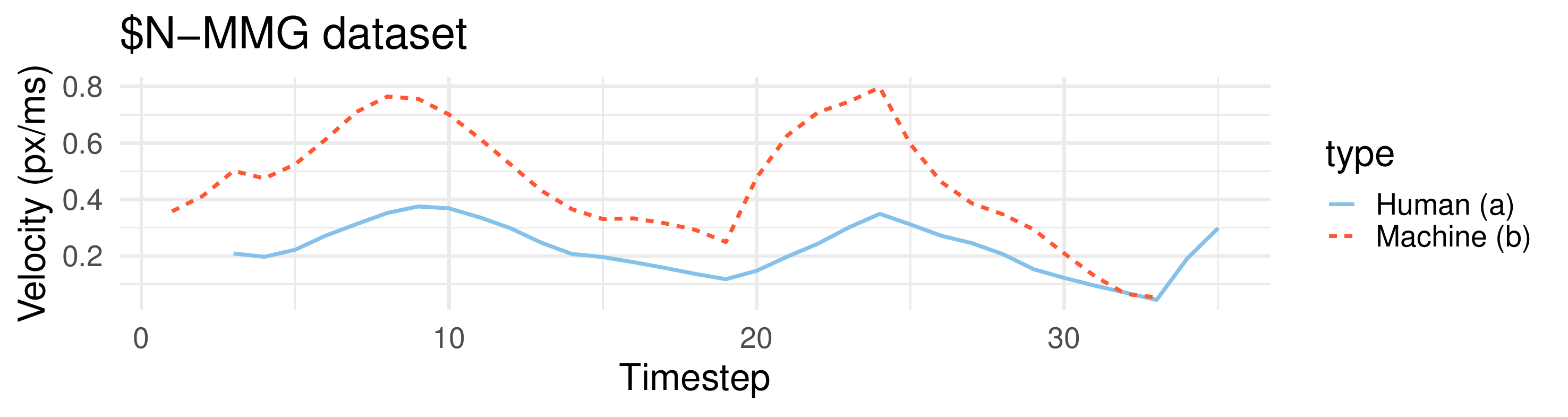}
  \end{minipage}
  \setcounter{subfigure}{0}
  \begin{minipage}[t]{0.49\textwidth}
      \hfill\subfloat[]{\includegraphics[height=0.6cm]{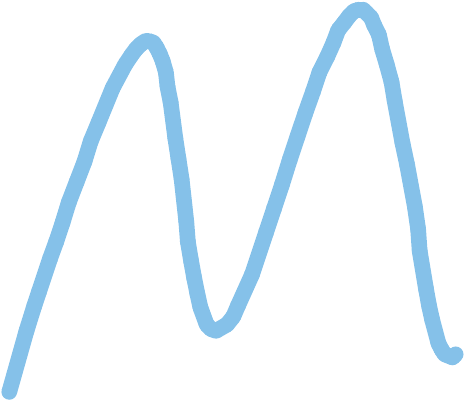}}\hspace{0.2em}
            \subfloat[]{\includegraphics[height=0.6cm]{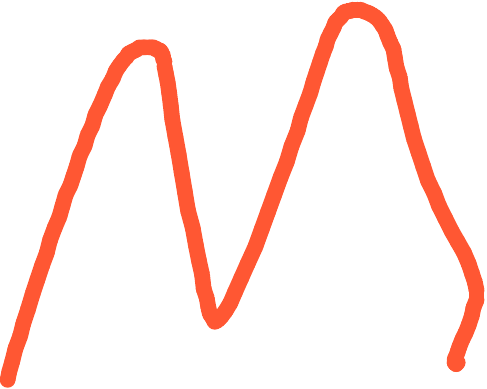}} \\
      \includegraphics[width=0.95\textwidth]{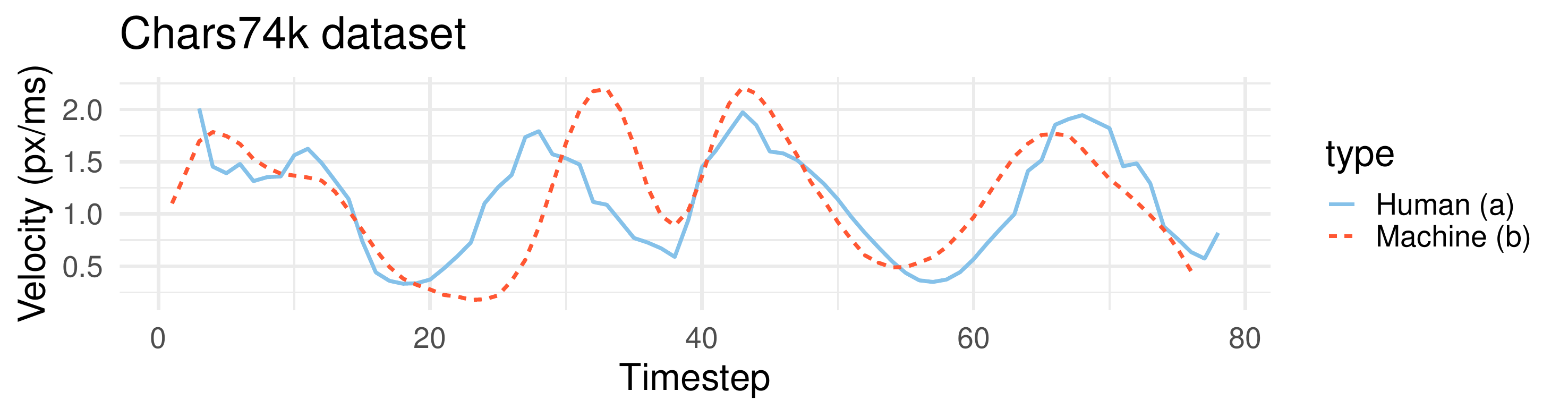}
  \end{minipage}
  \setcounter{subfigure}{0}
  \begin{minipage}[t]{0.49\linewidth}
      \hfill\subfloat[]{\includegraphics[height=0.7cm]{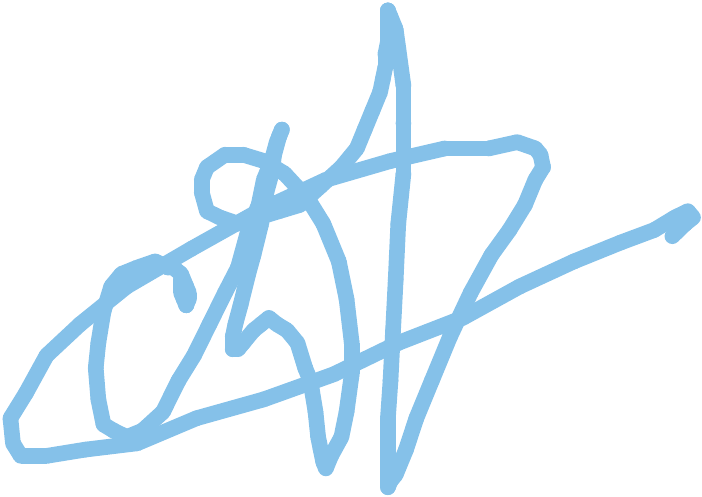}}\hspace{0.2em}
            \subfloat[]{\includegraphics[height=0.7cm]{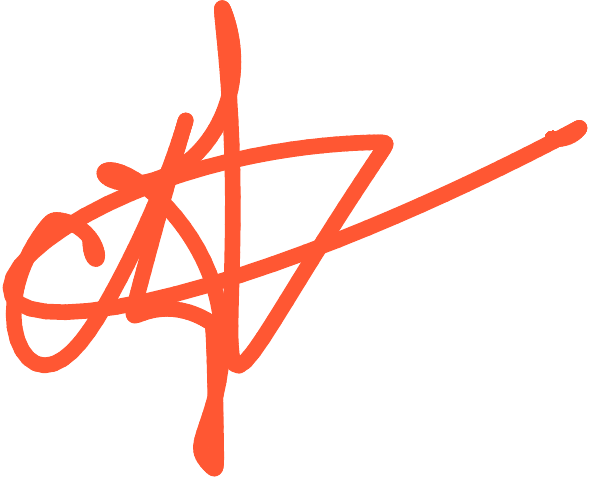}} \\
      \includegraphics[width=0.95\textwidth]{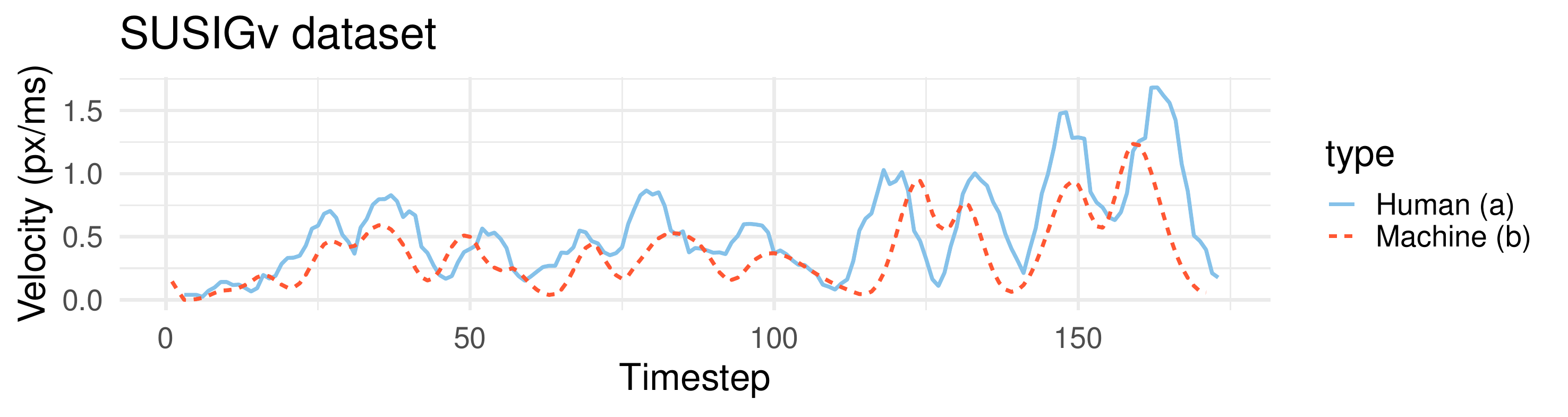}
  \end{minipage}

  \caption{
    Velocity profile examples from our evaluated datasets,
    describing how a handwriting movement ``unfolds'' over time.
    A moving average filter of size 3 is applied to remove artificial jitter introduced by the input device.
    For each human movement, a synthetic version is generated with the \slm model
    and plotted together with their human counterpart.
    It can be observed that synthetic and human samples are visually similar
    but the synthetic velocity profiles are smoother than their human counterparts.
  }
  \label{fig:velocity-profiles}
\end{figure*}

The following procedure was adopted
to create a machine-generated movement execution
for a given human sample in each of these datasets.
First, each human sample is modeled (reconstructed)
with the ScriptStudio stroke extractor~\cite{o2009development}.
Then, the \slm model parameters of the reconstructed sample are perturbed
using an additive noise distribution~\cite{Leiva15_g3}
whose values were derived in previous work~\cite{Galbally12a, 7775072}.
After these perturbations, a new synthetic sample is obtained,
which simulates a new articulation instance of the original human sample.
\autoref{fig:datasets}
provides some examples of human and synthetic samples,
and \autoref{fig:velocity-profiles} provides examples
both in visual and temporal form.

\subsection{Convolutional Neural Nets}

We analyze a static representation of human and machine-generated data,
to compare against previous work that used this representation with end-users~\cite{Galbally12b, Leiva17_slm, Leiva17_gestures}.
We investigate different convolutional neural network (CNN) architectures.
CNNs have emerged as the master algorithm in computer vision in recent years~\cite{chollet2016xception}
and have spurred further breakthroughs in machine learning.
Concretely, we tested the following popular architectures:
VGG16~\cite{Simonyan15},
ResNet50~\cite{he2015deep},
DenseNet~\cite{huang2016densely},
Inception~\cite{szegedy2014going},
and Xception~\cite{chollet2016xception}.

\begin{figure}[!ht]
    \centering
    \includegraphics[width=0.95\linewidth]{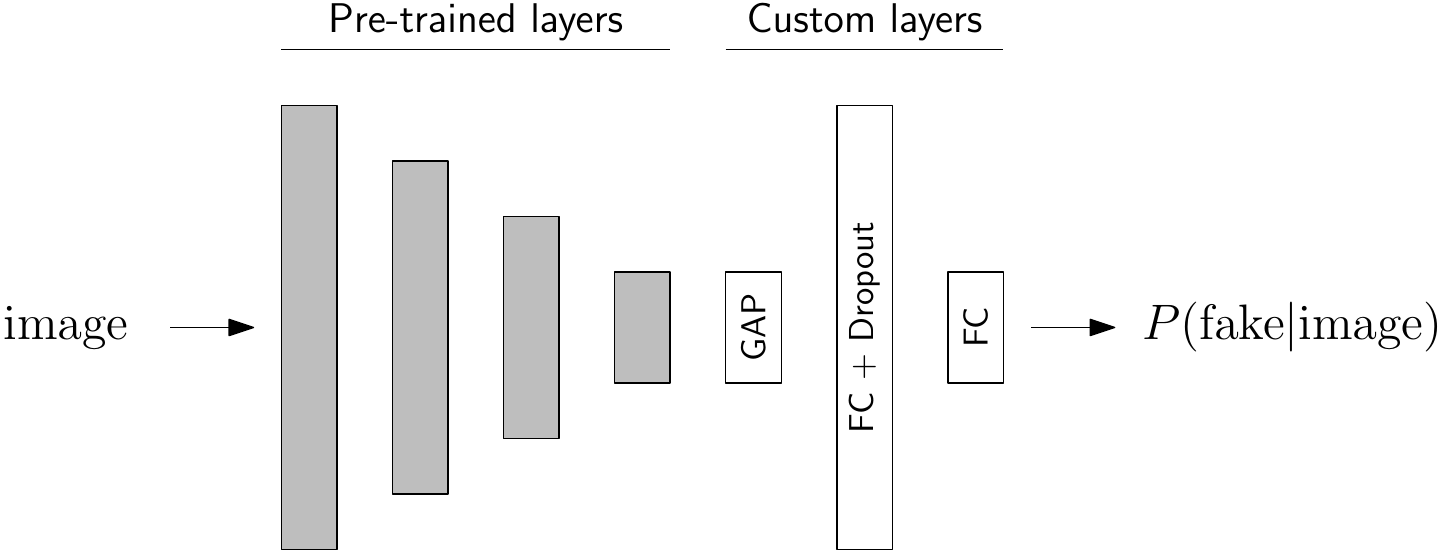}
    \caption{Conceptual representation.
      Our CNN models were built on top of existing pre-trained network architectures,
      which we extended and fine-tuned for our classification task.}
    \label{fig:cnn-models}
\end{figure}

All these CNN architectures are publicly available
and have provided state-of-the-art performance in image classification tasks.
These models were trained on the ImageNet database~\cite{Deng09_imagenet},
so we fine-tune them to our datasets via transfer learning~\cite{Pan10_transfer}.
To this aim:
(1)~the input layer is modified to accept images of any size;
(2)~all layers from the pre-trained architecture are frozen so that they become non-trainable;
(3)~a global average pooling (GAP) layer is added after the last of the pre-trained layers,
followed by a fully connected (FC) layer of $4096$ neurons with ReLU activation and Dropout rate $q=0.5$;
and
(4)~the last softmax layer\footnote{All pretrained models were designed to distinguish among 1000 classes.}
    is replaced by an FC layer with one neuron and sigmoid activation.\footnote{
    Alternatively, an FC layer with $2$ neurons and softmax activation can be used.
    However, since we are interested in determining if a movement is human or not,
    a single neuron with sigmoid activation is a more elegant design choice.}
A key advantage of GAP layers is that they enforce correspondences between feature maps and output categories,
which feels much more native to the convolution structure and help to prevent overfitting~\cite{Lin14_gap}.
Our additional layers are trained with backpropagation, as usual in neural nets.
\autoref{fig:cnn-models} conceptually illustrates our pre-trained CNN architectures.

We also train a custom CNN model from scratch,
aiming for a straightforward but high-performance classifier.
The model has two stacked Convolutional layers with $64$ and $32$ filters of 3x3 kernel size and ReLU activation,
followed by a GAP layer, an FC layer of $2048$ neurons with ReLU activation and Dropout rate $q=0.25$,
and an FC layer of one neuron with sigmoid activation as output layer.
This model thus promotes a rather simple architecture with several but small receptive fields,
inspired by the human feed-forward vision system~\cite{Kheradpisheh16}.
Note that the FC layer has significantly less neurons that in our pre-trained CNNs
and the Dropout rate is less aggressive
since this architecture is much simpler.

All CNN models are trained with the popular Adam optimizer~\cite{Kingma15_adam}
with learning rate $\eta=0.0005$ and decay rates $\beta_1=0.9, \beta_2=0.999$.
The loss function to minimize is binary cross-entropy, since our task is a two-class classification problem.
We feed all CNN models with images of 160x160\,px resolution,
which did not cause noticeable pixel distortions after resizing, as can be seen in \autoref{fig:datasets}.
The models are trained up to 400 epochs in batches of 64 images.
We use early stopping with patience of 40 epochs, to prevent overfitting,
i.e., if some monitoring metric (in our case, classification accuracy)
does not improve in 40 consecutive epochs, training is finalized and the best model weights are retained.
We train the models on 70\% of the data, validate on 10\% of the data, and test on the remaining 20\% of the data.
The results of these experiments are shown in the top rows of \autoref{tbl:results-dl-gds}.

\subsection{Recurrent Neural Nets}

We investigate different recurrent neural network (RNN) architectures,
which are designed to process sequential data,
such as the handwriting movements we analyze in this paper,
and can model temporal dynamic behavior,
so they seem ideal candidates for this classification task.
Currently there are no pre-trained deep learning architectures for handwriting movements,
so we designed our own RNN models from scratch.
We tested the vanilla RNN cell as well as the popular LSTM~\cite{Hochreiter97_lstm} and GRU~\cite{Cho14_gru} cells,
together with their bi-directional variant~\cite{Graves05_bidir},
where the input sequence is processed in both forward and backward direction.

Since velocity is considered the fundamental control variable in human handwriting~\cite{Bernier05, Hartwig01}
our RNN models use velocity as single input feature,
which is computed as the point-wise Euclidean distance (in px) divided by the time offset (in ms)
between two consecutive timesteps:
\begin{equation}
\label{eq:rnn-feats}
  v_i = \dfrac{\sqrt{\Delta x_i^2 + \Delta y_i^2}}{t_{i} - t_{i-1}}
\end{equation}
where $\Delta x_i = x_{i} - x_{i-1}$ and $\Delta y_i = y_{i} - y_{i-1}$, $\forall i > 1$.
In sum, input data are converted from $(x,y,t)_i$ sequences to velocity sequences $v_i$ at every $i$th timestep.
Among other benefits, velocity is both rotation and translation invariant,
and is rather challenging to simulate reliably~\cite{Elarian14}.

\begin{figure}[!ht]
    \def\w{0.24\linewidth}

    \includegraphics[width=\w]{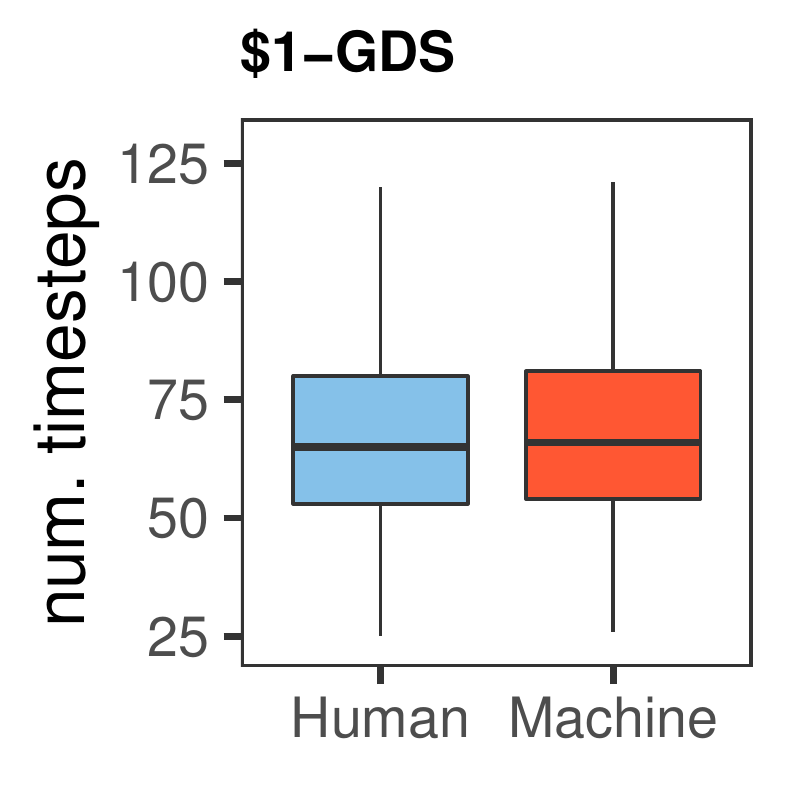}\hfill
    \includegraphics[width=\w]{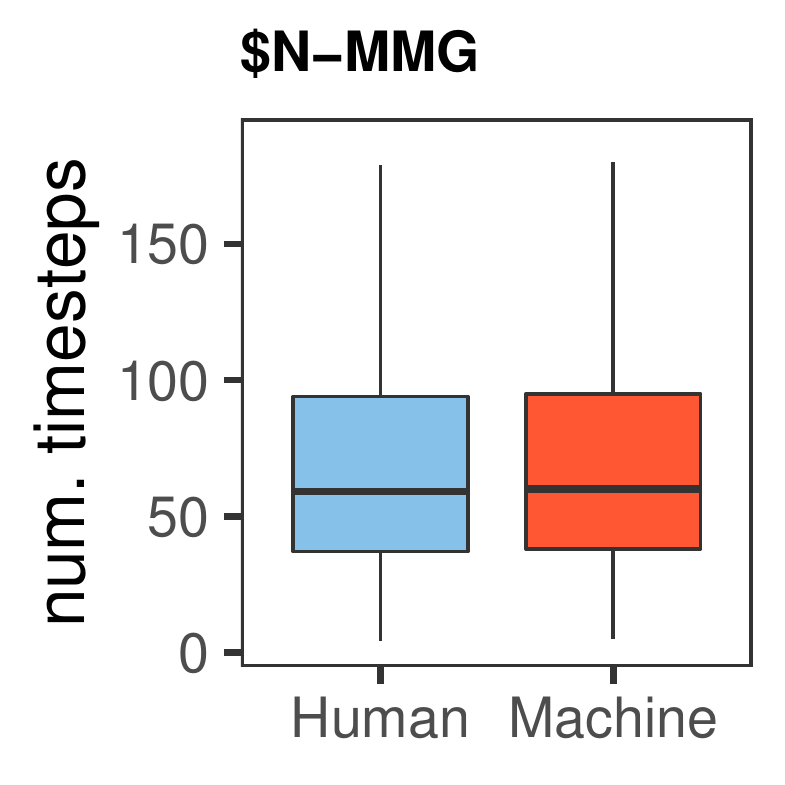}\hfill
    \includegraphics[width=\w]{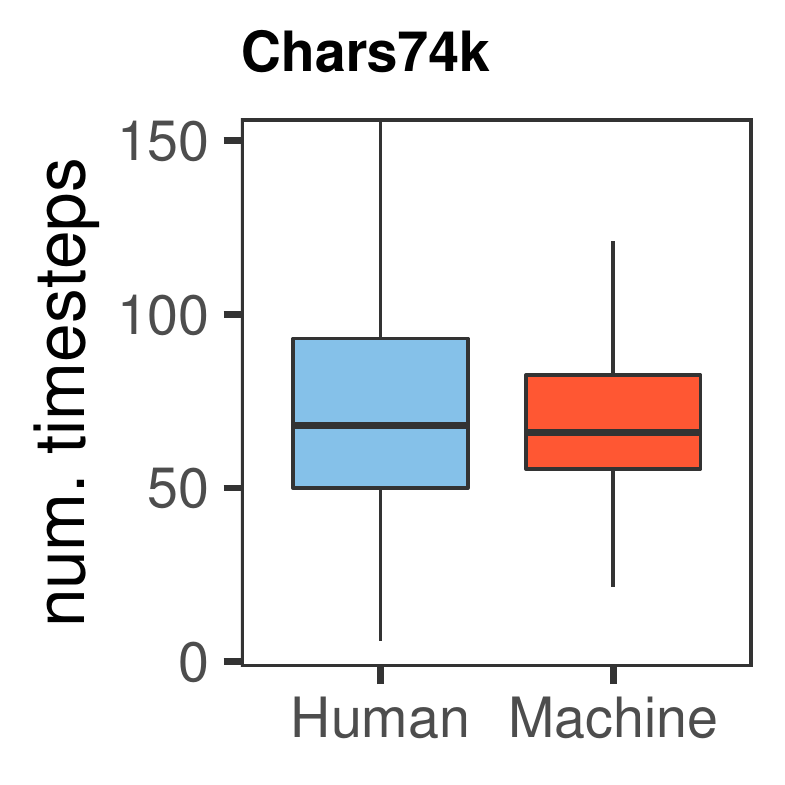}\hfill
    \includegraphics[width=\w]{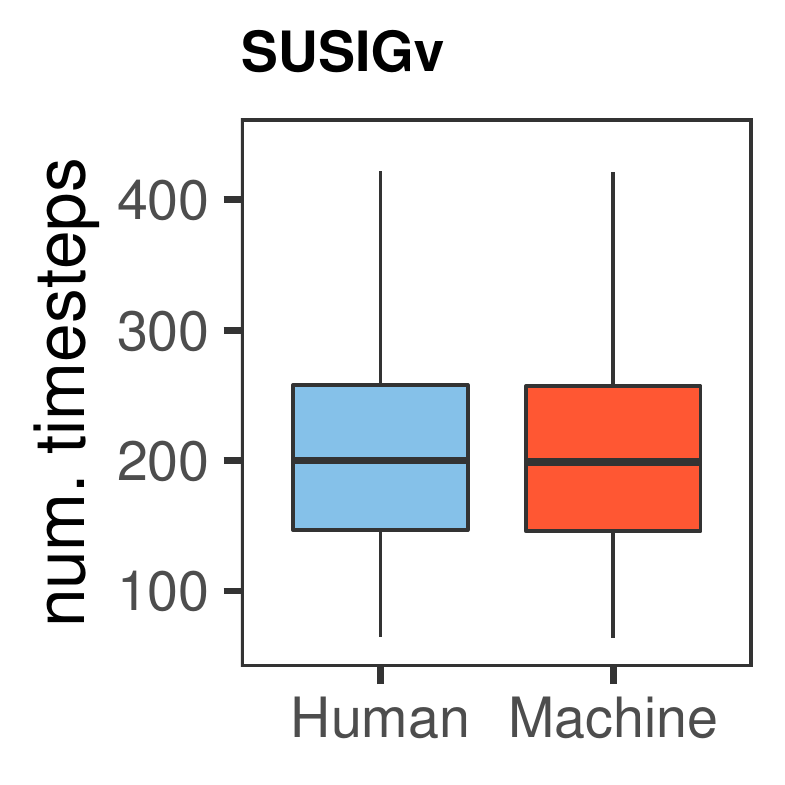}

    \caption{Analysis of sequence length (number of timesteps) to determine the maximum capacity of our RNN models.}
    \label{fig:seq-lengths}
\end{figure}

The model architecture of our RNNs is rather simple:
an input layer followed by a recurrent layer (either vanilla, LSTM, or GRU) with hyperbolic tangent activation,
embedding size of 100 units, and Dropout rate $q=0.25$,
followed by a FC layer with one neuron and sigmoid activation as output layer.
We set a maximum capacity for our RNNs to be 400 timesteps,
so that longer sequences are truncated to 400 sequence points
before feeding them to the input layer.
\autoref{fig:seq-lengths} shows that this is an adequate cap for the datasets analyzed in this paper.
We noticed that the synthetic sequences in the Chars74k and SUSIGv datasets
were sampled at twice the frequency of the human sequences.
Consequently, we downsampled them by half, i.e., by removing one point every two consecutive points,
to ensure that both human and synthetic sequences have about the same length,
as it happens in \$1-GDS and \$N-MMG.

All RNN models are trained with the Adam optimizer
using the same hyperparameters ($\eta, \beta_1, \beta_2$) used in the CNN models.
The loss function is also binary cross-entropy, as the task remains the same (two-class classification).
We feed the RNN models in batches of 128 sequences each and use up to 400 epochs for training,
as we did with the CNN models.
We also use early stopping with patience of 40 epochs, to prevent overfitting,
with classification accuracy as monitoring metric.
Finally, we also train the models on 70\% of the data, validate on 10\% of the data, and test on the remaining 20\% of the data.
The results of these experiments are shown in the bottom rows of \autoref{tbl:results-dl-gds}.

\section{Results}

Classification accuracy is not the only relevant performance metric.
Area Under the ROC curve (AUC), for example, helps to determine the discriminatory power of any classifier.
Further, assuming that human data are the positive cases and that synthetic data are the negative cases,
we report the classic Precision and Recall metrics as well.
On the one hand, Precision quantifies the number of positive class predictions that actually belong to the positive class.
On the other hand, Recall quantifies the number of positive class predictions made out of all positive cases in a dataset.
For completeness, we also report the F-measure, or balanced F1 score,
which is the harmonic mean of Precision and Recall~\cite{sasaki2007truth}.

We begin by determining the most suitable data representation (i.e. images or sequences)
and the most suitable neural network architecture for classification.
We experiment with the \$1-GDS dataset, since it is large enough
and is available both in off-line and on-line form~\cite{Leiva17_gestures}.
\autoref{tbl:results-dl-gds} shows the results of these experiments.
Top rows are CNN-based image classifiers that use spatial information only.
Bottom rows are RNN-based sequence classifiers that use both spatial and temporal information:
at every timestep the RNN is fed with the velocity of the pen tip (see \autoref{eq:rnn-feats}).
Notice that by switching from a spatial to a spatiotemporal domain
the RNN classifiers achieve remarkable classification performance.
We also consider the 1-nearest neighbor classifier with dynamic time warping (1NN-DTW) as a baseline model.
Arguably, 1NN-DTW has proven itself to be an exceptionally strong baseline for time series classification~\cite{Kate16}.

\begin{table}[!ht]
    \caption{Experiment results for the \$1-GDS dataset.
        Top rows are CNN-based image classifiers.
        Bottom rows are RNN-based sequence classifiers.}
    \label{tbl:results-dl-gds}

    \centering
    \setlength{\tabcolsep}{0.55em}

    \begin{tabular}{l *5r}
    \toprule
    \thead{System} & \thead{Precision} & \thead{Recall} & \thead{F-measure} & \thead{Accuracy} & \thead{AUC} \\
    \midrule
    VGG16         & 68.50     & 68.48     & 68.45     & 68.48     & 75.39     \\
    ResNet50      & 68.97     & 68.95     & 68.95     & 68.95     & 76.34     \\
    Xception      & 71.15     & 71.03     & 70.98     & 71.03     & 79.58     \\
    DenseNet      & 71.41     & 71.41     & 71.41     & 71.41     & 78.74     \\
    Inception     & \bf 75.09 & \bf 74.69 & \bf 74.57 & \bf 74.69 & \bf 82.82 \\
    Custom CNN    & 74.32     & 74.28     & 74.27     & 74.28     & 82.50     \\
    \midrule
    1NN-DTW       & 85.32     & 83.97     & 83.80     & 83.97     & 83.88     \\
    \midrule
    Vanilla RNN   & 94.93     & 94.51     & 94.49     & 94.51     & 94.46     \\
    LSTM          & 95.66     & 95.27     & 95.25     & 95.27     & 97.45     \\
    Bi-LSTM       & 95.14     & 94.73     & 94.72     & 94.73     & 96.98     \\
    GRU           & \bf 95.78 & \bf 95.39 & \bf 95.38 & \bf 95.39 & \bf 98.20 \\
    Bi-GRU        & 95.62     & 95.20     & 95.19     & 95.20     & 97.76     \\
    \bottomrule
    \end{tabular}
\end{table}

The perceptual experiments conducted on the gesture datasets~\cite{Leiva17_slm, Leiva17_gestures}
concluded that participants were unable to judge whether gesture images were produced by a human or by a machine,
with classification rates close to 50\%.
In this paper, we can see that it is possible to tell human and synthetic data apart
using spatial information only (top rows in \autoref{tbl:results-dl-gds})
with higher accuracy than previous work, ranging from 68\% to 75\%.
Note that our custom CNN model is substantially less complex than the other CNNs
(it has only 72K weights and takes up 1.8M of memory, see \autoref{tbl:models-complexity})
but performs very similarly to the best pre-trained model (Inception: 30M weights, 347M of memory).
Nevertheless, we believe this accuracy range is not competitive enough for a real-world biometric application.
It is only when we incorporate temporal information that we start to observe noticeable improvements.
For example, the 1NN-DTW classifier outperforms all CNNs.
Furthermore, using velocity as the only input feature to a GRU architecture, all our performance metrics are above 94\%.
Of special mention is the AUC of 98\%, which indicates a highly discriminative power of the model.

Now that we know that RNN models outperform CNN models for this classification task,
we repeat the experiments for the remaining datasets.
We use the very same model architectures, hyperparameters, and configuration.
\autoref{tbl:results-dl} summarizes the results for our GRU classifier,
which is the overall best performer.
We report results for all RNN models in the Appendix.
We can see that our GRU model is able to distinguish between human and machine handwriting with remarkable accuracy,
ranging from 93\% to 97\%, except for the \$N-MMG dataset which was 87\%.
Precision, Recall, and F-measure also suggest excellent performance.
In any case, the AUC score is above 92\%, which indicates an outstanding discriminative power.
Nevertheless, there is still some room for improvement regarding the \$N-MMG dataset.
Interestingly, most symbols in this dataset comprise multi-stroke sequences,
and the time between those sequences (in-air strokes) is not taken into account by the \slm model.
We argue that this introduces additional noise in the velocity distribution
and therefore the synthetic and human \$N-MMG trajectories are a bit more difficult to distinguish.
A closer look at \autoref{tbl:input-device},
which splits the \$N-MMG dataset according to the two input devices available,
reveals that gestures articulated with the stylus are more difficult to classify
than gestures articulated with the finger.
This happens when the model is trained either on stylus-only or finger-only samples.
We elaborate more in `\nameref{sec:devices}' section.

\begin{table}[!ht]
    \caption{Experiment results with our GRU classifier.}
    \label{tbl:results-dl}

    \centering
    \setlength{\tabcolsep}{0.65em}

    \begin{tabular}{l *6r}
    \toprule
    \thead{Dataset} & \thead{Precision} & \thead{Recall} & \thead{F-measure} & \thead{Accuracy} & \thead{AUC} \\
    \midrule
    \$1-GDS     & 95.78 & 95.39 & 95.38 & 95.39 & 98.20 \\
    \$N-MMG     & 87.41 & 86.98 & 86.94 & 86.98 & 92.07 \\
    Chars74k    & 97.06 & 97.04 & 97.04 & 97.04 & 99.30 \\
    SUSIGv      & 93.68 & 93.35 & 93.34 & 93.35 & 95.43 \\
    \bottomrule
    \end{tabular}
\end{table}

For completeness, in the Appendix
we report the performance of all the RNN models
in all of our evaluated datasets (\autoref{tbl:models-performance})
as well as their complexity (\autoref{tbl:models-complexity}).
Overall, the classification results provided by the other models are similar
to what we observed in the analysis of the \$1-GDS dataset,
where the GRU classifier and its bi-directional variant outperformed the other models.

\section{Discussion and Future Work}

Synthetic handwritten generators such as the \slm model can produce human-like movements
that are almost indistinguishable from actual human movements if those are represented as off-line images.
However, if handwriting data are represented as on-line point sequences
it is possible to tell human and machine-generated movements apart with remarkable accuracy.
In other words, the distinctive badge of a human action is not \emph{what} it is written but \emph{how} it is written.
Our approach allows distinguishing next-generation impersonators of human handwriting from genuine human writers.
It thus has important implications for biometric and forensics systems, human behavior analysis, and motor control understanding.
Crucially, different data generation methods deserve to be considered to demonstrate the generalization ability of our GRU classifier.
So far, the \slm model is the only model we are aware of that can reconstruct generic, human-like spatio-temporal sequences.
We are aware of many other competing models that unfortunately ignore the temporal information~\cite{graves2013generating, Ha18, Taranta16_gpsr}.

We have investigated computational models to exploit the main differences
between real and synthetic handwriting specimens in both images and sequences to great advantage.
We found that RNNs outperform CNNs in this task.
Still, our RNN models could rely on other features as well such as the raw signal,
pen-tip position, or even pressure information (if available),
which could be combined to improve further the classification results.
In addition, adversarial training could help to make our models more robust,
provided that temporal data can be generated alongside spatial data,
which is rather challenging nevertheless~\cite{Elarian14}.
We leave these analyses as an interesting opportunity for future work.

Our work has the potential to enhance the security, reliability, and effectiveness of computer systems susceptible to spoof attacks.
Used wisely, a biometric-like handwriting verification system based on our findings can make users' life more comfortable.
Calm technology~\cite{Weiser96} can toil quietly in the background,
for example continuously authenticating account-holders
without badgering them for additional passwords or two-factor authentication.
Used unwisely, however, the system could become yet another electronic spy on people's privacy,
permitting strangers to monitor the user's every move.

It has been proposed recently that mobile devices could become
our Personal Digital Bodyguards (PDBs)~\cite{MartinAlbo16c_pdb, Plamondon18_pdb},
taking advantage of human movement control models and handwriting recognition technologies.
It is expected that, in a near future, PDBs will protect people's sensitive data with strong verification measures,
provide equipment security with writer authentication and recognition (e-security)
to monitor the user's fine motor control, which can detect stress, aging and health problems (e-health).
In the hands of children, these tools will turn into interactive toys helping them
to learn and master their fine motricity and become better writers and students (e-learning).
Such a vision might look out of reach, according to the current status of handheld technology,
though it is expected that breakthroughs made in these research domains
will put pressure on device vendors in the forthcoming years
in order to catch up and incorporate these e-applications ``by default'' in their operating systems.

\section{Conclusion}

We have contributed computational models to tackle the liveness detection problem
via handwriting symbols (isolated characters, digits, gestures, and signatures) and deep learning architectures.
We have found that if symbols are presented as off-line images, they can fool state-of-the-art classifiers
but can be distinguished with remarkable accuracy if they are presented as on-line sequences.
We conclude that an accurate detection of fake movements has more to do with \emph{how} users write,
rather than \emph{what} they write.

\section*{Acknowledgments}

We thank the anonymous referees for their constructive feedback.
L.\,A. Leiva acknowledges support from the Finnish Center for Artificial Intelligence (FCAI).
This research is supported by the Spanish MINECO (TEC2016-77791-C4-1-R project),
the European Union (FEDER program), and the NSERC (grant RGPIN-2015-06409).

\appendix

\subsection{Models complexity}

\autoref{tbl:models-complexity} summarizes the
complexity of our models,
informed by the usual proxy metrics in deep learning.
The `Params' column denotes the number of trainable model weights,
the `FLOPS' (Floating Point Operations Per Second) column
denotes the number of multiply-and-accumulate operations
(higher FLOPS denote more complexity),
and the `Memory' column denotes the model operational footprint.

\begin{table}[!ht]
  \caption{Deep learning models complexity.}
  \label{tbl:models-complexity}
  \centering

    \begin{tabular}{l *3r}
    \toprule
    \thead{System} & \thead{Params} & \thead{FLOPS} & \thead{Memory}\\
    \midrule
    VGG16         &  16M  &  33M &  81M \\
    ResNet50      &  32M  &  63M & 367M \\
    Xception      &  30M  &  58M & 336M \\
    DenseNet      &  26M  &  51M & 302M \\
    Inception     &  30M  &  60M & 347M \\
    Custom CNN    &  72K  & 289K & 1.8M \\
    \midrule
    Vanilla RNN   & 41K   &  40K & 148K \\
    LSTM          & 41K   & 161K & 507K \\
    GRU           & 31K   & 120K & 392K \\
    Bi-LSTM       & 83K   & 322K &   1M \\
    Bi-GRU        & 63K   & 241K & 765K \\
    \bottomrule
    \end{tabular}
\end{table}

\subsection{Models performance}

Given that RNN models outperformed CNN models in our liveness detection tasks,
we report in \autoref{tbl:models-performance} the performance of all the RNNs over all our evaluated datasets.
Notice that the performance of GRU and Bi-GRU models is very similar.
Therefore we decided to use GRU as the main RNN classifier in this paper,
since it is a simpler architecture.

\begin{table}[!ht]
  \caption{RNN models performance on all datasets.
    We also include the 1NN-DTW classifier as a baseline model.}
  \label{tbl:models-performance}
    \centering
    \setlength{\tabcolsep}{0.25em}

    \begin{tabular}{ll *5r}
    \toprule
    \thead{Dataset} & \thead{Model} & \thead{Precision} & \thead{Recall} & \thead{F-measure} & \thead{Accuracy} & \thead{AUC} \\
    \midrule
    \$1-GDS  & Vanilla    & 94.93     & 94.51     & 94.49     & 94.51     & 94.46     \\
             & LSTM       & 95.66     & 95.27     & 95.25     & 95.27     & 97.45     \\
             & Bi-LSTM    & 95.14     & 94.73     & 94.72     & 94.73     & 96.98     \\
             & GRU        & \bf 95.78 & \bf 95.39 & \bf 95.38 & \bf 95.39 & \bf 98.20 \\
             & Bi-GRU     & 95.62     & 95.20     & 95.19     & 95.20     & 97.76     \\
             & 1NN-DTW    & 85.32     & 83.97     & 83.80     & 83.97     & 83.88     \\
    \midrule
    \$N-MMG  & Vanilla    & 77.94     & 77.65     & 77.59     & 77.65     & 83.59     \\
             & LSTM       & 81.84     & 80.08     & 79.79     & 80.08     & 81.25     \\
             & Bi-LSTM    & 85.89     & 85.66     & 85.63     & 85.66     & 90.20     \\
             & GRU        & 87.41     & 86.98     & 86.94     & 86.98     & 92.07     \\
             & Bi-GRU     & \bf 87.52 & \bf 87.01 & \bf 86.96 & \bf 87.01 & \bf 92.14 \\
             & 1NN-DTW    & 71.28     & 62.02     & 57.51     & 62.02     & 62.18     \\
    \midrule
    Chars74k & Vanilla    & 91.31     & 90.34     & 90.28     & 90.34     & 91.64     \\
             & LSTM       & 92.92     & 92.90     & 92.90     & 92.90     & 98.46     \\
             & Bi-LSTM    & 95.43     & 95.27     & 95.26     & 95.27     & 99.25     \\
             & GRU        & \bf 97.06 & \bf 97.04 & \bf 97.04 & \bf 97.04 & 99.30     \\
             & Bi-GRU     & 96.60     & 96.60     & 96.60     & 96.60     & \bf 99.49 \\
             & 1NN-DTW    & 92.55     & 91.52     & 91.47     & 91.52     & 91.53     \\
    \midrule
    SUSIGv   & Vanilla    & 84.66     & 84.57     & 84.56     & 84.57     & 90.28     \\
             & LSTM       & 65.98     & 65.78     & 65.62     & 65.78     & 72.48     \\
             & Bi-LSTM    & 88.37     & 87.41     & 87.32     & 87.41     & 92.21     \\
             & GRU        & 93.68     & 93.35     & 93.34     & 93.35     & 95.43     \\
             & Bi-GRU     & \bf 95.00 & \bf 94.68 & \bf 94.67 & \bf 94.68 & \bf 97.29 \\
             & 1NN-DTW    & 72.65     & 65.16     & 61.82     & 65.16     & 64.85     \\
    \bottomrule
    \end{tabular}
\end{table}

\subsection{GRU robustness}

To further demonstrate the robustness of our GRU classifier,
we train it on different splits of the original training data
and report the results in \autoref{tbl:gru-robustness}.
The `Train' and `Test' columns denote the number of samples in each partition.
In all cases, the model is fine-tuned on 20\% of the training data.
The smaller the training split, the more likely that user-dependent samples will be left out.
As can be observed in the table, classification performance remains about the same in all splits.
Notice that when training on 99\% of the data, the model (1)~has almost full knowledge of the dataset distribution
and (2)~is tested on a small number of samples, therefore classification performance is usually higher.

\begin{table}[!ht]
  \caption{GRU robustness on all datasets.}
  \label{tbl:gru-robustness}
    \centering
    \setlength{\tabcolsep}{0.425em}

    \begin{tabular}{lr rl *5r}
    \toprule
    \thead{Dataset} & \thead{Split} & \thead{Train} & \thead{Test} &
    \thead{Prec.} & \thead{Recall} & \thead{F1} & \thead{Acc.} & \thead{AUC} \\
    \midrule
    \$1-GDS  & 10\% &  1056 & 9504  & 95.37 & 94.91 & 94.89 & 94.91 & 96.61 \\
             & 20\% &  2112 & 8448  & 95.39 & 94.92 & 94.91 & 94.92 & 97.49 \\
             & 40\% &  4224 & 6336  & 95.47 & 95.03 & 95.02 & 95.03 & 97.52 \\
             & 80\% &  8448 & 2112  & 96.03 & 95.69 & 95.68 & 95.69 & 98.18 \\
             & 99\% & 10455 & 106   & 96.48 & 96.23 & 96.22 & 96.23 & 98.00 \\
    \midrule
    \$N-MMG  & 10\% &  1919 & 17277 & 85.88 & 84.19 & 84.01 & 84.19 & 86.88 \\
             & 20\% &  3840 & 15356 & 87.81 & 87.26 & 87.21 & 87.26 & 92.32 \\
             & 40\% &  7678 & 11518 & 87.75 & 87.20 & 87.16 & 87.20 & 92.28 \\
             & 80\% & 15356 & 3840  & 88.19 & 87.45 & 87.38 & 87.45 & 92.39 \\
             & 99\% & 19004 & 192   & 87.31 & 86.98 & 86.95 & 86.98 & 93.21 \\
    \midrule
    Chars74k & 10\% &  675 & 6083   & 91.47 & 91.44 & 91.43 & 91.44 & 96.76 \\
             & 20\% & 1351 & 5407   & 97.28 & 97.28 & 97.28 & 97.28 & 99.57 \\
             & 40\% & 2703 & 4055   & 98.57 & 98.57 & 98.57 & 98.57 & 99.87 \\
             & 80\% & 5407 & 1351   & 97.87 & 97.86 & 97.85 & 97.86 & 99.88 \\
             & 99\% & 6690 & 68     & 98.58 & 98.53 & 98.53 & 98.53 & 99.91 \\
    \midrule
    SUSIGv   & 10\% &  376 & 3384   & 84.35 & 84.13 & 84.11 & 84.13 & 91.38 \\
             & 20\% &  752 & 3008   & 93.40 & 93.35 & 93.35 & 93.35 & 96.32 \\
             & 40\% & 1504 & 2256   & 92.92 & 92.69 & 92.68 & 92.69 & 95.80 \\
             & 80\% & 3008 & 752    & 95.75 & 95.48 & 95.46 & 95.48 & 97.47 \\
             & 99\% & 3722 & 38     & 100.0 & 100.0 & 100.0 & 100.0 & 100.0 \\
    \bottomrule
    \end{tabular}
\end{table}

\subsection{Effect of input device}
\label{sec:devices}

We train our GRU classifier on the \$N-MMG dataset conditioned on stylus or finger articulations,
since this dataset has both types of input data, and test the classifier on both types of input.
The results of these experiments are shown in \autoref{tbl:input-device}.
It can be observed that our classifier performs much better when tested on finger-only samples,
no matter if it was trained on stylus-only or finger-only samples.
We argue that, in this dataset, the stylus samples are of poor quality;
e.g., previous work~\cite{Leiva15_g3, Leiva17_slm}
has indicated that many samples have sparse coordinates and duplicated timestamps.
This explains the lower-than-usual performance of our GRU classifier when tested on these stylus samples.

\begin{table}[!ht]
  \caption{Effect of input device (\$N-MMG dataset).}
  \label{tbl:input-device}
    \centering
    \setlength{\tabcolsep}{0.5em}

    \begin{tabular}{ll *5r}
    \toprule
    \thead{Train} & \thead{Test} &
    \thead{Precision} & \thead{Recall} & \thead{F-measure} & \thead{Accuracy} & \thead{AUC} \\
    \midrule
    Stylus & Stylus & 83.27 & 79.31 & 78.75 & 79.31 & 86.88 \\
    Stylus & Finger & 93.21 & 92.29 & 92.25 & 92.29 & 97.25 \\
    Finger & Finger & 95.47 & 95.24 & 95.24 & 95.24 & 97.03 \\
    Finger & Stylus & 79.55 & 78.82 & 78.73 & 78.82 & 85.19 \\
    \bottomrule
    \end{tabular}
\end{table}

\balance


\begin{thebibliography}{10}
\providecommand{\url}[1]{#1}
\csname url@samestyle\endcsname
\providecommand{\newblock}{\relax}
\providecommand{\bibinfo}[2]{#2}
\providecommand{\BIBentrySTDinterwordspacing}{\spaceskip=0pt\relax}
\providecommand{\BIBentryALTinterwordstretchfactor}{4}
\providecommand{\BIBentryALTinterwordspacing}{\spaceskip=\fontdimen2\font plus
\BIBentryALTinterwordstretchfactor\fontdimen3\font minus
  \fontdimen4\font\relax}
\providecommand{\BIBforeignlanguage}[2]{{
\expandafter\ifx\csname l@#1\endcsname\relax
\typeout{** WARNING: IEEEtran.bst: No hyphenation pattern has been}
\typeout{** loaded for the language `#1'. Using the pattern for}
\typeout{** the default language instead.}
\else
\language=\csname l@#1\endcsname
\fi
#2}}
\providecommand{\BIBdecl}{\relax}
\BIBdecl

\bibitem{Andrews18}
N.~Andrews, ``{Can I Get Your Digits?} illegal acquisition of wireless phone
  numbers for {SIM}-swap attacks and wireless provider liability,'' \emph{Nw.
  J. Tech. \& Intell. Prop.}, vol.~16, no.~2, 2018.

\bibitem{Menotti15}
D.~{Menotti}, G.~{Chiachia}, A.~{Pinto}, W.~R. {Schwartz}, H.~{Pedrini}, A.~X.
  {Falcão}, and A.~{Rocha}, ``Deep representations for iris, face, and
  fingerprint spoofing detection,'' \emph{IEEE Trans. Inf. Forensics Secur.},
  vol.~10, no.~4, 2015.

\bibitem{Leiva13_smt}
L.~A. Leiva and R.~Viv\'{o}, ``Web browsing behavior analysis and interactive
  hypervideo,'' \emph{ACM Trans. Web}, vol.~7, no.~4, 2013.

\bibitem{Shirali-Shahreza11}
S.~Shirali-Shahreza and M.~Shirali-Shahreza, ``Multilingual highlighting
  {CAPTCHA},'' in \emph{Proc. ITNG}, 2011.

\bibitem{Leiva15_mucaptcha}
L.~A. Leiva and F.~Álvaro, ``$\mu$captcha: Human interaction proofs tailored
  to touch-capable devices via math handwriting,'' \emph{Int. J. Hum.-Comput.
  Interact.}, vol.~31, no.~7, 2012.

\bibitem{Ramahia14_capctha}
C.~Ramahia, R.~Plamondon, and V.~Govindaraju, ``A sigma-lognormal model for
  handwritten text {CAPTCHA} generation,'' in \emph{Proc. ICPR}, 2014.

\bibitem{diaz2015modeling}
M.~Diaz-Cabrera, M.~A. Ferrer, and A.~Morales, ``Modeling the lexical
  morphology of western handwritten signatures,'' \emph{PloS One}, vol.~10,
  no.~4, 2015.

\bibitem{Plamondon18_pdb}
R.~Plamondon, G.~Pirlo, E.~Anquetil, C.~Rémi, H.-L. Teuling, and M.~Nakagawa,
  ``Personal digital bodyguards for e-security, e-learning and e-health: A
  prospective survey,'' \emph{Pattern Recognit.}, vol.~81, 2018.

\bibitem{Galbally12b}
J.~Galbally, R.~Plamondon, J.~Fierrez, and J.~Ortega-García, ``Synthetic
  on-line signature generation. {Part II}: Experimental validation,''
  \emph{Pattern Recognit.}, vol.~45, no.~7, 2012.

\bibitem{8052226}
M.~A. {Ferrer}, S.~{Chanda}, M.~{Diaz}, C.~K. {Banerjee}, A.~{Majumdar},
  C.~{Carmona-Duarte}, P.~{Acharya}, and U.~{Pal}, ``Static and dynamic
  synthesis of {Bengali} and {Devanagari} signatures,'' \emph{IEEE T.
  Cybernetics}, vol.~48, no.~10, 2018.

\bibitem{Leiva17_slm}
L.~A. Leiva, D.~Martín-Albo, and R.~Plamondon, ``The {Kinematic Theory}
  produces human-like stroke gestures,'' \emph{Interact. Comput.}, vol.~29,
  no.~4, 2017.

\bibitem{Leiva17_gestures}
L.~A. Leiva, ``Large-scale user perception of synthetic stroke gestures,'' in
  \emph{Proc. DIS}, 2017.

\bibitem{Bhattacharya17}
U.~Bhattacharya, R.~Plamondon, S.~Chowdhury, P.~Goyal, and S.~Parui, ``A
  sigma-lognormal model based approach to generating large synthetic online
  handwriting samples databases,'' \emph{Int. J. Doc. Anal. Recogn.}, vol.~20,
  no.~71, 2017.

\bibitem{Reznakova17}
M.~Režnáková, L.~Tencer, R.~Plamondon, and M.~Cheriet, ``Forgetting of
  unused classes in missing data environment using automatically generated
  data: Application to on-line handwritten gesture command recognition,''
  \emph{Pattern Recognit.}, vol.~72, 2017.

\bibitem{plamondon2014recent}
R.~Plamondon, C.~O’reilly, J.~Galbally, A.~Almaksour, and {\'E}.~Anquetil,
  ``Recent developments in the study of rapid human movements with the
  {Kinematic Theory}: Applications to handwriting and signature synthesis,''
  \emph{Pattern Recogn. Lett.}, vol.~35, 2014.

\bibitem{Almaksour11}
A.~Almaksour, E.~Anquetil, R.~Plamondon, and C.~O'Reilly, ``Synthetic
  handwritten gesture generation using sigma-lognormal model for evolving
  handwriting classifiers,'' in \emph{Proc. IGS}, 2011.

\bibitem{diaz2019perspective}
M.~Diaz, M.~A. Ferrer, D.~Impedovo, M.~I. Malik, G.~Pirlo, and R.~Plamondon,
  ``A perspective analysis of handwritten signature technology,'' \emph{ACM
  Comput. Surv.}, vol.~51, no.~6, 2019.

\bibitem{Plamondon00}
R.~Plamondon and S.~N. Srihari, ``On-line and off-line handwriting recognition:
  a comprehensive survey,'' \emph{IEEE Trans. Pattern Anal. Mach. Intell.},
  vol.~22, no.~1, 2000.

\bibitem{Ghiani15_fingerprint}
L.~Ghiani, D.~A. Yambay, V.~Mura, G.~L. Marcialis, F.~Roli, and S.~A.
  Schuckers, ``Review of the fingerprint liveness detection ({LivDet})
  competition series: 2009 to 2015,'' \emph{Image Vision Comput.}, vol.~58,
  2017.

\bibitem{Xin17_face}
Y.~Xin, Y.~Liu, Z.~Liu, X.~Zhu, L.~Kong, D.~Wei, W.~Jiang, and J.~Chang, ``A
  survey of liveness detection methods for face biometric systems,''
  \emph{Sensor Review}, vol.~37, no.~3, 2017.

\bibitem{Chen18_iris}
Y.~Chen and W.~Zhang, ``Iris liveness detection: A survey,'' in \emph{Proc.
  IEEE BigMM}, 2018.

\bibitem{Hourcade08}
J.~P. Hourcade and T.~R. Berkel, ``Simple pen interaction performance of young
  and older adults using handheld computers,'' \emph{Interact. Comput.},
  vol.~20, no.~1, 2008.

\bibitem{Acien20_becaptcha}
A.~Acien, A.~Morales, J.~Fierrez, and R.~Vera-Rodriguez, ``{BeCAPTCHA-Mouse}:
  Synthetic mouse trajectories and improved bot detection,'' arXiv 2005.00890,
  2020.

\bibitem{graves2013generating}
A.~Graves, ``Generating sequences with recurrent neural networks,'' arXiv
  1308.0850, 2013.

\bibitem{Ha18}
D.~Ha and D.~Eck, ``A neural representation of sketch drawings,'' in
  \emph{Proc. ICLR}, 2018.

\bibitem{Taranta16_gpsr}
E.~M. Taranta, II, M.~Maghoumi, C.~R. Pittman, and J.~J. LaViola, Jr., ``A
  rapid prototyping approach to synthetic data generation for improved {2D}
  gesture recognition,'' in \emph{Proc. UIST}, 2016.

\bibitem{Elarian14}
Y.~Elarian, R.~Abdel-Aal, I.~Ahmad, M.~T. Parvez, and A.~Zidouri, ``Handwriting
  synthesis: classifications and techniques,'' \emph{Int. J. Doc. Anal.
  Recogn.}, vol.~17, no.~4, 2014.

\bibitem{hollerbach1981oscillation}
J.~M. Hollerbach, ``An oscillation theory of handwriting,'' \emph{Biol.
  Cybern.}, vol.~39, no.~2, 1981.

\bibitem{gangadhar2007oscillatory}
G.~Gangadhar, D.~Joseph, and V.~S. Chakravarthy, ``An oscillatory neuromotor
  model of handwriting generation,'' \emph{Int. J. Doc. Anal. Recogn.},
  vol.~10, no.~2, 2007.

\bibitem{choudhury2019synthesis}
H.~Choudhury and S.~M. Prasanna, ``Synthesis of handwriting dynamics using
  sinusoidal model,'' in \emph{Proc. ICDAR}, 2019.

\bibitem{Plamondon95a}
R.~Plamondon, ``A kinematic theory of rapid human movements. {Part I}: Movement
  representation and control,'' \emph{Biol. Cybern.}, vol.~72, no.~4, 1995.

\bibitem{Plamondon95b}
------, ``A kinematic theory of rapid human movements. {Part II}: Movement time
  and control,'' \emph{Biol. Cybern.}, vol.~72, no.~4, 1995.

\bibitem{Plamondon06}
R.~Plamondon and M.~Djioua, ``A multi-level representation paradigm for
  handwriting stroke generation,'' \emph{Hum. Mov. Sci.}, vol.~25, no. 4--5,
  2006.

\bibitem{Leiva15_g3}
L.~A. Leiva, D.~Martín-Albo, and R.~Plamondon, ``Gestures à go go: Authoring
  synthetic human-like stroke gestures using the kinematic theory of rapid
  movements,'' \emph{ACM Trans. Intell. Syst. Technol.}, vol.~7, no.~2, 2016.

\bibitem{7775072}
M.~{Diaz}, A.~{Fischer}, M.~A. {Ferrer}, and R.~{Plamondon}, ``Dynamic
  signature verification system based on one real signature,'' \emph{IEEE T.
  Cybernetics}, vol.~48, no.~1, 2018.

\bibitem{8585714}
M.~A. {Ferrer}, M.~{Diaz}, C.~{Carmona-Duarte}, and R.~{Plamondon}, ``A
  biometric attack case based on signature synthesis,'' in \emph{Proc. ICCST},
  2018.

\bibitem{MartinAlbo16_ipm}
D.~Martín-Albo, L.~A. Leiva, J.~Huang, and R.~Plamondon, ``Strokes of insight:
  User intent detection and kinematic compression of mouse cursor trails,''
  \emph{Inform. Process. Manag.}, vol.~56, no.~6, 2016.

\bibitem{Wobbrock07}
J.~O. Wobbrock, A.~D. Wilson, and Y.~Li, ``Gestures without libraries, toolkits
  or training: A \$1 recognizer for user interface prototypes,'' in \emph{Proc.
  UIST}, 2007.

\bibitem{Anthony10}
L.~Anthony and J.~O. Wobbrock, ``A lightweight multistroke recognizer for user
  interface prototypes,'' in \emph{Proc. GI}, 2010.

\bibitem{kholmatov2009susig}
A.~Kholmatov and B.~Yanikoglu, ``{SUSIG}: an on-line signature database,
  associated protocols and benchmark results,'' \emph{Pattern Anal. Appl.},
  vol.~12, no.~3, 2009.

\bibitem{o2009development}
C.~O'Reilly and R.~Plamondon, ``Development of a {Sigma--Lognormal}
  representation for on-line signatures,'' \emph{Pattern Recognit.}, vol.~42,
  no.~12, 2009.

\bibitem{Galbally12a}
J.~Galbally, R.~Plamondon, J.~Fierrez, and J.~Ortega-García, ``Synthetic
  on-line signature generation. {Part I}: Methodology and algorithms,''
  \emph{Pattern Recognit.}, vol.~45, no.~7, 2012.

\bibitem{chollet2016xception}
F.~Chollet, ``Xception: Deep learning with depthwise separable convolutions,''
  in \emph{Proc. CVPR}, 2017.

\bibitem{Simonyan15}
K.~Simonyan and A.~Zisserman, ``Very deep convolutional networks for
  large-scale image recognition,'' in \emph{Proc. ICLR}, 2015.

\bibitem{he2015deep}
K.~He, X.~Zhang, S.~Ren, and J.~Sun, ``Deep residual learning for image
  recognition,'' in \emph{Proc. CVPR}, 2016.

\bibitem{huang2016densely}
G.~Huang, Z.~Liu, L.~van~der Maaten, and K.~Q. Weinberger, ``Densely connected
  convolutional networks,'' in \emph{Proc. CVPR}, 2017.

\bibitem{szegedy2014going}
C.~Szegedy, W.~Liu, Y.~Jia, P.~Sermanet, S.~Reed, D.~Anguelov, D.~Erhan,
  V.~Vanhoucke, and A.~Rabinovich, ``Going deeper with convolutions,'' in
  \emph{Proc. CVPR}, 2015.

\bibitem{Deng09_imagenet}
J.~{Deng}, W.~{Dong}, R.~{Socher}, L.~{Li}, {Kai Li}, and {Li Fei-Fei},
  ``{ImageNet}: A large-scale hierarchical image database,'' in \emph{Proc.
  CVPR}, 2009.

\bibitem{Pan10_transfer}
S.~J. {Pan} and Q.~{Yang}, ``A survey on transfer learning,'' \emph{IEEE T.
  Knowl. Data En.}, vol.~22, no.~10, 2010.

\bibitem{Lin14_gap}
M.~Lin, Q.~Chen, and S.~Yan, ``Network in network,'' in \emph{Proc. ICLR},
  2014.

\bibitem{Kheradpisheh16}
S.~R. Kheradpisheh, M.~Ghodrati, M.~Ganjtabesh, and T.~Masquelier, ``Deep
  networks can resemble human feed-forward vision in invariant object
  recognition,'' \emph{Sci. Rep.}, vol.~6, no. 32672, 2016.

\bibitem{Kingma15_adam}
D.~P. Kingma and J.~L. Ba, ``Adam: A method for stochastic optimization,'' in
  \emph{Proc. ICLR}, 2015.

\bibitem{Hochreiter97_lstm}
S.~Hochreiter and J.~Schmidhuber, ``Long short-term memory,'' \emph{Neural
  Comput.}, vol.~9, no.~8, 1997.

\bibitem{Cho14_gru}
K.~Cho, B.~van Merrienboer, C.~Gulcehre, F.~Bougares, H.~Schwenk, and
  Y.~Bengio, ``Learning phrase representations using {RNN} encoder-decoder for
  statistical machine translation,'' in \emph{Proc. EMNLP}, 2014.

\bibitem{Graves05_bidir}
A.~Graves and J.~Schmidhuber, ``Framewise phoneme classification with
  bidirectional {LSTM} and other neural network architectures,'' \emph{Neural
  Netw.}, vol.~18, no. 5--6, 2005.

\bibitem{Bernier05}
P.-M. Bernier, R.~Chua, and I.~M. Franks, ``Is proprioception calibrated during
  visually guided movements?'' \emph{Exp. Brain Res.}, vol. 167, 2005.

\bibitem{Hartwig01}
H.~R. Siebner, C.~Limmer, A.~Peinemann, P.~Bartenstein, A.~Drzezga, and
  B.~Conrad, ``Brain correlates of fast and slow handwriting in humans: a
  {PET}-performance correlation analysis,'' \emph{Eur. J. Neurosci.}, vol.~14,
  2001.

\bibitem{sasaki2007truth}
Y.~Sasaki, ``The truth of the {F-measure},'' 2007.

\bibitem{Kate16}
R.~J. Kate, ``Using dynamic time warping distances as features for improved
  time series classification,'' \emph{Data Min. Knowl. Discovery}, vol.~30,
  no.~2, 2016.

\bibitem{Weiser96}
M.~Weiser and J.~S. Brown, ``Designing calm technology,'' \emph{PowerGrid
  Journal}, vol.~1, no.~1, 1996.

\bibitem{MartinAlbo16c_pdb}
D.~Martín-Albo, L.~A. Leiva, and R.~Plamondon, ``On the design of personal
  digital bodyguards: Impact of hardware resolution on handwriting analysis,''
  in \emph{Proc. ICFHR}, 2016.

\end{thebibliography}
\end{document}